\documentclass{article}

\usepackage{arxiv}

\usepackage[utf8]{inputenc} 
\usepackage[T1]{fontenc}    
\usepackage{hyperref}       
\usepackage{url}            
\usepackage{booktabs}       
\usepackage{amsfonts}       
\usepackage{microtype}      
\usepackage{lipsum}         
\usepackage{graphicx}
\usepackage{natbib}
\usepackage{doi}
\usepackage{wrapfig}
\usepackage{caption}
\usepackage{subcaption}
\usepackage{float}
\usepackage{multicol}
\usepackage{enumitem}
\usepackage{xcolor}         
\usepackage{xspace}
\usepackage[disable]{todonotes}

\usepackage{amsmath,amssymb,amsthm,mathabx,mathtools} 
\usepackage{bbold} 
\usepackage{nicefrac}       
\usepackage{thmtools}       
\usepackage{thm-restate}    
\usepackage[capitalise,noabbrev]{cleveref}    
\usepackage[ruled,linesnumbered,vlined]{algorithm2e} 
\usepackage{tikz} 
\usepackage{tikz-3dplot}
\usepackage{pgfplots}
\usepackage{standalone}

\usetikzlibrary{arrows.meta}
\usetikzlibrary{external}
\usepackage{amsmath,amsfonts,bm}
\usepackage{xargs}

\def\eqref#1{equation~\ref{#1}}

\def\1{\bm{1}}

\def\vtheta{{\bm{\theta}}}

\def\ve{{\bm{e}}}

\DeclareMathAlphabet{\mathsfit}{\encodingdefault}{\sfdefault}{m}{sl}
\SetMathAlphabet{\mathsfit}{bold}{\encodingdefault}{\sfdefault}{bx}{n}

\newcommand{\R}{\mathbb{R}}

\DeclareMathOperator*{\argmax}{arg\,max}
\DeclareMathOperator*{\argmin}{arg\,min}

\usepackage{amssymb,amsthm,mathabx,mathtools}
\usepackage{bbold} 
\usepackage{nicefrac}       
\usepackage{thmtools}       
\usepackage{thm-restate}    

\newcommand{\utheta}{u_{\vtheta}}
\newcommand{\uthetaT}[1][t]{u_{\vtheta_{#1}}}
\newcommand{\Mf}{\mathcal{M}}
\newcommand{\feat}{\widehat{\phi}}
\newcommand{\ft}[1][t]{\widehat{\phi}_{#1}}
\newcommand{\Using}{\widehat{U}}

\newcommand{\fUsing}{U_\vtheta}
\newcommand{\fUsi}[1][i]{U_{\vtheta,{#1}}}
\newcommand{\Vsing}{\widehat{V}}
\newcommand{\Vsi}[1][i]{\widehat{V}_{t,{#1}}}
\newcommand{\fVsing}{V_\vtheta}
\newcommand{\fVsi}[1][i]{V_{t,{#1}}}
\newcommand{\Dsing}{\widehat{\Delta}}
\newcommandx{\Dsi}[2][1=i,2=t]{\widehat{\Delta}_{#2,#1}}
\newcommand{\fDsing}{\Delta_\vtheta}
\newcommand{\fDsi}[1][i]{\Delta_{\vtheta,{#1}}}
\newcommand{\FRk}{\textrm{r}_{\text{svd}}}
\newcommandx{\TMN}[2][1=N,2=M]{T^{#2}_{#1}\Mf}
\newcommand{\TZN}[1][N]{T^{0}_{#1}\Mf}
\newcommandx{\eTMN}[2][1=N,2=M]{\widehat{T^{#2}_{#1}\Mf}}
\newcommand{\eTZN}[1][N]{\widehat{T^{0}_{#1}\Mf}}
\newcommand{\rce}{\textrm{RCE}}
\newcommand{\ePr}{\epsilon}
\newcommand{\cutoff}{\alpha_t}
\newcommand{\Ccutoff}{\alpha}
\newcommand{\Nflat}{N_{\text{flat}}}
\newcommand{\NflatInf}{N_{\text{flat}}^{\infty}}
\newcommand{\rkk}{\textrm{r}_{\text{cutoff}}}
\newcommand{\rkint}{\textrm{r}_{\text{int}}}
\newcommand{\rkeps}{\textrm{r}_{\ePr}}

\newcommand{\liftoff}{\lambda}
\newcommand{\rkmax}{\textrm{r}_{\rm max}}
\newcommand{\rkmin}{\textrm{r}_{\rm min}}
\newcommand{\rkOne}{\textrm{r}_1}
\newcommand{\rkTwo}{\textrm{r}_2}
\newcommand{\cumapprox}{\widehat{\Sigma}} 
\newcommand{\rotatedgrad}{\widehat{c}}

\newcommand{\Floss}{\cL}

\newcommand{\FLgrad}{\nabla \cL}
\newcommand{\ELgrad}{\widehat{\nabla \cL}}

\newcommand{\CC}{\mathbb{C}}

\newcommand{\KK}{\mathbb{K}}
\newcommand{\NN}{\mathbb{N}}

\newcommand{\RR}{\mathbb{R}}

\newcommand{\cA}{\mathcal{A}}
\newcommand{\cB}{\mathcal{B}}
\newcommand{\cC}{\mathcal{C}}
\newcommand{\cG}{\mathcal{G}}

\newcommand{\cH}{\mathcal{H}}
\newcommand{\cL}{\mathcal{L}}
\newcommand{\cM}{\mathcal{M}}

\newcommand{\cT}{\mathcal{T}}

\newcommand{\dd}{\textrm{d}}

\newcommand{\dI}{\textrm{I}}

\newcommand{\dL}{\textrm{L}}

\newcommand{\braket}[2]{\left\langle #1\,,\, #2\right\rangle}
\newcommand{\norm}[1][\cdot]{\left\|#1\right\|}

\DeclareMathOperator*\Span{Span}
\DeclareMathOperator\Ker{Ker}
\DeclareMathOperator\Ima{Im}

\DeclareMathOperator\diag{diag}
\newcommand{\Spectrum}{\sigma}

\def\forallt{\textrm{for all }}

 \newcommand{\ie}{\textit{i.e.}\xspace}
 
 \newcommand{\cf}{\textit{cf.}\xspace}
 
 \newcommand{\apriori}{\textit{a priori}\,}
 
 \newcommand{\via}{\textit{via}~}

\declaretheorem[name=Definition,style=definition]{Def}
\declaretheorem[name=Definition,style=definition,numbered=no]{Def*}
\declaretheorem[name=Proposition,style=plain]{Prop}
\declaretheorem[name=Proposition,style=plain,numbered=no]{Prop*}

\declaretheorem[name=Theorem,style=plain,numbered=no]{Th*}
\declaretheorem[name=Lemma,style=plain]{LM}
\declaretheorem[name=Lemma,style=plain,numbered=no]{LM*}
\declaretheorem[name=Corollary,style=plain]{Cor}
\declaretheorem[name=Corollary,style=plain,numbered=no]{Cor*}
\declaretheorem[name=Remark,style=remark]{Rk}

\usepackage{environ}
\newcommand\literallabel[2]{%
  \label{#1}
  \expandafter\gdef\csname literallabel_#1\endcsname{#2}
  #2
}
\NewEnviron{literaleq}[1]{%
  \begin{equation}
  \label{#1}
  \expandafter\gdef\csname literallabel_#1\expandafter\endcsname\expandafter{\BODY}
  \BODY
  \end{equation}%
}
\newcommand\literalref[1]{\csname literallabel_#1\endcsname}

\NewEnviron{literalign}[1]{%
  \begin{align}
  \label{#1}
  \expandafter\gdef\csname literallabel_#1\expandafter\endcsname\expandafter{\BODY}
  \BODY
  \end{align}
}
\newcommand\literalignref[1]{\csname literallabel_#1\endcsname}

\newcommand{\method}{AMStraMGRAM\xspace}

\makeatletter
\renewcommand{\todo}[2][]{\tikzexternaldisable\@todo[#1]{#2}\tikzexternalenable}
\makeatother
\newcommand{\todoAS}[2][]{\todo[color = blue!20,#1]{\textbf{AS:} #2}}

\newcommand{\todoCR}[2][]{\todo[color = teal!70,#1]{\textbf{CR:} #2}}

\SetNlSty{textbf}{\color{black}}{}

\SetCommentSty{mycommentfont}

\title{\method: Adaptive Multi-cutoff Strategy Modification for ANaGRAM}

\date{\today}

\author{%
  Nilo Schwencke\\
  LISN$-$Université Paris-Saclay$-$INRIA-Saclay\\
  \texttt{nilo.schwencke@protonmail.com}
  \And
  Cyriaque Rousselot\\
  LISN$-$Université Paris-Saclay$-$INRIA-Saclay\\
  \texttt{cyriaque.rousselot@inria.fr}
  \And
  Alena Shilova\\
  LISN$-$Université Paris-Saclay$-$INRIA-Saclay\\
  \texttt{alena.shilova@inria.fr}
  \And
  Cyril Furtlehner\\
  LISN$-$Université Paris-Saclay$-$INRIA-Saclay\\
  \texttt{cyril.furtlehner@inria.fr}
}

\renewcommand{\shorttitle}{\textit{\method} : Adaptive Multi-cutoff Strategy for ANaGRAM}

\hypersetup{
pdftitle={\method: Adaptive Multi-cutoff Strategy Modification for ANaGRAM},
pdfsubject={math.NA, cs.NE, cs.LG},
pdfauthor={Nilo Schwencke, Cyriaque Rousselot, Alena Shilova, Cyril Furtlehner},
pdfkeywords={PINNs, SciML, PDEs, NatGrad, NTK, Natural Gradient, Physics-Informed Neural Networks},
}

\begin{document}
\maketitle

\begin{abstract}
  Recent works have shown that natural gradient methods can significantly outperform standard optimizers when training physics-informed neural networks (PINNs). In this paper, we analyze the training dynamics of PINNs optimized with ANaGRAM, a natural-gradient-inspired approach employing singular value decomposition with cutoff regularization. Building on this analysis, we propose a multi-cutoff adaptation strategy that further enhances ANaGRAM's performance. Experiments on benchmark PDEs validate the effectiveness of our method, which allows to reach machine precision on some experiments. To provide theoretical grounding, we develop a framework based on spectral theory that explains the necessity of regularization and extend previous shown connections with Green's functions theory.
\end{abstract}


\section{Introduction}

Physics-informed neural networks (PINNs) have recently emerged as a promising alternative for the numerical solution of partial differential equations (PDEs) \citep{raissiPhysicsinformedNeuralNetworks2019}. By leveraging neural networks as universal function approximators \citep{leshnoMultilayerFeedforwardNetworks1993}, PINNs replace traditional mesh-based discretizations with sampling-based collocation methods, enabling a straightforward extension to high-dimensional domains. This mesh-free formulation not only circumvents the “curse of dimensionality” inherent in grid-based approaches, but also allows continuous evaluation of the solution throughout the domain without explicit mesh generation \citep{cuomoScientificMachineLearning2022}.

Despite these advantages, achieving low training error with PINNs remains a major challenge \citep{wangExpertsGuideTraining2023,urbanUnveilingOptimizationProcess2025,kiyaniWhichOptimizerWorks2025,deryckOperatorPreconditioningPerspective2024}.
Open questions include how to select and distribute collocation points, how to balance the PDE residual against boundary-condition penalties, and which optimization strategies most effectively minimize the composite loss \citep{krishnapriyanCharacterizingPossibleFailure2021a,wangUnderstandingMitigatingGradient2021,mcclennySelfAdaptivePhysicsInformedNeural2022}.

A different line of research has recently reexamined PINNs from the perspective of functional geometry \citep{mullerAchievingHighAccuracy2023,mullerPositionOptimizationSciML2024a,jniniGaussNewtonNaturalGradient2024}, providing a mathematically principled view of the training dynamics.
In this vein, the ANaGRAM algorithm \citep{schwencke2025anagram} applies a natural-gradient update \citep{amariNaturalGradientWorks1998,ollivierRiemannianMetricsNeural2015a}, based on a reinterpretation and generalization of the neural tangent kernel (NTK; \citet{jacotNeuralTangentKernel2018}) as the kernel of the projection onto the neural network's tangent space.
This leads to a notion of the empirical natural gradient that projects the true functional gradient onto the empirical tangent space, yielding significantly faster convergence and lower errors compared to standard optimizers on PDE benchmarks.

Nevertheless, while ANaGRAM improves over standard optimizers, it still falls short of the accuracy attained by classical mesh-based methods, such as the finite element method \citep{grossmannCanPhysicsinformedNeural2024}.
Moreover, its final performance is highly affected by the way the pseudo-inverse of the feature matrix is computed. In particular, ANaGRAM sets a fixed level of \emph{cutoff}: a value below which the singular values of the feature matrix are ignored, \ie it controls how much loss signal is incorporated into an update. ANaGRAM's cutoff is currently chosen manually, as no automatic selection procedure has been proposed.

In this paper, we study the performance and training dynamics of ANaGRAM, with a particular focus on the role of the chosen cutoff. Typically, the training loss of ANaGRAM exhibits the slow convergence at the early iterations followed by a sudden drop at the end of the training -- similar behavior is shown by the eNGD method \citep{mullerAchievingHighAccuracy2023}. We discover that it is closely connected to what we further refer as the \emph{flattening phenomenon}, which we define and characterize using the \emph{reconstruction error}: a novel metric that measures how much of the loss signal is lost by different choices of cutoffs. Relying on the adaptive multi-cutoff strategy, our new algorithm \method  manages to capitalize on this phenomenon, resulting in a significant improvement (of several orders of magnitude) on various PDE benchmarks. To complement our empirical findings, we also present a functional-analytic view linking cutoff (and ridge regularization) to (generalized) Green operator theory, clarifying why cutoff regularization is essential and not just a mere fix to stabilize training.

\section{Problem Statement}

\subsection{Differential Operators and Physics-Informed Neural Networks (PINNs)}\label{subsec:PINNs-notation}

Let $\Omega \subset \RR^d$ be a domain. We introduce two operators, $D$ and $B$, defined on a Hilbert space $\cH$ of real-valued functions, acting respectively on $\Omega$ and on its boundary $\partial\Omega$:
\begin{align}\label{eqn:differential-operators-definition}
  D: & \left\{\begin{array}{lll}
                \cH & \to     & \dL^2(\Omega \to \RR,\mu) \\
                u   & \mapsto & D[u]
              \end{array}\right.,            &
  B: & \left\{\begin{array}{lll}
                \cH & \to     & \dL^2(\partial\Omega \to \RR,\sigma) \\
                u   & \mapsto & B[u]
              \end{array}\right. .
\end{align}

Here, $D$ denotes a differential operator, while $B$ represents a boundary operator.
A function $u \in \cH$ is said to be a \emph{classical solution} to the \emph{Partial Differential Equation} (PDE) associated with $D$ and $B$ if it satisfies
\begin{literaleq}{eqn:pinn-strong-pde}
  \begin{cases}
    D(u) = f \in \dL^2(\Omega \to \RR, \mu),            & \text{in } \Omega,         \\
    B(u) = g \in \dL^2(\partial\Omega \to \RR, \sigma), & \text{on } \partial\Omega,
  \end{cases}
\end{literaleq}

A \emph{physics-informed neural network} (PINN) approximates the solution $u$ by a parametric model $\utheta$, where $\utheta$ is a neural network with parameters $\vtheta \in \RR^P$. The learning objective is to minimize the empirical loss
\begin{equation}\label{eqn:empirical-pinn-loss}
  \ell_{D,B}(\vtheta) :=
  \frac{1}{2S_D} \sum_{i=1}^{S_D} \left(D[\utheta](x_i^D) - f(x_i^D)\right)^2
  + \frac{1}{2S_B} \sum_{i=1}^{S_B} \left(B[\utheta](x_i^B) - g(x_i^B)\right)^2.
\end{equation}

\subsection{PINNs Optimizers}
\label{sec:pinn-optimizers}

Training PINNs is notoriously challenging. Issues such as spectral bias, where networks struggle to learn high-frequency components, and the difficulty of balancing residual and boundary loss terms—often with vastly different magnitudes— result in unsatisfactory performance of standard deep learning optimizers \citep{wangUnderstandingMitigatingGradient2021, deryckOperatorPreconditioningPerspective2024, krishnapriyanCharacterizingPossibleFailure2021a, liuPreconditioningPhysicsInformedNeural2024}.

To mitigate these challenges, researchers have proposed various strategies. These include adaptive sampling approaches that focus on regions with high error \citep{krishnapriyanCharacterizingPossibleFailure2021a}, dynamic loss weighting schemes \citep{mcclennySelfAdaptivePhysicsInformedNeural2022}, and architectural modifications \citep{wangPirateNetsPhysicsinformedDeep2024}. Another promising line of research has focused on modifying the optimizers. In particular, two main branches of optimization approaches for PINNs have emerged:

\begin{enumerate}[label=(\roman*)]
  \item \textbf{Second-Order Methods.} These methods, based on Quasi-Newton techniques, particularly the BFGS algorithm \citep[Chapter 6]{nocedalNumericalOptimization1999} and its memory-efficient approximation L-BFGS \citep{liuLimitedMemoryBFGS1989}, address some of the training difficulties by considering the curvature of the loss landscape. This curvature arises from the non-linearities of both the neural network and the differential operators \citep{rathoreChallengesTrainingPINNs2024}. Recently, \citet{urbanUnveilingOptimizationProcess2025} extended this approach by modifying the self-scaled BFGS (SSBFGS; \citealp{albaaliNumericalExperienceClass1998}) and self-scaled Broyden (SSBroyden; \citealp{al-baaliWideIntervalEfficient2005}), along with other computational enhancements such as point resampling \citep{wuComprehensiveStudyNonadaptive2023} and boundary condition enforcement \citep{wangExpertsGuideTraining2023}, achieving state-of-the-art results \citep{kiyaniWhichOptimizerWorks2025}.

  \item \textbf{Natural Gradient Methods.} In contrast to second-order methods, natural gradient methods are \textbf{first-order} techniques\footnote{contrary to a widespread misconception, which arises from their analogy in the context of information theory} that provide a principled way to incorporate the geometry and metric structure of the problem space. Initially introduced in the context of information geometry by \citet{amariNaturalGradientWorks1998} and later extended by \citet{ollivierRiemannianMetricsNeural2015a}, these methods were introduced for PINNs by \citet{mullerAchievingHighAccuracy2023}. In subsequent work, \citet{schwencke2025anagram} connected these methods to kernel methods, yielding an efficient implementation they linked to Green's function theory \citep{duffyGreensFunctionsApplications2015}.
\end{enumerate}

\subsection{Natural Gradient Methods for PINNs}\label{subsec:natural-gradient-overview}

As a preliminary observation highlighted in \citet[Section 4.1]{schwencke2025anagram}, PINNs can be interpreted as a quadratic regression problem. This viewpoint arises naturally once the parametric model $\utheta$ is replaced with the following compound model:
\begin{equation}\label{eqn:coumpound-model-definition}
  (D,B)\circ u:\left\{\begin{array}{lllll}
    \RR^P   & \to     & \cH     & \to     & \dL^2(\Omega,\mu)\times \dL^2(\partial\Omega,\sigma) \\
    \vtheta & \mapsto & \utheta & \mapsto & (D[\utheta],B[\utheta])
  \end{array}\right..
\end{equation}

For ease of exposition, and without loss of generality, we restrict attention to regression in $\dL^2(\Omega,\mu)$.
Given $f \in \dL^2(\Omega, \mu)$, we define the associated empirical loss
\begin{equation}\literallabel{eqn:empirical-quadratic-loss}
  {\ell(\vtheta) := \frac{1}{2S} \sum_{i=1}^{S} \left(\utheta(x_i) - f(x_i)\right)^2},
\end{equation}
which can be seen as a discretization of the functional loss
\begin{equation}\literallabel{eqn:functional-loss}
  {\mathcal{L}(u) := \tfrac{1}{2} \norm[{u - f}]^2_{\dL^2(\Omega, \mu)}, \qquad u \in \dL^2(\Omega, \mu)}.
\end{equation}

The natural gradient approach seeks to compute the optimal update direction in function space and then pull it back to parameter space.
A single Fréchet derivative of the functional loss \cref{eqn:functional-loss} yields $\FLgrad_{|u} = u-f$.
The key insight is that admissible updates are constrained to the tangent space of the parametric model,
\begin{equation}\literallabel{eqn:tangent-space}
  {T_{\vtheta}\mathcal{M} := \mathrm{Im}(\mathrm{d}\utheta) = \Span \left( \partial_p \utheta : 1 \leq p \leq P \right) \subset \cH},
\end{equation}
where $\mathcal{M} := \mathrm{Im}(u) = \{ \utheta : \vtheta \in \RR^P \} \subset \cH$ is the manifold of functions parametrized by $\vtheta$.
Thus, the optimal update in function space is the projection of $\FLgrad_{|u}$ onto the tangent space (\cf\cref{fig:projection-illustration}), 
\begin{align}\literallabel{eqn:functional-natural-gradient-update}
  {u_{\vtheta_{t+1}} & \gets \uthetaT - \eta\, \Pi_{T_{\vtheta_t} \mathcal{M}}\!\left( \FLgrad_{\uthetaT} \right)\,;                                              
                     &
  \vtheta_{t+1}      & \gets \vtheta_t - \eta \, \mathrm{d} \uthetaT^\dagger\!\left( \Pi_{T_{\vtheta_t} \mathcal{M}}\!\left( \FLgrad_{\uthetaT} \right) \right)},
\end{align}
where the second equation is simply the pullback of the functional update to parameter space. 
We prove in \cref{app:proof-gram-matrix-nat-grad-formula} that this update is equivalent to the Gram–matrix formulation:
\begin{align}
  \literallabel{eqn:matrix-formula-natural-gradient-update}
  {\vtheta_{t+1}        & \gets\vtheta_t - \eta\,G_{\vtheta_t}^\dagger\nabla\ell(\vtheta_t)\,;
                        &
  {G_{\vtheta_t}}_{p,q} & :=\braket{\partial_p u_{\vtheta_t}}{\partial_q u_{\vtheta_t}}_{\dL^2(\Omega,\mu)}}.
\end{align}

\subsection{ANaGRAM: Empirical Natural Gradient}

The $O(P^3)$ complexity of matrix inversion in \cref{eqn:matrix-formula-natural-gradient-update} renders a direct implementation prohibitively expensive.
ANaGRAM \citep{schwencke2025anagram} circumvents this by exploiting a motivated approximation.
The key observation is that the update can be expressed in terms of the empirical feature matrix $\feat \in \RR^{S \times P}$ and the empirical functional residuals $\widehat{\Floss_{\vtheta}} \in \RR^{S}$:
\begin{align}\literallabel{eqn:eng-computation-trick}
  {\vtheta_{t+1}                    & \gets \vtheta_t - \eta\, \feat^\dagger \ELgrad_{\vtheta_t}; &
  \feat_{i,p}                       & := \partial_p \utheta(x_i);                                 &
  \left(\ELgrad_{\vtheta} \right)_i & := \utheta(x_i) - f(x_i)}.
\end{align}

Here, the pseudo-inverse is computed via singular value decomposition (SVD):
$\feat^\dagger = \Using \Dsing^\dagger \Vsing^T$ with $\feat = \Vsing \Dsing \Using^T$,
where $\Using \in \RR^{P \times \FRk}$, $\Dsing \in \RR^{\FRk \times \FRk}$, $\Vsing \in \RR^{S \times \FRk}$, and $\FRk = \min(P, S)$.
This reduces computational cost to $O(\min(PS^2, P^2S))$, which is tractable in practice.
A comparable complexity was later obtained by \citet{guzman-corderoImprovingEnergyNatural2025} using a Cholesky factorization approach.

For further details on the derivation of the empirical natural gradient, we refer to \citet{schwencke2025anagram}.
In what follows, we adopt a slight abuse of notation by omitting the explicit dependence on $\vtheta$ whenever it is clear from context.
When iteration indices matter, we explicitly write $t$ to emphasize the connection to $\vtheta_t$.

\todoAS[inline]{I added an above sentence, but I haven't changed the notations yet to follow this rule.}

\subsection{Regularization}\label{sec:regularization-nat-grad}

As discussed in \cref{app:why-regularization-is-necessary}, the type of problem we consider is ill-conditioned, which necessitates the use of regularization.
We distinguish between two main regularization schemes: (i) \emph{ridge regression}, which consists in adding a factor $\alpha^2 I_d$ (or, according to conventions, $\alpha^{-2} I_d$) to the Gram matrix $G_\vtheta$ in \cref{eqn:matrix-formula-natural-gradient-update} (or its approximation $\widehat{\cG}_{\vtheta}$), thereby making it invertible or (ii) \emph{cutoff regularization}, a scheme that applies a binary threshold (used in ANaGRAM):
\begin{equation}\label{eqn:cutoff-regularization}
  \Dsi^\dagger =
  \begin{cases}
    \Dsi^{-1}, & \text{if } \Dsi \geq \alpha, \\
    0,         & \text{otherwise}.
  \end{cases}
\end{equation}
Here $\alpha$ denotes the cutoff threshold. This regularization is the focus of our analysis in \cref{sec:anagram-insights}. For completeness, we provide a geometric interpretation of each scheme in \cref{app:geometric-interpretation-regularization}.
We further show that cutoff regularization extends previously established connections between natural gradient methods and Green's function theory \citep{schwencke2025anagram}. In particular, we obtain: 
\begin{restatable}{Th}{GreenFunction}\label{Th:general-generalized-green-function}
  The generalized Green's function of the operator $D$ in the regularized space $\cH_{D,\cH_0}^\alpha$ is given, for all $x,y \in \Omega$, by\vspace{-.2cm}
  \begin{equation}
    g_D(x,y) := D[k_D(x,\cdot)](y),
  \end{equation}
\end{restatable}
where $\cH_{D,\cH_0}^\alpha$ is a regularized space with reproducing kernel $k_D$, defined in \cref{sec:connection-with-green-functions}.

\section{Insights on ANaGRAM's Training Dynamics}\label{sec:anagram-insights}

In this section, we will look at  relevant quantities of interest to understand this empirical phenomenon.

\subsection{Reconstruction Error of Functional Gradient}\label{subsec:quantities-of-interest}

Let $\vtheta\in\RR^P$, the empirical feature matrix $\feat \in \mathbb{R}^{S \times P}$, and the empirical functional gradient $\ELgrad \in \mathbb{R}^S$ as defined in \cref{eqn:eng-computation-trick}.
Let us consider various empirical tangent spaces formed by taking different ranges of right singular vectors of $\feat = \Using \Dsing\Vsing^T$,  \ie $\eTMN = \Span(\Vsi\, : \, M\leq i \leq N)$.
For $1\leq N\leq \FRk$, reconstruction error measures how much information from the functional gradient signal is lost when considering only first $N$ components in SVD (the error caused by the projection onto the empirical tangent space $\eTZN$) is defined as follows
\begin{equation}
  \literallabel{eqn:reconstruction-loss-orthogonal}
  {\rce^S_N
    := \frac{1}{\sqrt{S}}\norm[\Vsing\Pi^0_N\Vsing^T\ELgrad - \ELgrad]_{\RR^S}} = \frac1{\sqrt{S}} \|\Pi^{\bot}_{\eTZN} \ELgrad - \ELgrad\|,
\end{equation}
where we define $\Pi^M_N \in \mathbb{R}^{\FRk \times \FRk}$ as a projection operator onto $\eTMN$:
\begin{equation}\label{eqn:PiMN-def}
  \Pi^M_N
  =\sum_{p=M+1}^N \ve^{(p)}\ve^{(p)^T},
\end{equation}
with $(\mathbf{e}^{(p)})_{1\leq p\leq\FRk}$ being the canonical basis of $\RR^{\FRk}$.

\begin{restatable}{Prop}{RCEproperties}\label{Prop:RCE-as-N-components-projection}
  $\rce^S_N$ is a non-increasing function of $N$, \ie for all $1\leq M, N \leq\FRk$:
  \begin{equation}
    M\leq N\implies \rce^S_M\geq\rce^S_N.
  \end{equation}
  Furthermore, assuming that $(x_i)_{i=1}^S$ are \textit{i.i.d} sampled from $\mu$, we have $\mu$-almost surely
  \begin{align}
    \lim\limits_{S\to\infty} \rce^S_N
    = \norm[\FLgrad_{\utheta}-\Pi^\bot_{\TZN}\FLgrad_{\utheta}]_{\dL^2(\Omega,\mu)}
    =\norm[\Pi^\bot_{\left[{\TZN}\right]^\bot}\FLgrad_{\utheta}]_{\dL^2(\Omega,\mu)},
  \end{align}
  where $\TMN = \Span(\fVsi\,:\,M\leq i\leq N)$,
  while $\left(\fVsi\right)_{1\leq i\leq \FRk}$ are the right singular-vectors of the differential $\dd \utheta$ ordered in a decreasing order according to their associated singular values.
\end{restatable}

\todoCR[inline]{Maybe this remark goes in the appendix}
\begin{Rk}\label{rk:empiricalSVDvsfullSVD}
  Note that
  $\Vsi \in \RR^{S}$ for $i \in 1, \dots, N$, the right singular vectors of $\feat$, can be seen as discretized versions of $\fVsi$ from \cref{Prop:RCE-as-N-components-projection}. Indeed, a weak convergence holds, \ie $\forall h \in \mathcal{H}$, $\frac1S \sum_{j =1}^S \Vsi[i,j] h_j  = \frac1S \sum_{j =1}^S \fVsi (x_j)  h (x_j) \stackrel{S \to \infty}\to \left\langle \fVsi, h\right\rangle_{L^2}$.
\end{Rk}

\todoAS[inline]{After the above remark or inside of it, should we mention a connection between $\eTMN$ and  $\TMN$?}

Proof of \cref{Prop:RCE-as-N-components-projection} can be found in \cref{app:proof-RCE-as-N-components-projection}.
From \cref{Prop:RCE-as-N-components-projection} RCE is related to the concept of \emph{expressivity bottleneck} illustrated in \cite{verbockhaven2024growing}, and measures what part of the learning signal is not captured by truncating at $N$ components for natural gradient computation.
Therefore, this metric allows us to explicitly estimate and compare different cutoff choices. Note that this metric incurs no additional computational cost since ANaGRAM already computes the required SVD.
\subsection{Empirical Observations: Flattening}\label{subsec:empirical-observations}

\begin{figure}[H]
  \centering
  \includegraphics[width=0.8\linewidth]{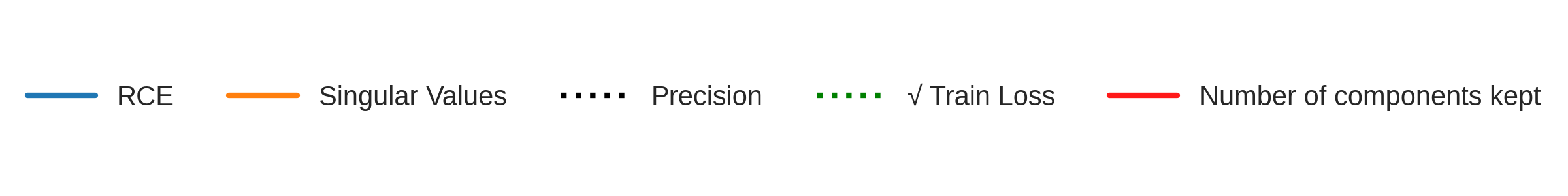}\\[0.8em]

  \begin{subfigure}[t]{.48\textwidth}
    \centering
    \includegraphics[width=\linewidth]{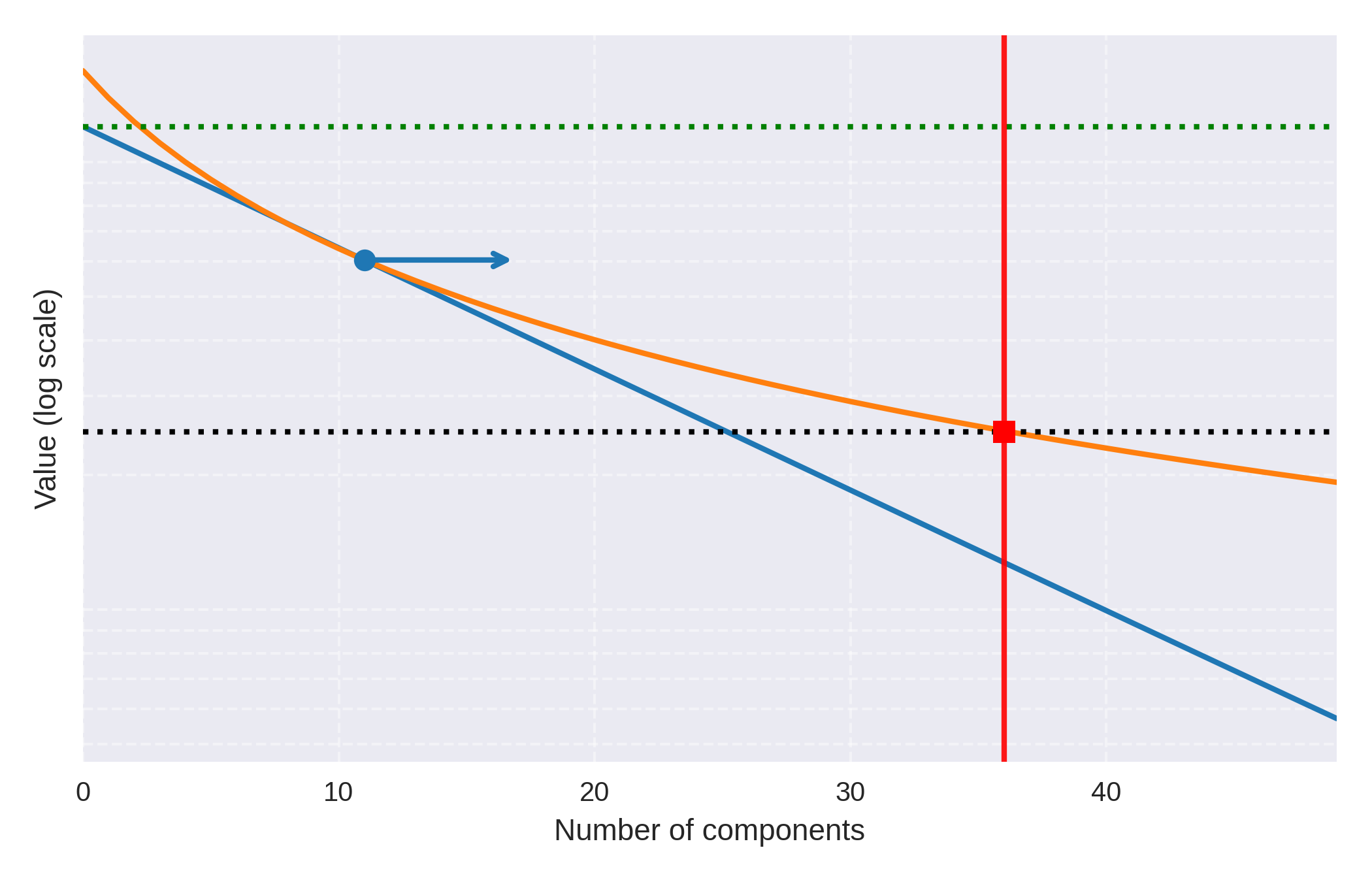}
    \caption{Early iterations, RCE at intersection with singular values is above the desired precision threshold.}
    \label{fig:anagram_first_iterations}
  \end{subfigure}\hfill
  \begin{subfigure}[t]{.48\textwidth}
    \centering
    \includegraphics[width=\linewidth]{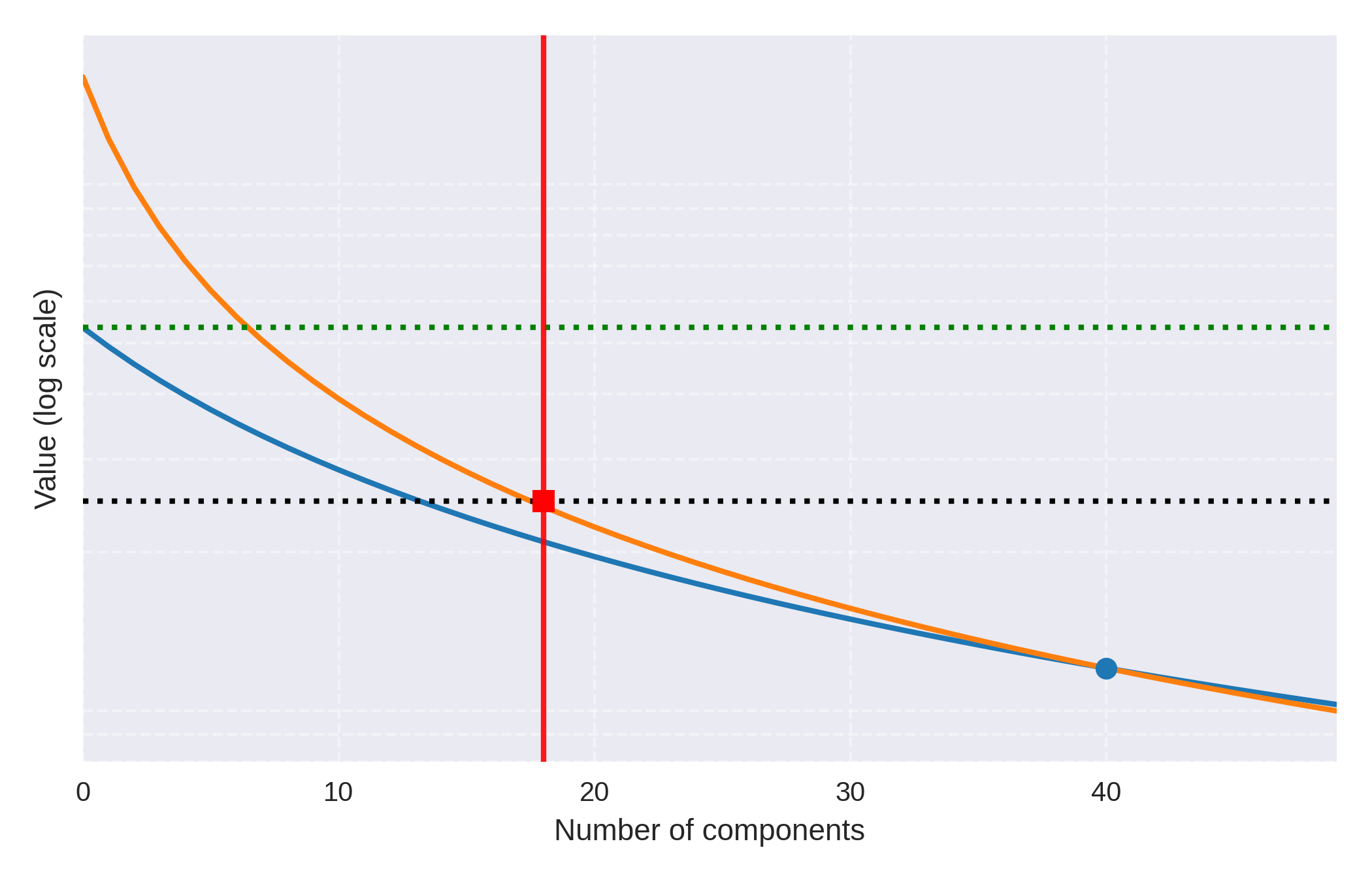}
    \caption{The RCE and singular values intersection drops below precision.}
    \label{fig:anagram_cross_precision}
  \end{subfigure}

  \vspace{0.75em}

  \begin{subfigure}[t]{.48\textwidth}
    \centering
    \includegraphics[width=\linewidth]{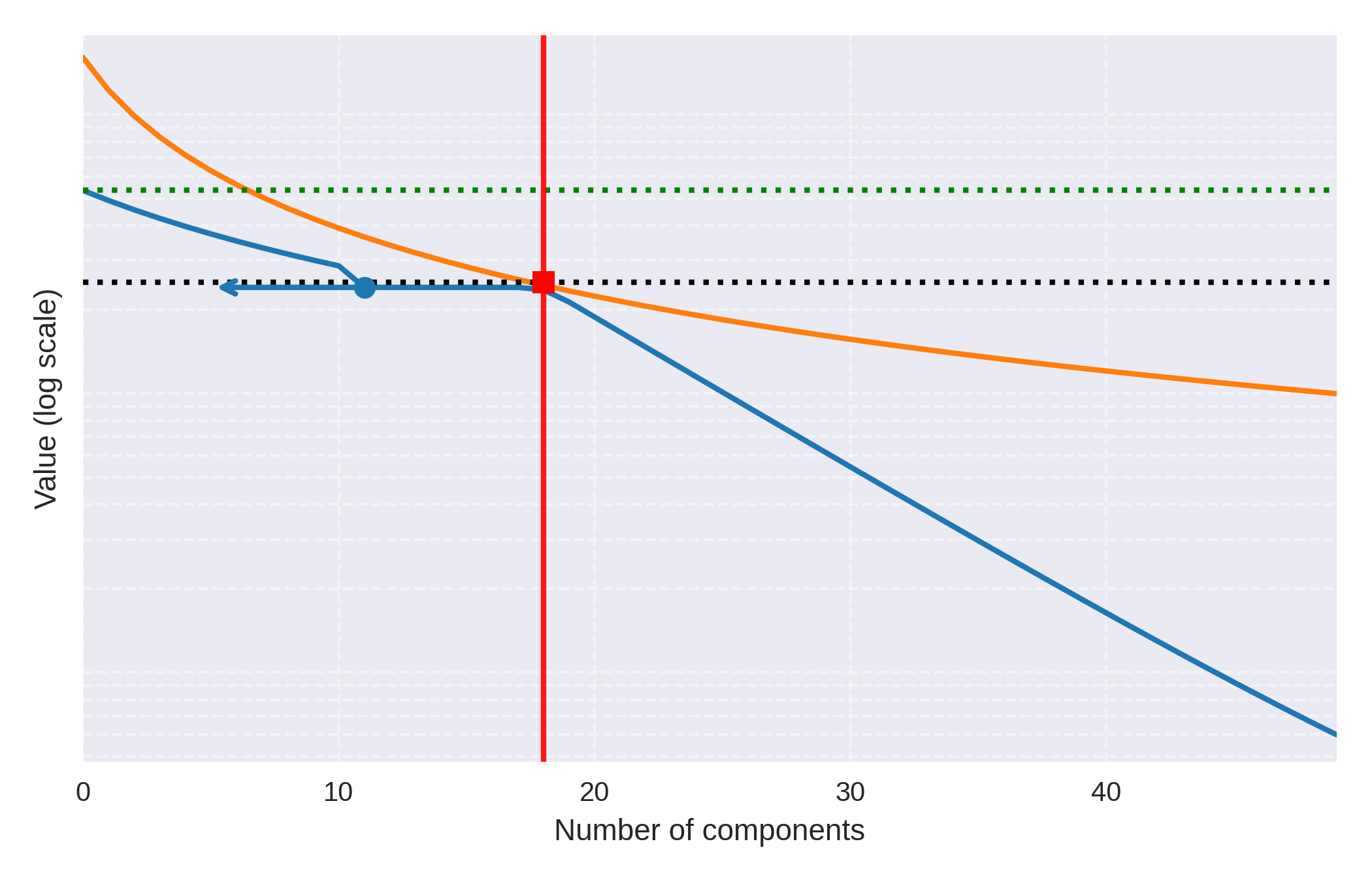}
    \caption{Beginning of the flattening: a plateau of RCE starts from $\rkk$ and propagates toward zero.}
    \label{fig:anagram_flattening}
  \end{subfigure}\hfill
  \begin{subfigure}[t]{.48\textwidth}
    \centering
    \includegraphics[width=\linewidth]{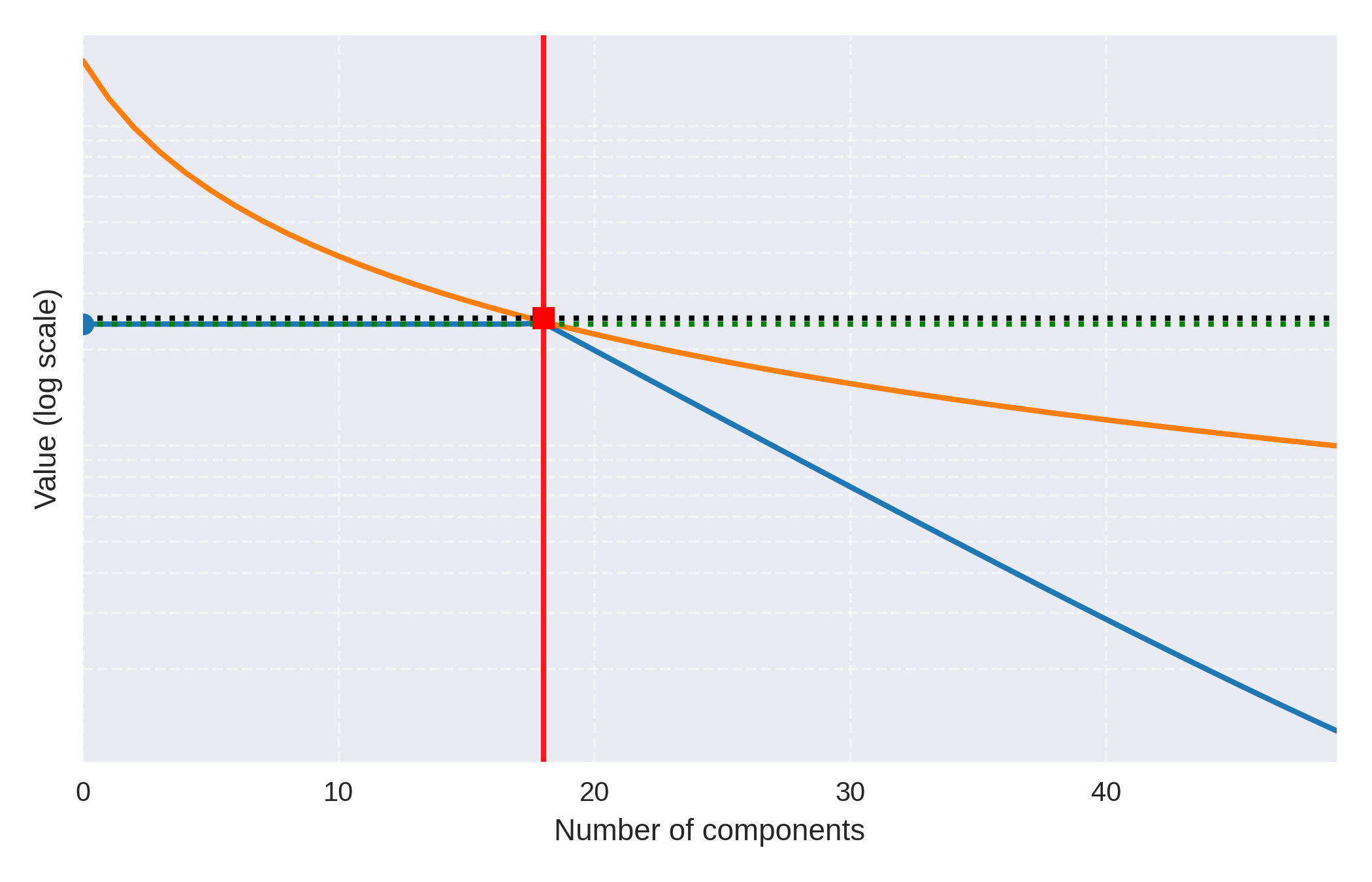}
    \caption{Final stage: full flattening and convergence.}
    \label{fig:anagram_final}
  \end{subfigure}

  \caption{\textbf{ANaGRAM training dynamics.} Legend (top) and four key phases: (a) initial evolution, (b) reconstruction–singular value intersection passes  target precision, (c) emergence of the flattening regime, (d) complete flattening yielding final loss level. Despite changing scale, target precision is constant and fixed across all plots. The number of ANaGRAM's retained components $\rkk$ is at intersection of precision line with singular values curve.}
  \label{fig:anagram_training_evolution_main}
\end{figure}

Here we illustrate the evolution of training loss and reconstruction error, where \cref{fig:anagram_training_evolution_main} schematically outlines key stages of ANaGRAM's training dynamics. The plot of a real experiment is provided in \cref{app:anagram_training_dynamics}.

Let $\Ccutoff$ is a cutoff level (also referred to as precision) and $\rkk$ denote the number of components retained by the cutoff, i.e., $\rkk(t) = \max\{j : \Dsi[j] \geq \Ccutoff\}$.
In \cref{fig:anagram_training_evolution_main}, we observe different stages of the training.
First, the reconstruction error is above the wanted precision (\cref{fig:anagram_iter_0}). As the training progresses, the training loss drops and the reconstruction error drops until reaching the cutoff precision (\cref{fig:anagram_iter_40}). Eventually, the reconstruction error drops below the cutoff threshold (\cref{fig:anagram_iter_90}). During this phase, the training loss (corresponding to the RCE for 0 component (green line in the figure)) is not decreasing a lot.

Then, a phenomenon that we call "flattening" occurs: once the reconstruction error is small compared to the cutoff precision value, reconstruction error
\textit{flattens} over the interval $[\Nflat, \rkk]$, where $\Nflat$ is the smallest number such as
\begin{equation}\label{eqn:informal-flattening-statement}
  \rce^S_{\Nflat} - \rce^S_{\rkk} \approx 0.
\end{equation}
Eventually, the phenomenon propagates toward low numbers of retained components (\cref{fig:anagram_iter_150}) and $\Nflat = 0$. Reconstruction error is now constant for all retained components and the training ends with training loss at cutoff precision. We refer a reader to \cref{app:proof-RCE-as-N-components-projection} to have a more theoretical insight on what is happening during the flattening.
\todoAS{Add some comments in the correspodingin appendix}

\begin{Rk}\label{Rk:RCE0-is-loss}
  This phenomenon sheds light on the sharp drop in training loss observed near the end of optimization, as reported in \cite{schwencke2025anagram}. By combining \cref{eqn:empirical-quadratic-loss,eqn:reconstruction-loss-orthogonal,eqn:eng-computation-trick} and using that $\Pi^0_0 = 0$, we obtain
  \begin{equation}
    {\rce^S_0}^2
    \stackrel{\ref{eqn:reconstruction-loss-orthogonal}}{=}
    \frac{1}{S} \norm[\Vsing \Pi^0_0 \Vsing^t \ELgrad - \ELgrad ]^2_{\RR^S}
    =
    \frac{1}{S} \norm[\ELgrad]^2_{\RR^S}
    \stackrel{\ref{eqn:eng-computation-trick}}{=}
    \frac{1}{S} \sum_{i=1}^S \left(\utheta(x_i) - f(x_i)\right)^2
    \stackrel{\ref{eqn:empirical-quadratic-loss}}{=} \ell(\vtheta).
  \end{equation}
  Thus, the last iteration of flattening is \textbf{directly responsible for the sudden drop of train loss} at the end of the training.
\end{Rk}

\begin{Rk}
  We see that for higher precision than the cutoff value ($N > \rkk$), the RCE is still decreasing as we increase the number of components kept. This indicates that there is still information to capture in the functional eigenspace composed of components associated to lower eigenvalues, see also \cref{app:proof-RCE-as-N-components-projection}.
\end{Rk}

The final interesting observation is that
\begin{equation}
  \rce^S_0 - \rce^S_{\rkk} \simeq 0 \qquad \Leftrightarrow \qquad \Pi^0_{\rkk} \Vsing^T \ELgrad \approx 0.
\end{equation}
Thus, the flattening phenomenon means that the projection of the signal onto the first $\rkk$ components retained by the cutoff is negligible. In other words, the optimization has extracted all the \emph{usable} signal from these components at this cutoff level.

\subsection{Incomplete Flattening and Adaptive Strategies}\label{subsec:incomplete-and-instant-flattening}

In practice, for some experiments we observe that the flattening may remain incomplete with $\lim_{t \to \infty} \Nflat = \NflatInf> 0$: the system remains in a state similar to that shown in \cref{fig:anagram_flattening} and never (at least not within a reasonable number of iterations) reaches the configuration illustrated in \cref{fig:anagram_final}.
A natural question arises: \emph{what happens if we adjust the cutoff to retain exactly $\NflatInf$ components?}

If we try this trick in practice (see \cref{fig:adapted-cutoff-flattening}), then a single natural gradient step with an adjusted cutoff can be enough to get immediate and complete flattening ($\Nflat = 0$) and eventually dramatically reduce training loss.
This abrupt flattening when restricting cutoff to low number of feature is typically accompanied by a learning rate found by the line search to be very close to one. A possible explanation is that this may represent an iteration in the \emph{lazy training} regime (NTK and the feature matrix are nearly constant), where we regress linearly (and thus fast) based on learned features. This hypothesis should be further explored in future work.

This empirical insight motivates the use of an adaptive algorithm: by dynamically adjusting cutoffs, we can hope to accelerate convergence and achieve higher precision.

\section{Algorithmic Design: Exploiting Flattening}\label{sec:algorithmic-design}

Building upon the empirical analysis presented in \cref{sec:anagram-insights}, we develop a principled algorithm that controls and exploits the flattening phenomenon identified in ANaGRAM's training dynamics. Our approach is based on tracking the relationship between reconstruction error and singular values to automatically determine well-adapted cutoff in order to reach the target precision (error) $\ePr$ at the end of the training. This well-adapted cutoff should vary from one iteration to another to adjust to the currently learned weights and training dynamics in such a way to avoid early flattening (if flattening happens too early, the training stagnates at higher values of losses) and when intersection between RCE and singular values goes below the target precision $\ePr$, we enforce the flattening, so that the final training loss also drops to $\ePr$.

\subsection{Adaptive Cutoff Strategy}
In what follows, we 
suggest an adaptive cutoff rank $\rkk$ that indicates how much components of $\Dsing$ are retained for the next update of ANaGRAM.
Our algorithm operates by dynamically selecting cutoff ranks based on the relationship between reconstruction error and singular values:
\begin{equation}\label{eqn:adaptive-cutoff-rule}
  \rkk(t) = \begin{cases}
    \rkint(t)  := \max\left\{j : \rce^S_j(t) \leq \Dsi[j][t] \right\} & \text{if } \rce^S_{\rkint(t)}(t) > \epsilon \text{ (intersection rank)}, \\
    \rkeps(t) := \max\left\{j : \rce^S_j(t) \geq \epsilon\right\}     & \text{if } \rce^S_{\rkint(t)}(t) \leq \epsilon \text{ (precision rank)}.
  \end{cases}
\end{equation}

The algorithm terminates when $\rkeps(t) = 0$, indicating that the reconstruction error $\rce^S_0$ that is equal to the training error is indeed below the predefined precision threshold.

For ease of presentation, we provide only the core elements of \method in Algorithm~\ref{alg:principled-adaptive} consisting in adaptively choosing, which $\rkk$ to apply for $\Dsing$ at each update of ANaGRAM. The final algorithm is explained in \cref{sec:practical-implementation}. Final Algortihm~\ref{alg:anagram} addresses some irregularities observed in evolution of RCE and singular values that we explain in more details in \cref{app:practical-justifications}.

\SetKwProg{DefFun}{Function}{}{end}
\SetKwFunction{SVD}{SVD}
\SetKwFunction{RCE}{ReconstructionErrors}
\SetKwFunction{FindIntersection}{FindIntersection}
\SetKwFunction{FE}{FindElbow}
\SetKwFunction{CountPrecision}{CountPrecisionComponents}
\SetKwFunction{ANaGRAMStep}{AnagramStep}

\begin{figure}[H]
  \centering
  \captionsetup[subfigure]{width=.95\linewidth}
  \begin{minipage}{\textwidth}
    \centering
    \includegraphics[width=\linewidth]{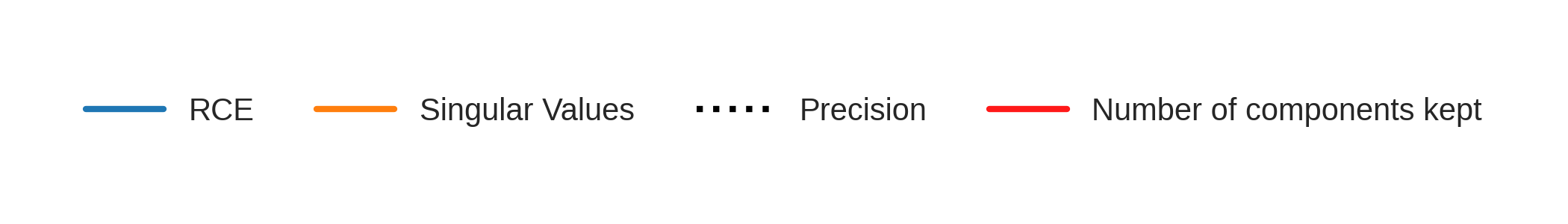}
  \end{minipage}

  \begin{subfigure}[t]{.48\textwidth}
    \centering
    \includegraphics[width=\linewidth]{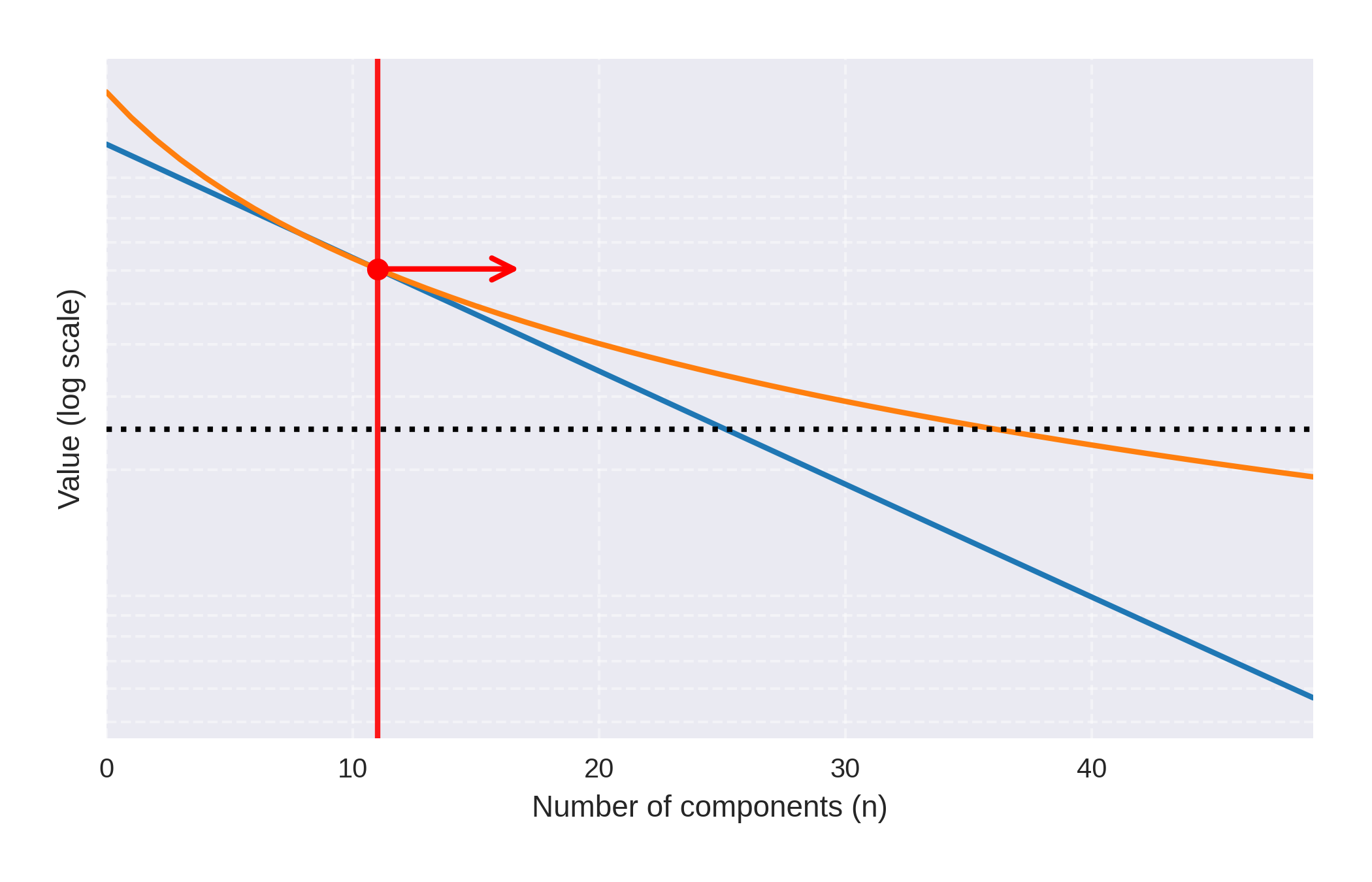}
    \caption{Early iterations ($\rkk = \rkint$).}
    \label{fig:adaptive_dynamics_early}
  \end{subfigure}%
  \hfill
  \begin{subfigure}[t]{.48\textwidth}
    \centering
    \includegraphics[width=\linewidth]{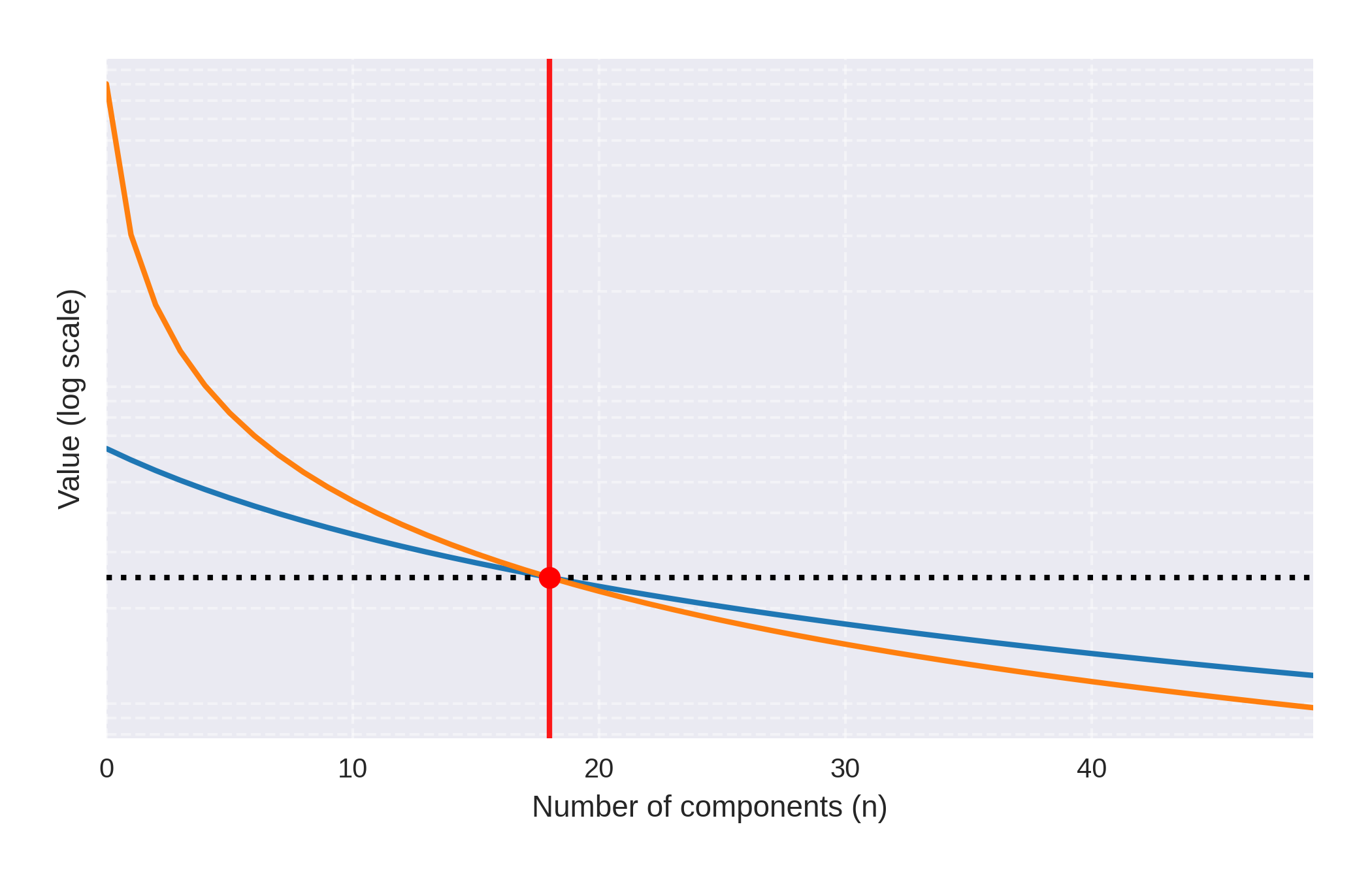}
    \caption{Intersection at precision ($\rkk = \rkint = \rkeps$) triggers a switch between different cutoff strategies.}
    \label{fig:adaptive_dynamics_intersection}
  \end{subfigure}


  \begin{subfigure}[t]{.48\textwidth}
    \centering
    \includegraphics[width=\linewidth]{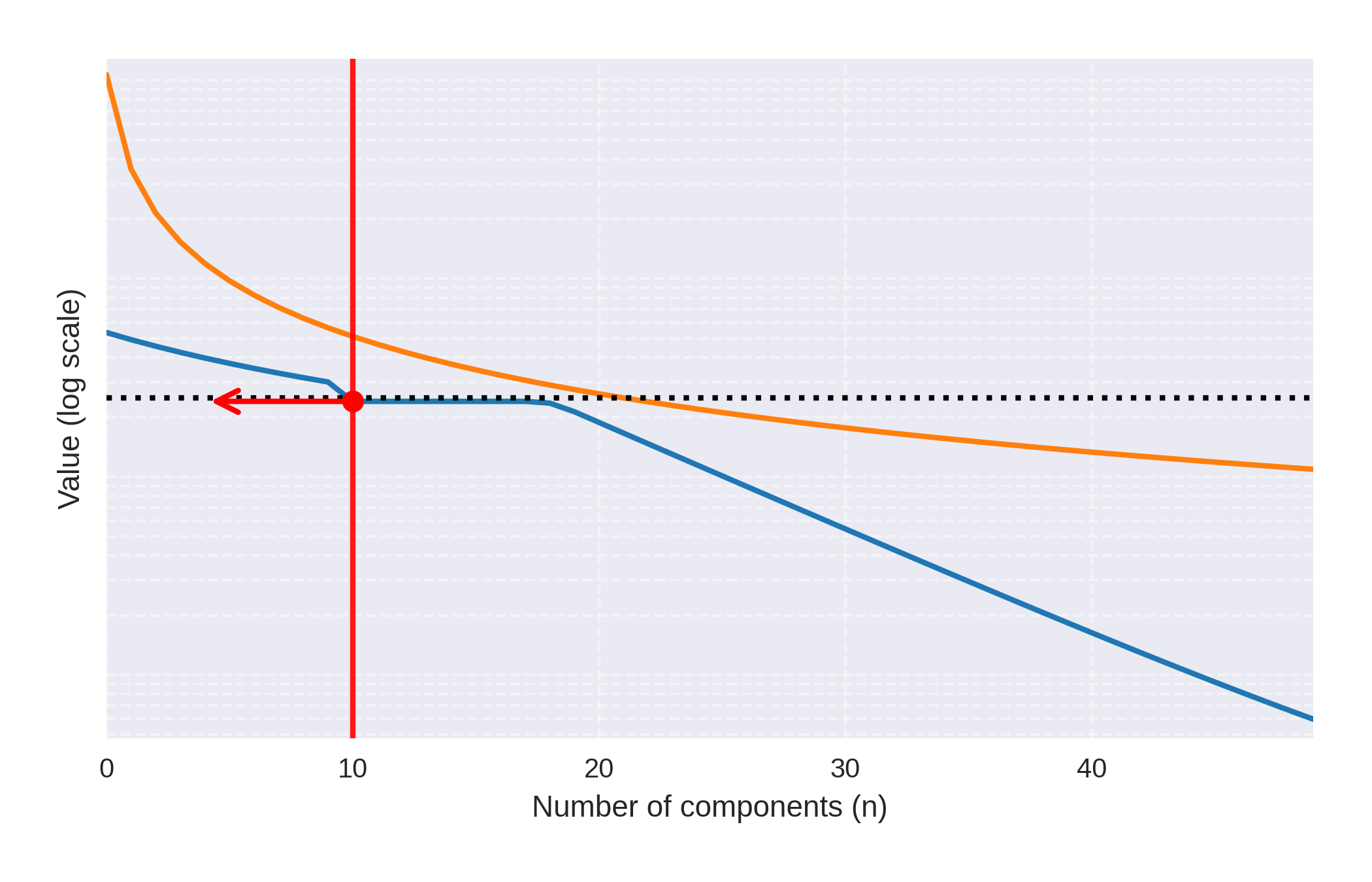}
    \caption{Flattening: error plateaus across retained components $\rkk = \rkeps$.}
    \label{fig:adaptive_dynamics_flattening}
  \end{subfigure}%
  \hfill
  \begin{subfigure}[t]{.48\textwidth}
    \centering
    \includegraphics[width=\linewidth]{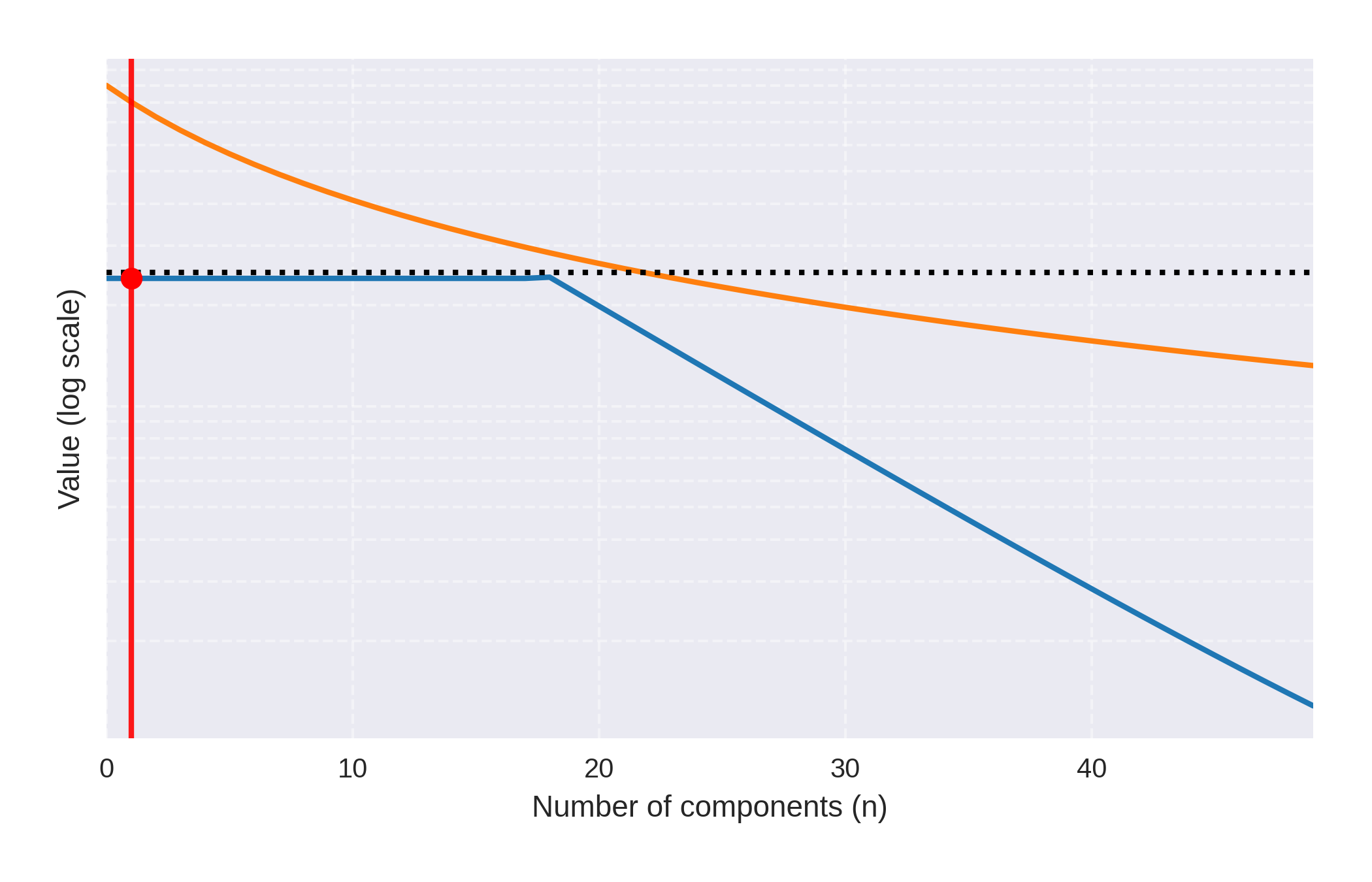}
    \caption{Final iteration: full flattening and convergence.}
    \label{fig:adaptive_dynamics_final}
  \end{subfigure}

  \caption{\textbf{Dynamics of the adaptive multi-cutoff strategy in AMStraMGRAM.}
    Progression from (a) initial exploration, (b) intersection reaches precision, (c) flattening onset, to (d) converged state. Red arrows (when present) indicate the retained rank dynamics (pointing right -- increasing, pointing left -- decreasing). Legends are shown below.}
  \label{fig:adaptive_strategy_dynamics}
\end{figure}

\todoAS[inline]{Move into Discussion after experiments}
\subsection{Geometrical Interpretation of the Adaptive Strategy}

The algorithm exploits the geometric relationship between the empirical tangent space and the functional gradient.  By tracking the intersection, we maximize the projection of the functional gradient onto the empirical tangent space while staying out of flattening. Once the intersection reach the precision level, we exploit the flattening phenomenon to achieve prescribed precision.

According to \cref{Prop:RCE-as-N-components-projection}, the reconstruction error $\rce^S_N$ measures how much of the functional gradient signal remains to be captured by the first $N$ components. The intersection point thus represents the good balance between signal capture and phase transition.

\section{Experimental Results}

We first compare in \cref{tab:comparison} our method implemented in JAX\footnote{\url{https://anonymous.4open.science/r/AMStraMGRAM-8D1B/}} with the ANaGRAM method \citep{schwencke2025anagram} on the benchmark problems presented in their paper, with modified datasets. As we see, for every equation, we perfom better.

\begin{table}[H]
  \centering
  \caption{Performance comparison between AMStraMGRAM (our method) and ANaGRAM \cite{schwencke2025anagram}. The adaptive strategy demonstrates significant improvements across all benchmark problems, with L2 error improvements of up to 8 orders of magnitude.}
  \resizebox{\columnwidth}{!}{%
    \begin{tabular}{lcccc}
      \toprule
      {Experiment}  & \multicolumn{2}{c}{Mean Squared Error (MSE)} & \multicolumn{2}{c}{$L_2$ Error}                                                              \\
      \cmidrule(lr){2-3} \cmidrule(lr){4-5}
                    & Ours                                         & ANaGRAM                         & Ours                             & ANaGRAM                 \\
      \midrule
      Heat Equation & \textbf{6.29e-29 $\pm$ 6.78e-30}             & 8.56e-11 $\pm$ 7.05e-11         & \textbf{2.32e-14 $\pm$ 1.14e-14} & 1.28e-06 $\pm$ 1.75e-06 \\
      Laplace 2D    & \textbf{1.46e-28 $\pm$ 1.87e-29}             & 4.27e-13 $\pm$ 4.66e-13         & \textbf{2.24e-15 $\pm$ 2.52e-16} & 3.49e-09 $\pm$ 3.58e-09 \\
      Laplace 5D    & \textbf{2.04e-08 $\pm$ 1.16e-08}             & 6.37e-08 $\pm$ 7.01e-08         & \textbf{2.12e-05 $\pm$ 8.15e-06} & 4.00e-05 $\pm$ 2.93e-05 \\
      Allen--Cahn   & \textbf{3.19e-11 $\pm$ 2.37e-11}             & 2.19e-04 $\pm$ 4.16e-04         & \textbf{5.87e-05 $\pm$ 6.25e-06} & 4.32e-03 $\pm$ 5.93e-03 \\
      \bottomrule
    \end{tabular}%
  }
  \label{tab:comparison}
\end{table}

We then compare our method with the baseline methods from \cite{urbanUnveilingOptimizationProcess2025} on the benchmark problems presented in their paper.
Note that in our case we do not need to enforce boundary constraints.
The methodology of sampling is also sighltly different, as we sample the data from a fixed grid, following the methodology of \cite{schwencke2025anagram}, while in \cite{urbanUnveilingOptimizationProcess2025} they perform batching of randomly sampled points.

\begin{table}[H]
  \centering
  \caption{Performance comparison between AMStraMGRAM (our method) and baseline \cite{urbanUnveilingOptimizationProcess2025} methods. Our method demonstrates improvements across benchmark problems, without requiring enforcement of boundary constraints.}
  \resizebox{\columnwidth}{!}{%
    \begin{tabular}{lcccc}
      \toprule
      {Experiment}                  & \multicolumn{2}{c}{Mean Squared Error (MSE)} & \multicolumn{2}{c}{$L_2$ Error}                                                                \\
      \cmidrule(lr){2-3} \cmidrule(lr){4-5}
                                    & Ours                                         & SSBroyden*                       & Ours                    & SSBroyden*                        \\
      \midrule
      One-dimensional Burgers (1DB) & \textbf{2.99e-12 $\pm$ 9.26e-13}             & 2.92e-10 $\pm$ 1.45e-10
                                    & \textbf{1.5e-06 $\pm$ 9.43e-7}               & 1.59e-06   $\pm$ 1.02e-6                                                                       \\
      Non-Linear Poisson (k=1)      & \textbf{8.51e-24 $\pm$ 2.24e-24}             & 3.03e-16   $\pm$ 3.82e-16        & 6.81e-10 $\pm$ 1.41e-09 & \textbf{9.29e-12  $\pm$ 5.85e-12} \\
      Allen--Cahn (AC)              & 3.19e-11 $\pm$ 2.37e-11                      & \textbf{6.42e-12 $\pm$ 5.52e-12} & 5.87e-05 $\pm$ 6.25e-06 & \textbf{3.94e-06 $\pm$ 1.72e-06}  \\
      \bottomrule
    \end{tabular}%
  }
  \footnotesize{* refer to method from \cite{urbanUnveilingOptimizationProcess2025} with adaptive sampling and hard constraint enforcement on boundary conditions.}

  \label{tab:comparison_with_urban}
\end{table}

\section{Limitations}
\begin{figure}[H]
  \centering
  \includegraphics[width=0.9\linewidth]{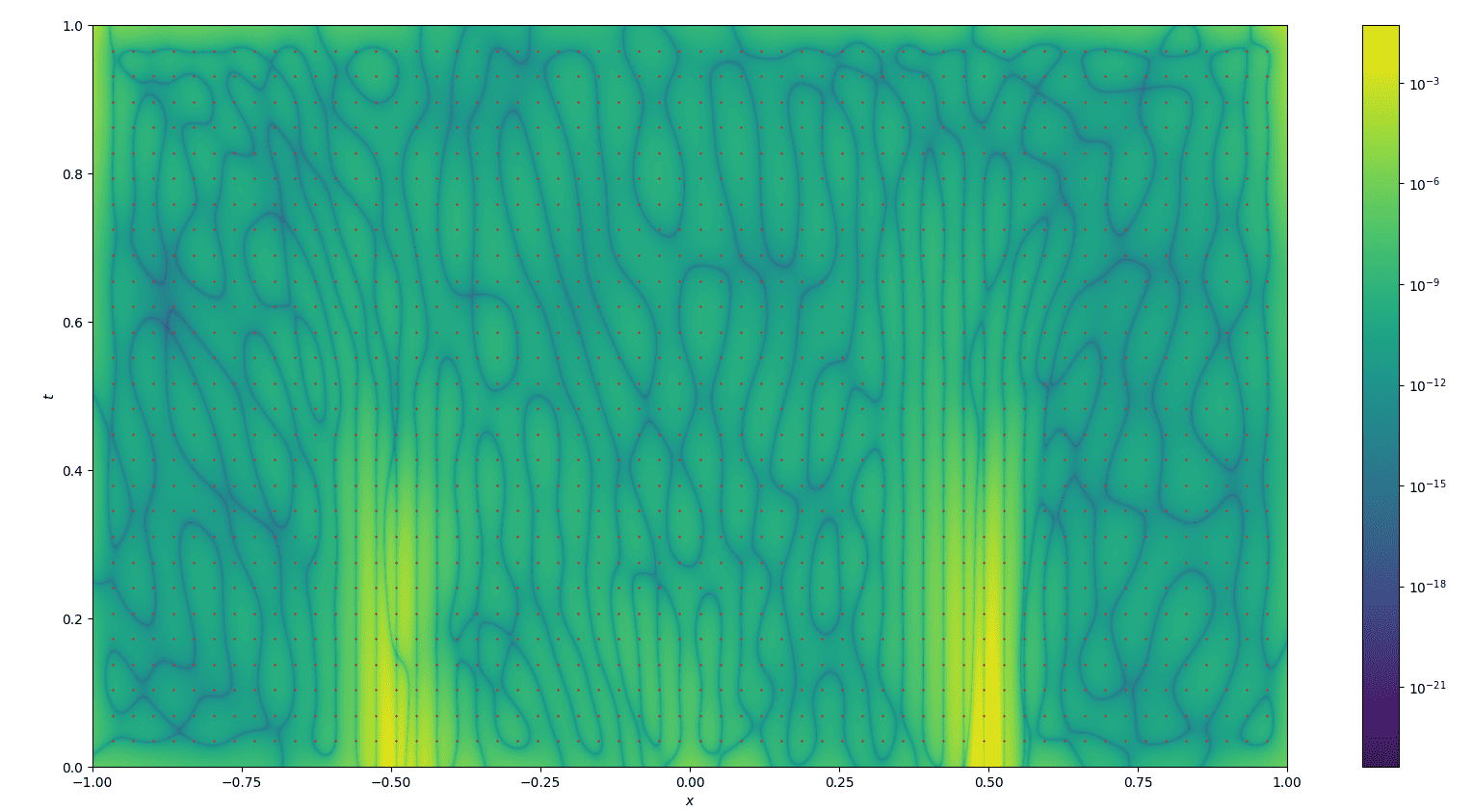}
  \caption{Allen–Cahn overfitting: residual lines align with sampling lines. Low-rank (post-cutoff) tangent projections fit exactly on sampled fibers while interpolation between them inherits weakly constrained oscillations in regions of steep interface curvature.}
  \label{fig:allencan}
\end{figure}

Despite its effectiveness, AMStraMGRAM can exhibit overfitting, particularly in problems with sharp features like the Allen--Cahn equation. The algorithm drives the training error to machine precision on the sampled points, but the learned function may develop high-frequency oscillations between them, especially in regions of high curvature where the approximation is the most challenging. These artifacts, visible as ``overfitting lines'' in \cref{fig:allencan}, are an imprint of the sampling lattice (see regions around $x=\pm0.5$). They arise because the SVD cutoff effectively projects the update onto a low-rank subspace of the tangent space. This subspace is often aligned with the grid axes, leading to anisotropic smoothing that perfectly fits the data on the grid lines but interpolates poorly in the under-sampled regions between them. Once the flattening phase begins, the training enters a quasi-linear regime that can ``lock in'' these geometric artifacts.

This phenomenon highlights that while our method significantly improves on ANaGRAM, the quality of the final solution remains fundamentally limited by the sampling strategy. Mitigating such overfitting requires co-designing the sampler and the optimizer. Potential remedies include adaptive sampling, where new collocation points are added in regions of high reconstruction error, or curriculum-based approaches that progressively refine the sampling grid.

\section{Conclusion}

In this work, we have introduced AMStraMGRAM, an adaptive multi-cutoff strategy that enhances the ANaGRAM natural gradient method for training PINNs. Our work provides an analytical framework to explain ANaGRAM’s convergence behavior, uncovering a \emph{flattening} phenomenon that clarifies its training dynamics. The proposed algorithm automatically adjusts cutoff regularization. Notably, AMStraMGRAM exhibits ``overfitting'' as demonstrated in Allen-Cahn experiments. These results underscore the potential of natural gradient optimization for PINNs while highlighting the critical role of sampling strategies in realizing their full accuracy.

Future research will focus on integrating residual-based methods to further stabilize training, establishing rigorous convergence guarantees for our adaptive cutoff scheme, and extending the approach to higher-dimensional PDEs and complex geometries. Exploring the interplay between network architecture and optimization—as well as further developing sampling techniques—will be essential to address the fundamental challenge of balancing optimization power with data representation. Ultimately, our findings suggest that with careful algorithmic design, PINNs can achieve the precision required for practical scientific computing, paving the way for mesh-free methods in computational science.

\newpage
\bibliographystyle{iclr2026/iclr2026_conference}
\bibliography{amstramgram}

\newpage

\appendix

\section{Illustration of Natural Gradient}

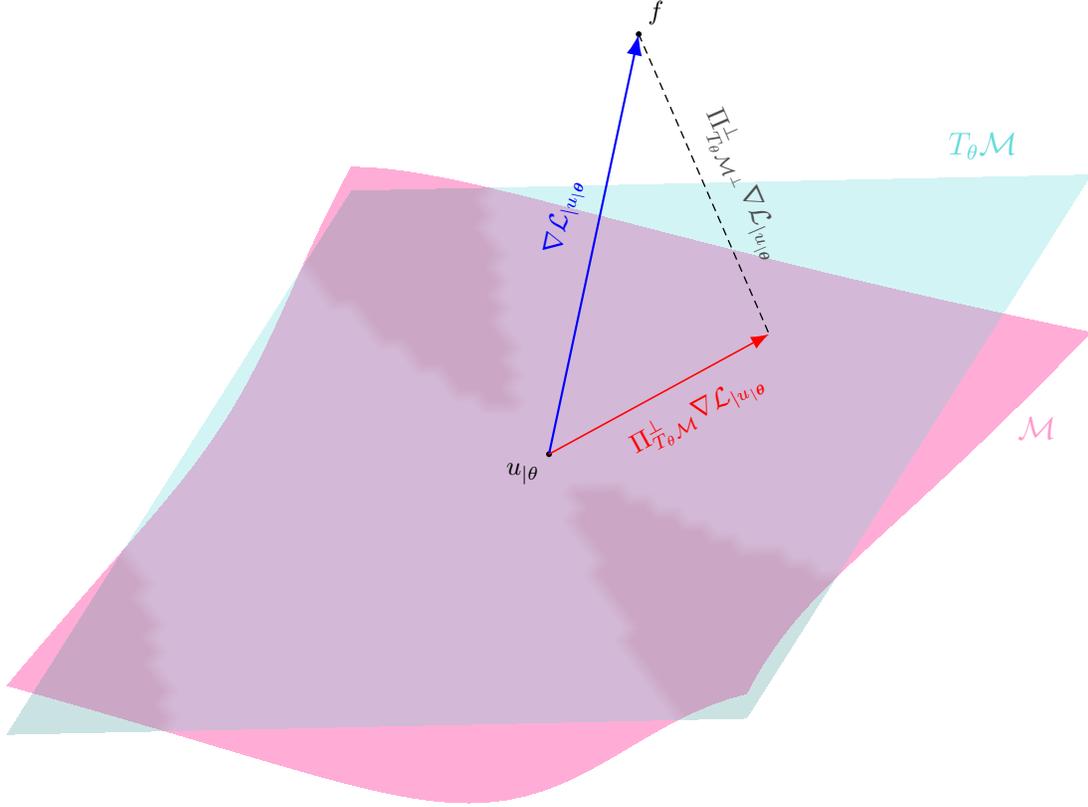
\begin{figure}[H]
  \centering
  \tikzsetnextfilename{NNmanifold}
\def\freq{5}
\newcommand{\surf}{tanh(x+y)/(1+sin(deg((x-y)/\freq))^2)}
\newcommand{\surfVar}[2]{tanh(#1+#2)/(1+sin(deg((#1-#2)/\freq))^2)}
\pgfmathsetmacro{\thetax}{-2.56}
\pgfmathsetmacro{\thetay}{3.47}
\pgfmathsetmacro{\thetaz}{0.39}

\def\xmin{\thetax-2.5} \def\xmax{\thetax+2.5}
\def\ymin{\thetay-2.5} \def\ymax{\thetay+2.5}

\pgfmathsetmacro{\tanhtheta}{tanh(\thetax+\thetay)}                     
\pgfmathsetmacro{\Ader}{1 - \tanhtheta*\tanhtheta}                      
\pgfmathsetmacro{\sintheta}{sin(deg((\thetax-\thetay)/\freq))}          
\pgfmathsetmacro{\costheta}{cos(deg((\thetax-\thetay)/\freq))}          
\pgfmathsetmacro{\B}{1 + \sintheta*\sintheta}                           
\pgfmathsetmacro{\Bx}{2*\sintheta*\costheta/\freq}                      
\pgfmathsetmacro{\By}{-2*\sintheta*\costheta/\freq}                     

\pgfmathsetmacro{\pp}{(\Ader*\B - \tanhtheta*\Bx)/(\B*\B)}              
\pgfmathsetmacro{\qq}{(\Ader*\B - \tanhtheta*\By)/(\B*\B)}
\def\p{0.28}
\def\q{0.23}

\pgfmathsetmacro{\gx}{0.35}
\pgfmathsetmacro{\gy}{0.55}
\pgfmathsetmacro{\gz}{2.8}
\pgfmathsetmacro{\elev}{25}   
\pgfmathsetmacro{\azim}{50}   

\pgfmathsetmacro{\nx}{-1*\p}
\pgfmathsetmacro{\ny}{-1*\q}
\pgfmathsetmacro{\nz}{1}
\pgfmathsetmacro{\ndotn}{\nx*\nx + \ny*\ny + \nz*\nz}
\pgfmathsetmacro{\gdotn}{\gx*\nx + \gy*\ny + \gz*\nz}
\pgfmathsetmacro{\vnx}{\gdotn/\ndotn * \nx}
\pgfmathsetmacro{\vny}{\gdotn/\ndotn * \ny}
\pgfmathsetmacro{\vnz}{\gdotn/\ndotn * \nz}
\pgfmathsetmacro{\vtx}{\gx - \vnx}
\pgfmathsetmacro{\vty}{\gy - \vny}
\pgfmathsetmacro{\vtz}{\gz - \vnz}

\pgfplotsset{
  colormap={cyanmap}{
    rgb255(0cm)=(50,110,110)
    rgb255(1cm)=(99,219,219)
  }
}
\pgfplotsset{
  colormap={magentaonly}{
    rgb255(0cm)=(255,153,204)
    rgb255(1cm)=(255,153,204)
  }
}

\begin{tikzpicture}
\begin{axis}[
  hide axis,
  view={\elev}{\azim},
  hide axis,
  axis lines=none,
  enlargelimits=false,
  width=16cm,  
  z buffer=sort,
  clip=false,
  xmin=\xmin, xmax=\xmax, ymin=\ymin, ymax=\ymax
]

\addplot3[
  surf, opacity=.8, fill=magenta!70,
  colormap name=magentaonly,
  draw=none, domain=\xmin:\xmax, y domain=\ymin:\ymax,
  mesh/ordering=y varies,
  samples=41, samples y=41,
  shader=interp
] ({x},{y},{\surf});
\node[font=\large\itshape, text={rgb,255:red,255;green,153;blue,204}] at (axis cs:{\xmax-0.25},{\ymax-0.25},{\surfVar{\xmax-0.25}{\ymax-0.25}-1}) {$\mathcal{M}$};

\addplot3[
  surf, draw=none,
  opacity=0.28,
  colormap name=cyanmap,
  domain=\xmin:\xmax, y domain=\ymin:\ymax,
  samples=41, samples y=41,
  point meta={ (\thetaz + \p*(x-\thetax) + \q*(y-\thetay) >
                \surf )
                ? 1  
                : 0.00  
              },
  mesh/ordering=y varies,
  shader=interp,  
] ({x},{y},{\thetaz + \p*(x-\thetax) + \q*(y-\thetay)});
\node[font=\large\itshape, text={rgb,255:red,99;green,219;blue,219}] at (axis cs:\xmax-.5,\ymax-.5,{\thetaz + .6 + \p*(\xmax-.5-\thetax) + \q*(\ymax-.5-\thetay)}) {$T_{\theta}\mathcal M$};

\fill (axis cs:\thetax,\thetay,\thetaz) circle[radius=1.1pt];
\node[anchor=north east] at (axis cs:\thetax,\thetay,\thetaz) {$u_{|\theta}$};

\fill (axis cs:\thetax+\gx,\thetay+\gy,\thetaz+\gz) circle[radius=1.1pt];
\node[anchor=south west] at (axis cs:\thetax+\gx,\thetay+\gy,\thetaz+\gz) {$f$};

\draw[-{Latex[length=2.8mm]}, line width=0.8pt,color=blue]
  (axis cs:\thetax,\thetay,\thetaz) -- (axis cs:\thetax+\gx,\thetay+\gy,\thetaz+\gz)
  node[pos=0.55, sloped, text=blue, above=0.01\textwidth] {$\nabla \mathcal L_{|u_{|\vtheta}}$};

\draw[-{Latex[length=2.4mm]}, line width=0.6pt,color=red]
  (axis cs:\thetax,\thetay,\thetaz) -- (axis cs:\thetax+\vtx,\thetay+\vty,\thetaz+\vtz)
  node[pos=0.6, sloped, below=0.01\textwidth, text=red] {$\Pi^\bot_{T_\theta\mathcal{M}}\nabla\mathcal L_{|u_{|\vtheta}}$};

\draw[densely dashed, line width=0.5pt]
  (axis cs:\thetax+\gx,\thetay+\gy,\thetaz+\gz) -- (axis cs:\thetax+\vtx,\thetay+\vty,\thetaz+\vtz)
  node[pos=0.55, sloped, above=0.01\textwidth, text=black!70] {$\Pi^\bot_{T_\theta\mathcal{M}^{\perp}}\nabla\mathcal{L}_{|u_{|\vtheta}}$};

\end{axis}
\end{tikzpicture}
  \caption{Illustration of the orthogonal projection of the functional gradient onto the tangent space. While the ideal update direction would be the functional gradient $\FLgrad_{|{\utheta}}$ (shown in blue), our model constrains us to follow directions within the tangent space $T_{\vtheta}\mathcal{M}$ (shown as a green plane). The optimal feasible direction is thus the orthogonal projection $\Pi^\perp_{T_{\vtheta}\mathcal{M}} \left( \FLgrad_{|{\utheta}} \right)$ (shown in red).}
  \label{fig:projection-illustration}
\end{figure}

\section{Our vocabulary}

\begin{itemize}
  \item \textbf{Domain ($\Omega$).}
  \item \textbf{Boundary ($\partial \Omega$).}
  \item \textbf{Differential operators ($D, B$).}
  \item \textbf{Cutoff ($\cutoff$).} A threshold below which the components of the matrix $\Dsing$ are truncated, \ie $\Dsing \leftarrow \begin{cases} \Dsing & \text{if } \Dsing \geq \cutoff,  \\ 0 & \text{else.} \end{cases}$
  \item \textbf{Full rank ($\FRk$).} A full rank of feature matrix $\feat$ that we assume, without loss of generality, to be equal to $\min(P,S)$.
  \item \textbf{Rank ($\rkk$).} A number of $\Dsing$ components that are retained when computing a pseudo-inverse of $\Dsing$ in ANaGRAM. Depending on a current regime of the training and a desired effect, it can be set at $\rkint$ or $\rkeps$.
  \item \textbf{Flattening.} The phenomenon described in \cref{subsec:empirical-observations}, when reconstruction error starts to stabilize for a range of possible ranks.
  \item \textbf{Flat cutoff ($\Nflat$).} A number of components that corresponds to the beginning of flattening in reconstruction error curve.
  \item \textbf{Feature matrix ($\feat \in \RR^{P\times S}$).} It is defined by a jacobian $\partial_{p} \utheta (x_i)$, which is used in an ANaGRAM's update to "project"  a functional gradient onto parameter space of $\vtheta$.
  \item \textbf{Precision ($\ePr$).}  A hyperparameter of \method that prescribes a target error level that the algorithm should achieve.
  \item \textbf{Intersection rank ($\rkint$).} Defined in \cref{eqn:adaptive-cutoff-rule}, roughly speaking it corresponds to a number of components at which reconstruction error and singular values curves are intersecting.
  \item \textbf{Precision rank ($\rkeps$).} Defined in \cref{eqn:adaptive-cutoff-rule}, it corresponds to a number of components at which reconstruction error curve and precision level are intersecting.
  \item \textbf{Functional gradient ($\FLgrad$).} A Frechet derivative of squared $L^2$ loss $\Floss$, its negative gives the "ideal" update direction in non-parametric case.
  \item \textbf{Empirical functional gradient ($\ELgrad \in \RR^{S}$).} A vector obtain by evaluating $\FLgrad$ on some finite number of samples $x_i \in \Omega$, for $i \in 1, \dots, S$.
  \item \textbf{Parametric model ($\utheta$).} A function parametrized with $\vtheta$ that serves to approximate a solution to a problem (regression or PDE). Typically, it is a neural network, where $\vtheta$ are its full set of weights.
  \item \textbf{Differential of the model ($d \, \utheta$).}  Defined as $d \, \utheta (h) = \sum_{p=1}^{P} h_p \frac{\partial u}{\partial \vtheta_p}=\lim\limits_{\varepsilon\to 0} \frac{u_{|\vtheta+\varepsilon h} - \utheta}{\varepsilon}$. It measures how much $\utheta$ changes in a given direction $h$.
  \item \textbf{Tangent space ($T_{\vtheta} \Mf$).} Image of a differential of the model, giving a space of possible updates for a model $\utheta$.
  \item \textbf{SVD components of $\feat$ ($\Using$, $\Dsing$, $\Vsing$).} In particular, $\feat = \Using \Dsing \Vsing^T$, where $\Using \in \RR^{P\times S}$ is a left singular vector matrix, $\Dsing \in \RR^{\FRk \times \FRk}$ is a diagonal matrix with singular values on a diagonal ordered in a decreasing order and $\Vsing$ is a right singular vector matrix.
  \item \textbf{Functional singular vectors ($\fVsi$).} Right singular vectors of the differential $d \utheta$.
  \item \textbf{Empirical tangent space ($\TMN$).} A subspace of tangent space $T_{\vtheta} \Mf$, restricted to a span of the right functional singular vectors $\fVsi$ corresponding to a range of components from $M$ to $N$, \ie $\Span(\fVsi \, : \, 1 \leq M\leq N \leq N)$.
  \item \textbf{Discretized empirical tangent space ($\eTMN$).}  A version of $\TMN$ discretized on a set of samples $\{x_i\}_{i=1}^S$ coming from $\Omega$.
  \item \textbf{Reconstruction error ($\rce^S_N$).} A measure identifying the portion of the functional gradient signal that is lost when restricting $\ELgrad$ to $\eTZN$.
  \item \textbf{Feature development phase.} The early phase in the training, during which high volatility is observed in both quantities of interest with high sensitivity to the choice of $\rkk$.
  \item \textbf{Flattening phase.} The later phase in the training, during which reconstruction error starts to flatten for some values of $N$, at the same time singular values dominate over reconstruction error for all retained components, resulting in a drop of training loss.
\end{itemize}

\todoAS[inline]{Draw a distinction between $N$ in $\rce$ and $\rkk$!!!!}

\todoAS[inline]{Add in the above vocabulary, all the defiinitions of the possible cutoffs/ranks}

\section{Practical Implementation Considerations}\label{sec:practical-implementation}

While the principled algorithm discussed in the main paper and summarized in Algorithm~\ref{alg:principled-adaptive} provides a sound framework, empirical observations reveal that additional mechanisms are necessary for robust performance across diverse PDE problems. This section describes additional modifications to make the algorithm more practical.

\subsection{The Dual Cutoff Strategy: Addressing Empirical Challenges}

Our experiments reveal that the single cutoff approach, while theoretically elegant, suffers from numerical instabilities and incomplete convergence in practice. We observed three critical issues:

\begin{enumerate}
  \item \textbf{Ignition failure:} The intersection between reconstruction error and singular values sometimes fails to evolve, preventing the algorithm from reaching lower error values.
  \item \textbf{Retreating dynamics:} The intersection rank may decrease during training, disrupting convergence.
  \item \textbf{Incomplete flattening:} Without additional stabilization, the flattening phenomenon may not complete, leading to suboptimal final accuracy.
\end{enumerate}

To address these challenges, we introduce a dual cutoff strategy inspired by the staged design of rocket launches:

\subsection{Three-Phase Training Dynamics}

\subsubsection{Ignition Phase}

We initialize two cutoffs:
\begin{itemize}
  \item \textbf{Minimum cutoff ($\rkmin$):} Set at the intersection point $\rkint(t)$
  \item \textbf{Maximum cutoff ($\rkmax$):} Set at the "elbow" of the singular value curve (see algorithm~\ref{alg:anagram-elbow})
\end{itemize}

The algorithm performs two natural gradient steps per iteration, one with each cutoff. If the intersection position remains static after both updates, we increment $\rkmax$ by one to promote exploration of additional gradient components.

This phase ends when $\rkmin$ reaches $\rkmax$—an event we term \textbf{liftoff}.

\subsubsection{Ascent Phase}

During ascent, both cutoffs track the moving intersection, but with a stability mechanism:
\begin{equation}
  \rkmax(t) = \max(\rkmax(t-1), \rkint(t)).
\end{equation}

This monotonicity constraint prevents the intersection rank from falling to zero, which would disrupt training dynamics.

\subsubsection{Stage Separation and Precision Locking}

When $\rce^S_{\rkint(t)}(t) \leq \epsilon$, we trigger \textbf{stage separation}:
\begin{itemize}
  \item $\rkmin$ is fixed at the precision level: $\rkmin = \rkeps(t)$
  \item $\rkmax$ continues tracking the intersection to maintain stability
\end{itemize}

The algorithm continues until $\rkmin = 0$ (\textbf{booster return}), indicating complete convergence.
The final algorithm that combines all three stages is mentioned in Algorithm~\ref{alg:anagram}.

\begin{algorithm}[H]
  \caption{Sketch of the Adaptative MultiCutoff Strategy for ANaGRAM (\method)}
  \label{alg:principled-adaptive}
  \DontPrintSemicolon
  \SetAlgoLined
  \KwIn{$\utheta:\RR^P\to\dL^2(\Omega,\mu)$, $\vtheta_0\in\RR^P$, $f\in\dL^2(\Omega,\mu)$, $(x_i)\in\Omega^S$, $\epsilon > 0$, $T_{\max} \in \NN$}

  \tcp{Initialization}
  $t \gets 0$\;
  $\ft[0] \gets \left(\partial_p \uthetaT[0](x_i)\right)_{i,p}$ for $i \in 1, \dots, S$ and $p \in 1, \dots, P$\;
  $\widehat{U}_0, \widehat{\Delta}_0, \widehat{V}_0^T \gets \SVD\left(\ft[0]\right)$\;
  $\ELgrad_0 \gets \left(\uthetaT[0](x_i) - f(x_i)\right)_i$ for $i \in 1, \dots, S$\;
  Compute $(\rce^S_j)$ for all $j \in 1, \dots \FRk$ following \cref{eqn:reconstruction-loss-orthogonal}\;

  \Repeat{$\rkeps = 0$ or $t \geq T_{\max}$}{
  \tcp{Compute adaptive ranks}
  Compute $\rkint$ and $\rkeps$ using expressions from \cref{eqn:adaptive-cutoff-rule}\;

  \tcp{Determine a final cutoff rank}
  \uIf{$\rce^S_{\rkint} > \epsilon$}{
  $\rkk \gets \rkint$ \tcp*{Track intersection}
  }
  \uElse{
    $\rkk \gets \rkeps$ \tcp*{Lock on precision}
  }

  \tcp{Natural gradient step}
  Set $\Dsing_t \gets \begin{cases} \Dsi & \text{if } i \leq \rkk, \\ 0 & \text{else;} \end{cases}$ \;
  Get new $\vtheta_{t+1}$ after one ANaGRAM step with \cref{eqn:eng-computation-trick}\;
  \tcp{Update for next iteration}
  $\ft[t+1] \gets \left(\partial_p \uthetaT[t+1](x_i)\right)_{i,p}$\;
  $\widehat{U}_{t+1}, \widehat{\Delta}_{t+1}, \widehat{V}_{t+1}^T \gets \SVD\left(\widehat{\phi}_{t+1}\right)$\;
  $\ELgrad_{t+1} \gets \left(\uthetaT[t+1](x_i) - f(x_i)\right)_i$\;
  Recompute $\rce^S_j$ for all $j \in 1, \dots \FRk$ following \cref{eqn:reconstruction-loss-orthogonal}\;
  $t \gets t + 1$\;
  }

  \KwOut{$\vtheta_t$}
\end{algorithm}

\subsection{Complete Practical Algorithm}

\SetKwFor{ForEachBlue}{\textcolor{blue}{foreach}}{\textcolor{blue}{do}}{endfch}
\SetKwFunction{SVD}{SVD}
\SetKwFunction{RCE}{ReconstructionErrors}
\SetKwFunction{FE}{FindElbow}
\begin{algorithm}[H]
  \caption{AMStraMGRAM : Adaptive Multicutoff Strategy Modification for ANaGRAM}
  \label{alg:anagram}
  \DontPrintSemicolon
  \KwIn{\vspace{-4mm}
    \begin{itemize}
      \item $u:\RR^P\to\dL^2(\Omega,\mu)$ ~\tcp{neural network architecture}
      \item $\vtheta_0\in\RR^P$ ~\tcp{initialization of the neural network}
      \item $f\in\dL^2(\Omega,\mu)$ ~\tcp{target function of the quadratic regression}
      \item $(x_i)\in\Omega^S$ ~\tcp{a batch in $\Omega$}
            \begingroup
            \color{blue}
            \SetCommentSty{bluecommentfont}
      \item $\epsilon > 0$ ~\tcp{precision level of the optimization}
            \endgroup
    \end{itemize}
  }
  \begingroup
  \color{blue}
  \SetCommentSty{bluecommentfont}
  \Begin(Initialization){
    $\liftoff\gets False$ ~\tcp{Liftoff indicator}
    $\widehat{\phi}_{\vtheta_0} \gets \left(\partial_p u_{\vtheta_0}(x_i)\right)_{1 \leq i \leq S,\, 1 \leq p \leq P}$ \tcp*{Computed \via auto-differentiation}
    $\widehat{U}_{\vtheta_0}, \widehat{\Delta}_{\vtheta_0}, \widehat{V}_{\vtheta_0}^t \gets \SVD\left(\widehat{\phi}_{\vtheta_0}\right)$\;
    $\ELgrad_{\vtheta_0} \gets \left(u_{\vtheta_0}(x_i) - f(x_i)\right)_{1 \leq i \leq S}$\;
    $\rce^S_0\gets\RCE\left(\widehat{V}_{\vtheta_0}^t,\ELgrad_{\vtheta_0} \right)$\;
    ${\rkmax}_0 \gets \FE\left((1,\ldots,\FRk),\widehat{\Delta}_{\vtheta_0}\right)$\;
  }
  \endgroup
  \Repeat{\color{blue}${\rkOne}_t=0$ or $t\geq T_{\max}$}{
  \label{algl:phi-cpmputation-anagram}
  \begingroup
  \SetCommentSty{bluecommentfont}
  \color{blue}
  ${\rkOne}_t \gets \#\left\{\, \rce^S_{0_j} \leq \widehat{\Delta}_{\vtheta_{t_j}}\,:\,1\leq j\leq\FRk\,\right\}$\; 
  ${\rkTwo}_t \gets \#\left\{\, \rce^S_{0_j} \geq \epsilon\,:\,1\leq j\leq\FRk\,\right\}$\;
  \tcc{with \# standing for the cardinal}
  ${\rkmin}_t \gets \min({\rkOne}_t, {\rkTwo}_t)$\;
  ${\rkmax}_t \gets \max({\rkOne}_t, {\rkmax}_{t-1})$\;

  \uIf{$\textrm{not}\,{\liftoff}_t$}{
  \uIf{${\rkmin}_t \geq {\rkmax}_t$}{
    ${\liftoff}_t \gets \text{True}$\;
  }
  \uElseIf{${\rkmin}_{t-1} = {\rkmin}_t$}{
        ${\rkmax}_t \gets {\rkmax}_t + 1$\;
      }
  }

  \endgroup
  \ForEachBlue{\color{blue}$\rkk\in\left\{{\rkmax}_t,{\rkmin}_t\right\}$}{
  $\widehat{\Delta}_{\vtheta_t} \gets \left(\widehat{\Delta}_{\vtheta_t, p} \text{ if } \color{blue}{p \geq \rkk} \color{black} \text{ else } 0\right)_{1 \leq p \leq P}$\;
  $\ELgrad_{\vtheta_t} \gets \left(u_{\vtheta_t}(x_i) - f(x_i)\right)_{1 \leq i \leq S}$\;
  $d_{\vtheta_t} \gets \widehat{V}_{\vtheta_t} \widehat{\Delta}_{\vtheta_t}^\dagger \widehat{U}_{\vtheta_t}^t \, \ELgrad_{\vtheta_t}$\;
  $\eta_t \gets \argmin_{\eta \in \RR^+} \sum_{1 \leq i \leq S} \left(f(x_i) - u_{\vtheta_t - \eta d_{\vtheta_t}}(x_i)\right)^2$ \tcp*{via line search}
  $\vtheta_{t+1} \gets \vtheta_t - \eta_t \, d_{\vtheta_t}$\;
  $\widehat{\phi}_{\vtheta_{t+1}} \gets \left(\partial_p u_{\vtheta_{t+1}}(x_i)\right)_{1 \leq i \leq S,\, 1 \leq p \leq P}$ \tcp*{Computed \via auto-differentiation}
  $\widehat{U}_{\vtheta_{t+1}}, \widehat{\Delta}_{\vtheta_{t+1}}, \widehat{V}_{\vtheta_{t+1}}^t \gets \SVD\left(\widehat{\phi}_{\vtheta_{t+1}}\right)$\;
  }
  }
\end{algorithm}
\subsection{Empirical Justification for Design Choices} \label{app:practical-justifications}

The dual cutoff strategy addresses specific empirical challenges we observed:

\begin{description}
  \item[Dual gradient steps:] Without the second cutoff, training dynamics sometimes stagnate. The dual approach provides both stability (via $\rkmin$) and exploration (via $\rkmax$).

  \item[Elbow initialization:] The elbow point marks where singular values cease contributing meaningful signal, providing a natural upper bound for exploration.

  \item[Monotonic $\rkmax$:] Prevents catastrophic retreat of the intersection point, which we observed in complex equations like Allen-Cahn.

  \item[Stage separation timing:] Triggered precisely when the intersection error drops below target precision, ensuring optimal utilization of the flattening phenomenon.
\end{description}

We see in the next section how this practical algorithm successfully improve empirical robustness.

\section{Algorithmic details}\label{app:algorithmic-details}

\begin{algorithm}[H]
  \SetAlgoLined
  \DontPrintSemicolon
  \caption{Find elbow}
  \DefFun{\FE}{
    \KwIn{\vspace{-3.5mm}
      \begin{itemize}
        \item[-] $(x_i)\in\RR^m$ ~\tcp{an increasing sequence of $m\in\NN$ points in $\RR$}
        \item[-] $\widehat{f}\in\RR^m$ ~\tcp{a decreasing function evaluated at points $(x_i)$}
      \end{itemize}}
    \tcc{Clockwise normal vector to $\left(x_m-x_1,\widehat{f}_m-\widehat{f}_1\right)$}
    $\overrightarrow{n}\gets \left(\widehat{f}_m-\widehat{f}_1, x_1-x_m\right)\in\RR^2$\;
    $\left(s_j\right)_{1\leq j\leq m}\gets \left(\braket{\overrightarrow{n}}{\left(x_j-x_1,\widehat{f}_j-\widehat{f}_1\right)}_{\RR^2}\right)_{1\leq j\leq m}$\;
    %
    \KwOut{$\argmax\limits_{1\leq j\leq m} s_j$}
  }
  \label{alg:anagram-elbow}
\end{algorithm}

\todoAS[inline]{If we have time, the algorithm FindElbow can be efficiently illustrated with a picture }

\begin{algorithm}[H]
  \DontPrintSemicolon
  \SetAlgoLined
  \caption{Reconstruction Errors}
  \DefFun{\RCE}{
  \KwIn{\vspace{-4mm}
    \begin{itemize}
      \item[-] $\widehat{V}^t\in\RR^{\FRk,S}$ ~\tcp{right singular vectors of the Jacobian $\widehat{\phi}$}
      \item[-] $\ELgrad\in\RR^{S}$ ~\tcp{Evaluated functional gradient}
    \end{itemize}}
  \Begin(Initialization){
    $\cumapprox\gets 0\in\RR^{S}$ ~\tcp{cumulative approximation of $\ELgrad$}
    $\rce^S\gets 0 \in\RR^{\FRk}$ ~\tcp{cumulated reconstruction erros}
    $\rotatedgrad\gets \widehat{V}^t \ELgrad\in\RR^{\FRk}$\;
  }
  \ForEach{$j\in (1, \ldots, \FRk)$}{
  $\cumapprox\gets \cumapprox + {\rotatedgrad}_j$\;
  $\rce^S_j\gets \norm[\cumapprox-\rotatedgrad]_2$\;
  }
  \KwOut{$\rce^S$}
  }
  \label{alg:anagram-reconstruction-error}
\end{algorithm}

\todoAS[inline]{Instead of Algorithm 3, I suggest to refer to \cref{eqn:reconstruction-loss-orthogonal} wherever we need to say that we compute reconstruction error. Just mention that as ANaGRAM already computes SVD, evaluating \cref{eqn:reconstruction-loss-orthogonal} can be done at a negligible extra cost. }

\section{Empirical example of Anagram Training Dynamics} \label{app:anagram_training_dynamics}

\begin{figure}[H]
  \centering
  \captionsetup[subfigure]{width=.95\linewidth}
  \begin{subfigure}[t]{.33\textwidth}
    \centering
    \includegraphics[width=\linewidth]{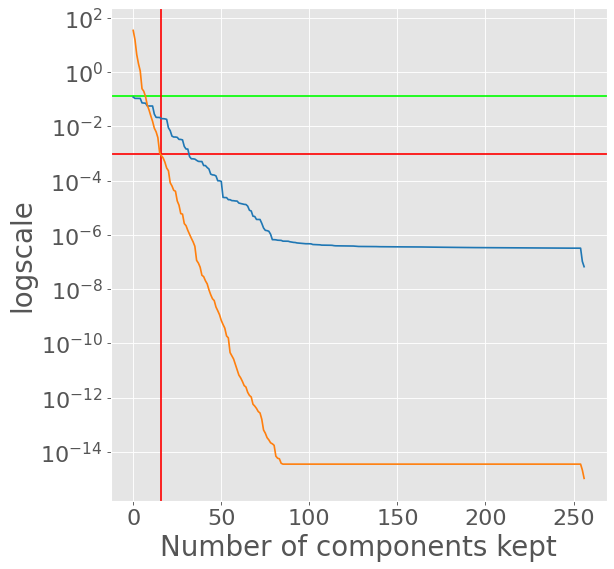}
    \caption{Iteration 0: intersection point between singular values and reconstruction error lies before cutoff.}
    \label{fig:anagram_iter_0}
  \end{subfigure}%
  \hfill
  \begin{subfigure}[t]{.33\textwidth}
    \centering
    \includegraphics[width=\linewidth]{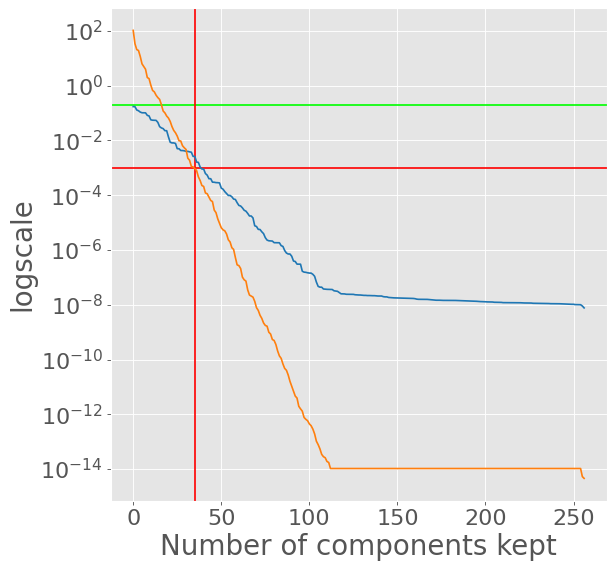}
    \caption{Iteration 40: intersection point shifts rightward toward cutoff.}
    \label{fig:anagram_iter_40}
  \end{subfigure}
  \hfill
  \begin{subfigure}[t]{.33\textwidth}
    \centering
    \includegraphics[width=\linewidth]{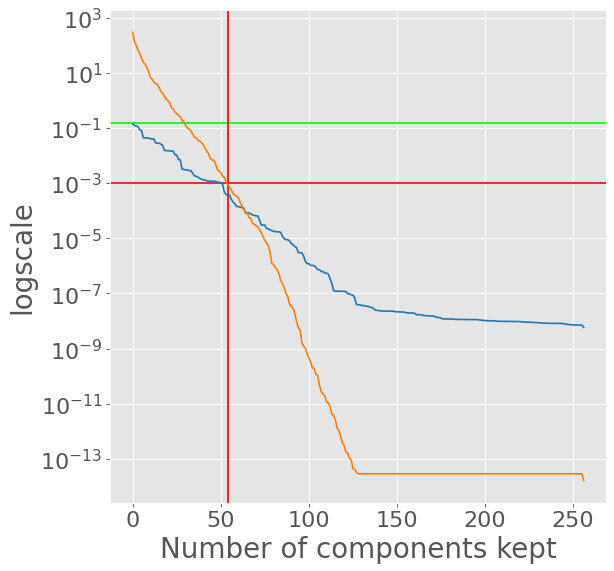}
    \caption{Iteration 90: intersection point passes the cutoff threshold.}
    \label{fig:anagram_iter_90}
  \end{subfigure}%
  \vskip\baselineskip
  \captionsetup[subfigure]{width=\linewidth} 
  \begin{subfigure}[t]{.49\textwidth}
    \raggedright
    \includegraphics[width=.67\linewidth]{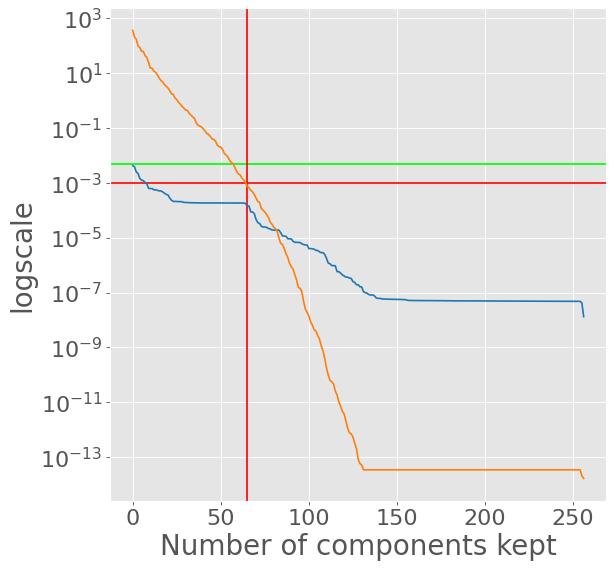}
    \caption{Iteration 120. Beginning of \emph{flattening}:
      reconstruction errors stabilizes at constant level before cutoff.
    }
    \label{fig:anagram_iter_120}
  \end{subfigure}
  \hfill
  \begin{subfigure}[t]{.01\textwidth}
    \centering
    \hspace*{-16\linewidth}\includegraphics[width=33\linewidth]{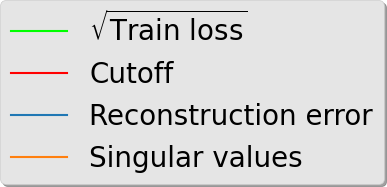}
  \end{subfigure}%
  \hfill
  \begin{subfigure}[t]{.49\textwidth}
    \raggedleft
    \includegraphics[width=.67\linewidth]{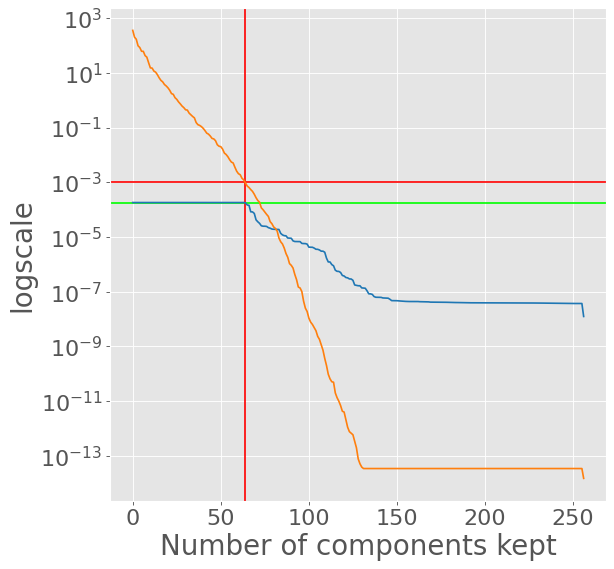}
    \caption{Iteration 150: Complete flattening. Training loss reaches the flattened reconstruction error level.}
    \label{fig:anagram_iter_150}
  \end{subfigure}
  \caption{\textbf{Evolution of quantities of interest during ANaGRAM training on heat equation.} The dynamics reveal two distinct phases culminating in reconstruction error flattening.}
  \label{fig:anagram_training_evolution}
\end{figure}

\begin{figure}[H]
  \centering
  \captionsetup[subfigure]{width=.95\linewidth}
  \begin{subfigure}[t]{.7\textwidth}
    \centering
    \includegraphics[width=\linewidth]{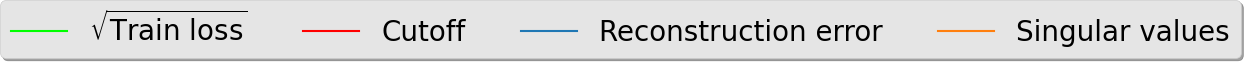}
  \end{subfigure}
  \vskip\baselineskip
  \begin{subfigure}[t]{.33\textwidth}
    \centering
    \includegraphics[width=\linewidth]{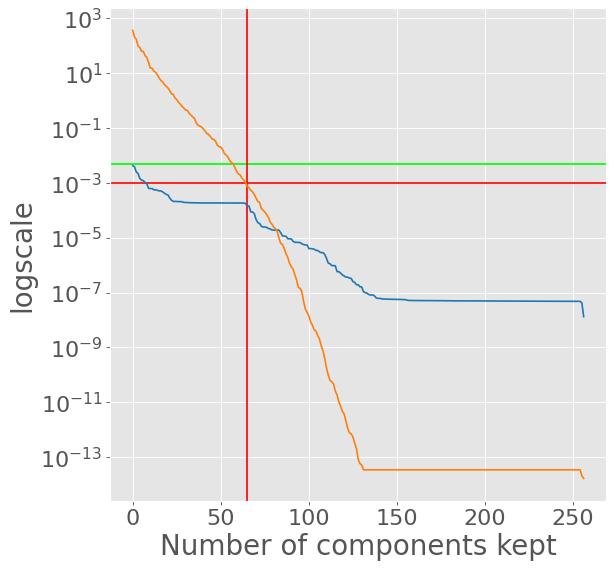}
    \caption{Same as \cref{fig:anagram_iter_120}: iteration 120 of ANaGRAM with a fixed cutoff at $10^{-3}$.}
  \end{subfigure}%
  \hfill
  \begin{subfigure}[t]{.33\textwidth}
    \centering
    \includegraphics[width=\linewidth]{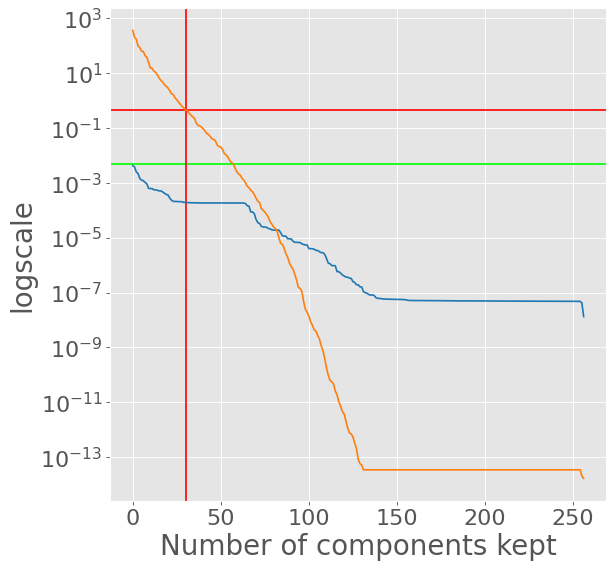}
    \caption{Still iteration 120, but now showing a new cutoff such that the number of retained components is 30, which is roughly the location of the "elbow" in the reconstruction error curve.}
  \end{subfigure}
  \hfill
  \begin{subfigure}[t]{.33\textwidth}
    \centering
    \includegraphics[width=\linewidth]{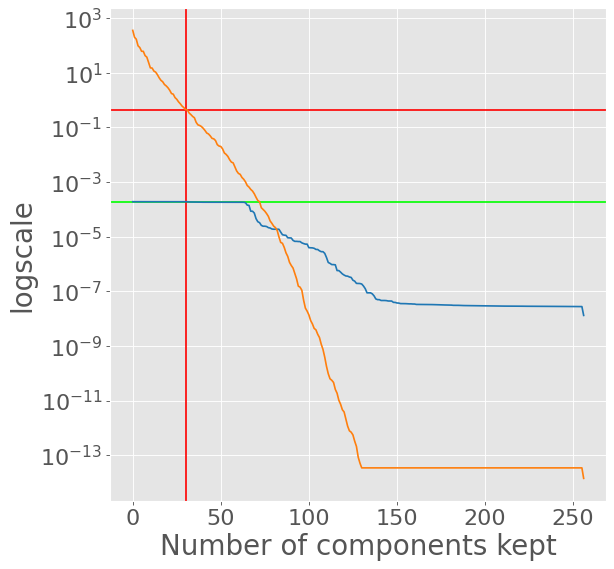}
    \caption{After applying \textbf{a single natural gradient step with the new cutoff}. The result is a completed flattening of the reconstruction error curve for all retained components, aligning with the previous flattening level. This reduces the square root of the training loss by two orders of magnitude in just one step.}
  \end{subfigure}%
  \caption{Illustration of ``instant flattening'' through adaptive cutoff adjustment. A single step with adjusted cutoff completes the flattening process.}
  \label{fig:adapted-cutoff-flattening}
\end{figure}

In \cref{fig:anagram_training_evolution}, we analyze ANaGRAM's training on the heat equation with a fixed cutoff threshold $\Ccutoff = 10^{-3}$ and line search for the learning rate. The training loss coincides with $\norm[{\ELgrad}]^2$. We can see the flattening phenomenon to occur on Iteration 120 and completed at 150. As discussed in the main paper, sometimes the flattening can be incomplete, and for many iterations remain without any further progress ($\Nflat$ never reaching zero). In this case, changing a cutoff threshold results in an immediate and complete flattening for all first components up to $\rkk$, which is demonstrated in \cref{fig:adapted-cutoff-flattening} for Iteration 120 of \cref{fig:anagram_training_evolution}.

\section{Deep dive on selected experiments}

In this section we look at curves of training and estimations obtained with \method on benchmark of PDEs.

\subsection{One Dimensional Burgers Equation}

\begin{figure}[H]
  \centering
  \includegraphics[width=\linewidth]{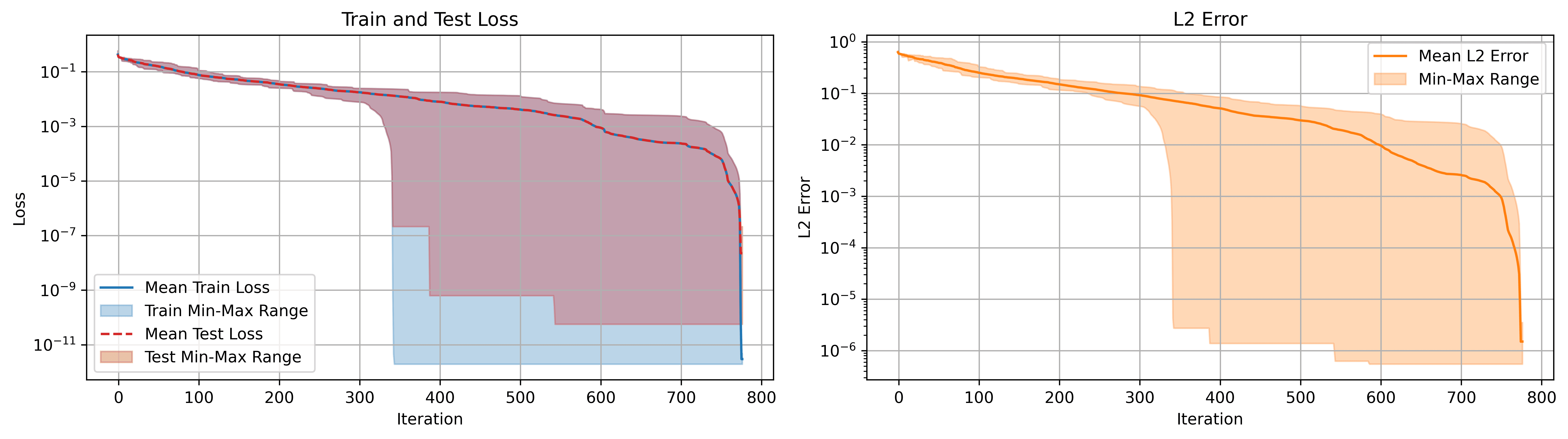}
  \caption{Training metrics for the One-Dimensional Burgers equation, showing convergence behavior with our adaptive multi-cutoff strategy.}
  \label{fig:1db_training_metrics}
\end{figure}

\begin{figure}[H]
  \centering
  \begin{subfigure}[t]{0.32\textwidth}
    \centering
    \includegraphics[width=\linewidth]{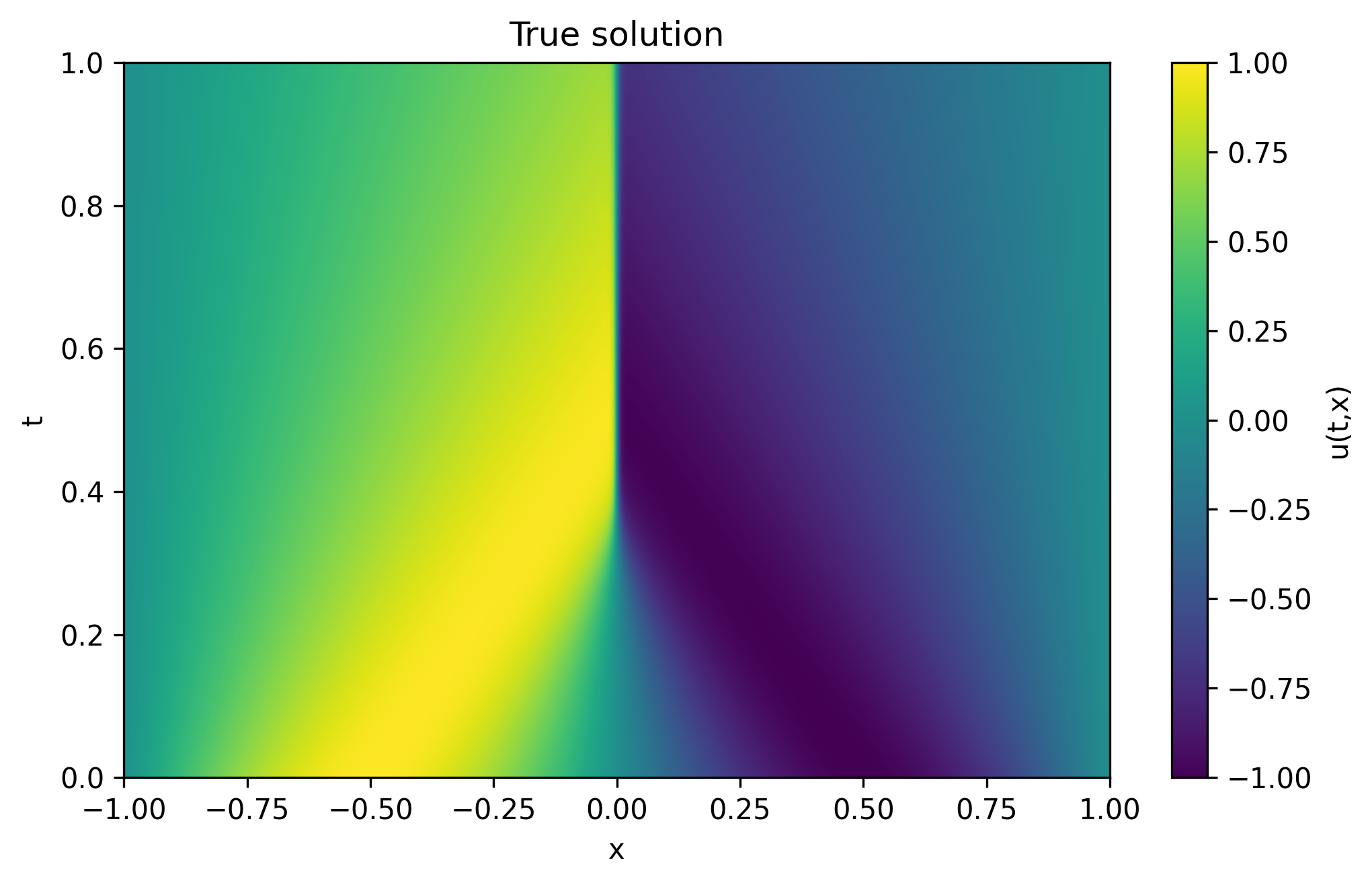}
    \caption{True solution}
    \label{fig:burgers_true}
  \end{subfigure}%
  \hfill
  \begin{subfigure}[t]{0.32\textwidth}
    \centering
    \includegraphics[width=\linewidth]{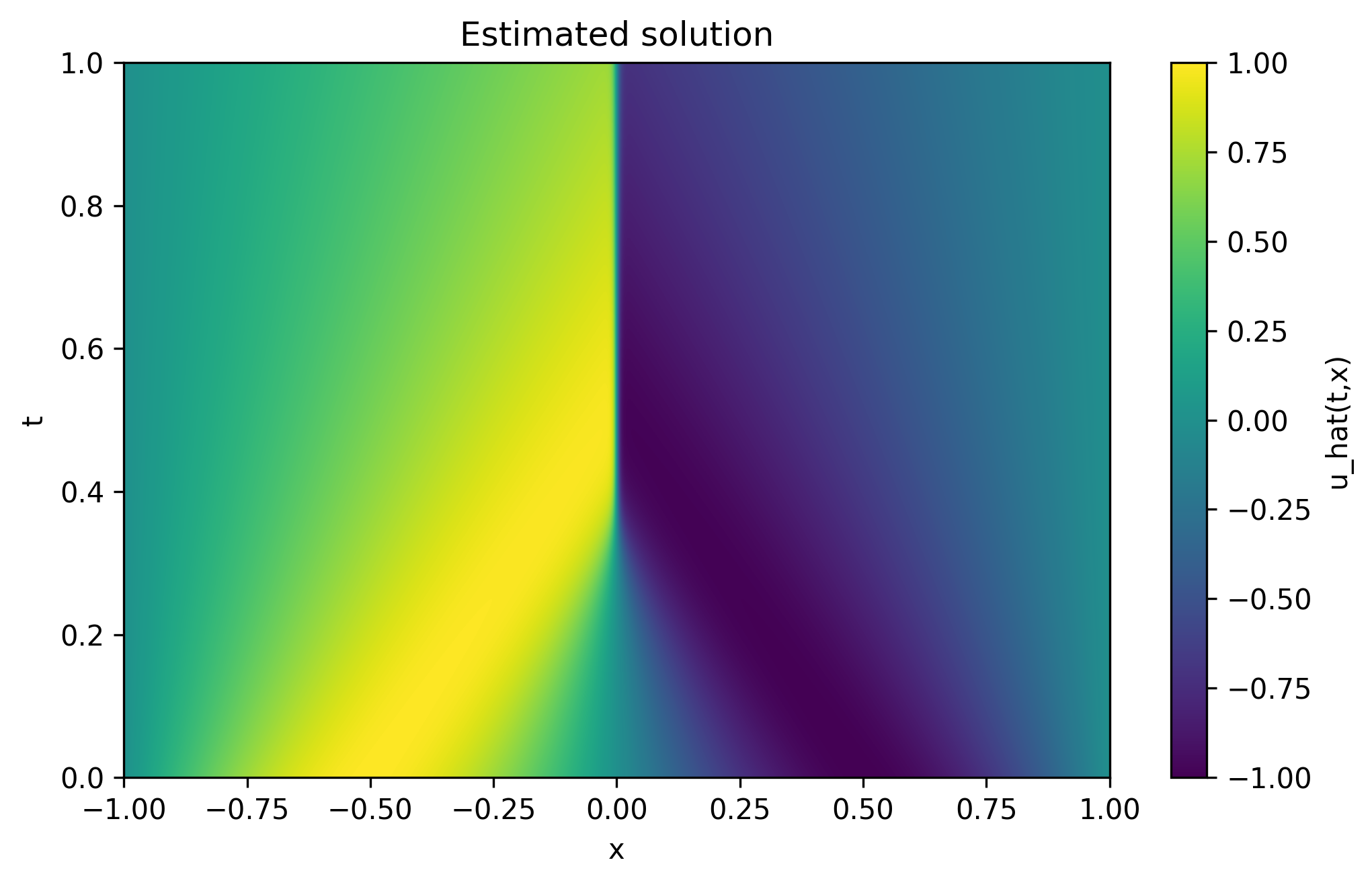}
    \caption{Estimated solution}
    \label{fig:burgers_estimated}
  \end{subfigure}%
  \hfill
  \begin{subfigure}[t]{0.32\textwidth}
    \centering
    \includegraphics[width=\linewidth]{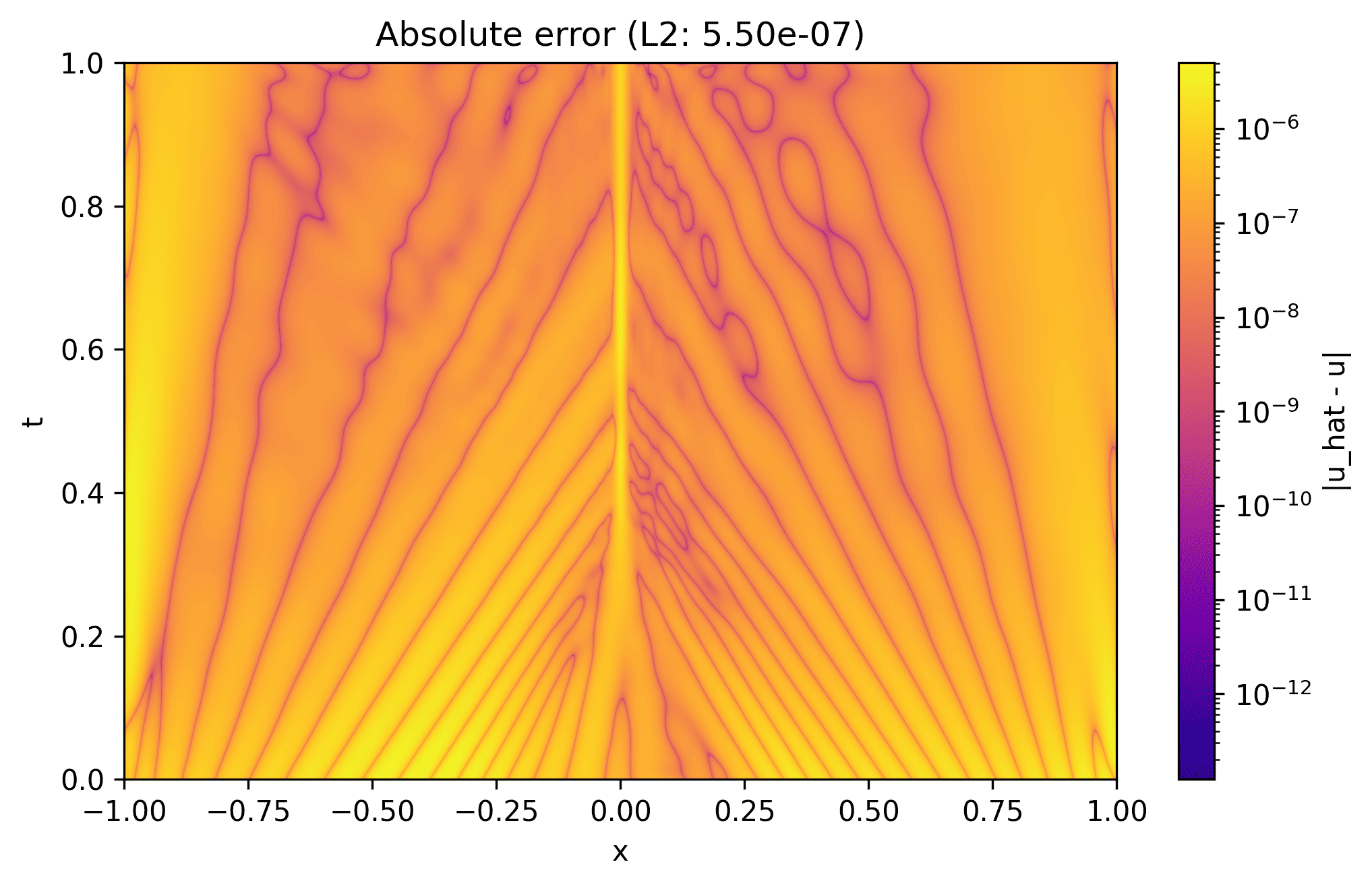}
    \caption{Error}
    \label{fig:burgers_error}
  \end{subfigure}
  \caption{Results for One Dimensional Burgers Equation with cutoff $10^{-6}$.}
  \label{fig:burgers_results}
\end{figure}
\subsection{Heat Equation}

\begin{figure}[H]
  \centering
  \includegraphics[width=\linewidth]{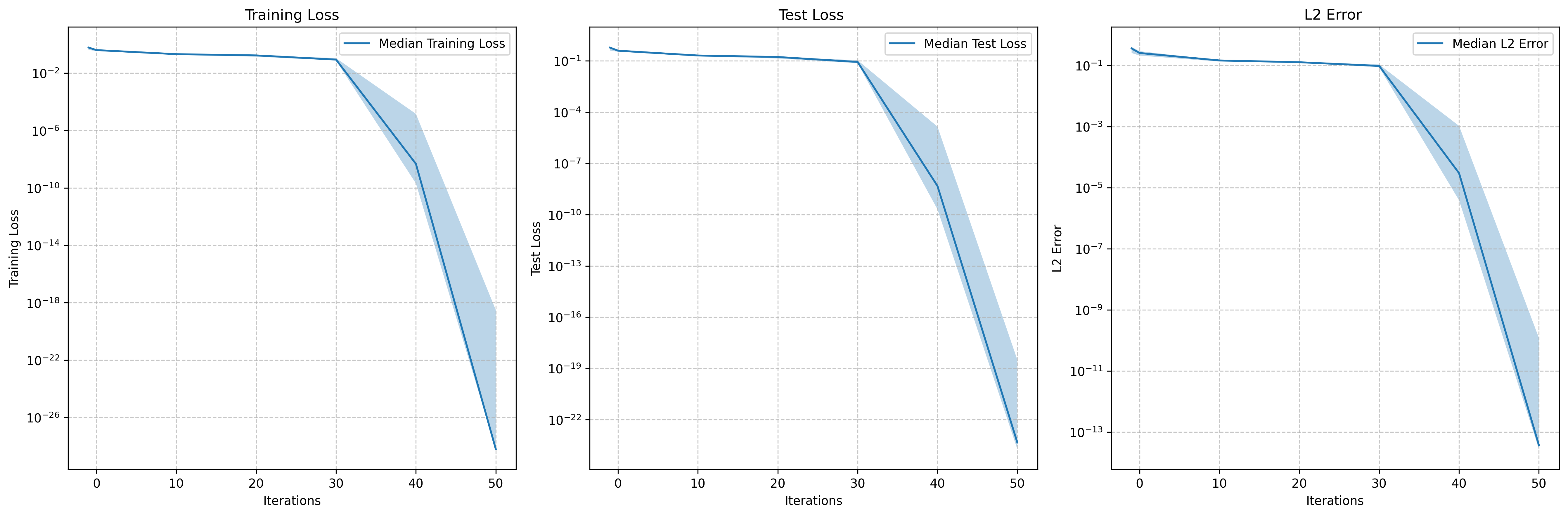}
  \caption{Convergence results for the Heat equation showing the $L_2$ error over iterations. Our method (\method) converges faster and reaches a lower final error than ANaGRAM and baselines. Variability across runs is due to differing feature development speed from the random initialization.}
  \label{fig:heat_convergence}
\end{figure}

\begin{figure}[H]
  \centering
  \begin{subfigure}[t]{0.32\textwidth}
    \centering
    \includegraphics[width=\linewidth]{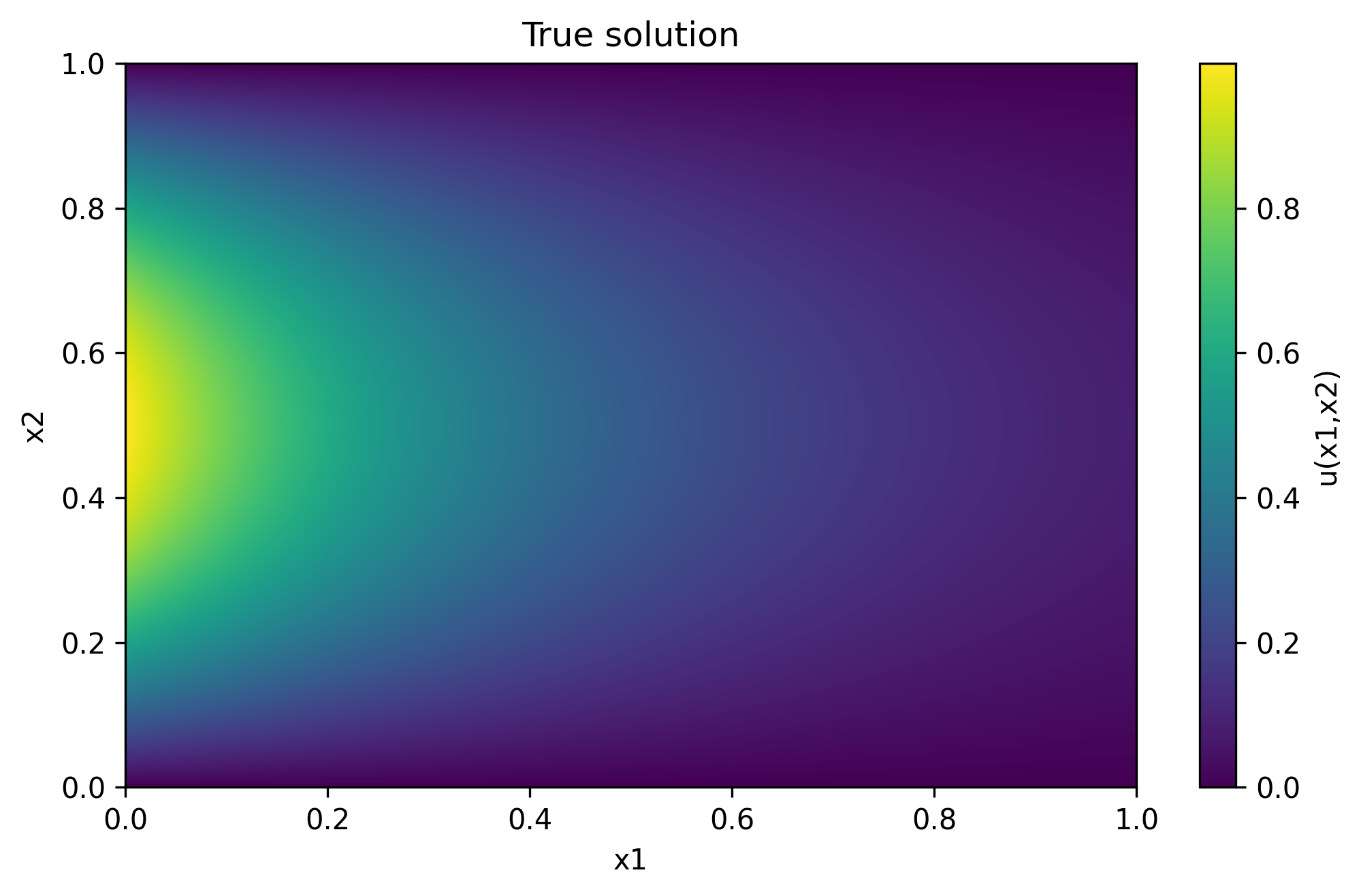}
    \caption{True solution}
    \label{fig:HEAT_true}
  \end{subfigure}%
  \begin{subfigure}[t]{0.32\textwidth}
    \centering
    \includegraphics[width=\linewidth]{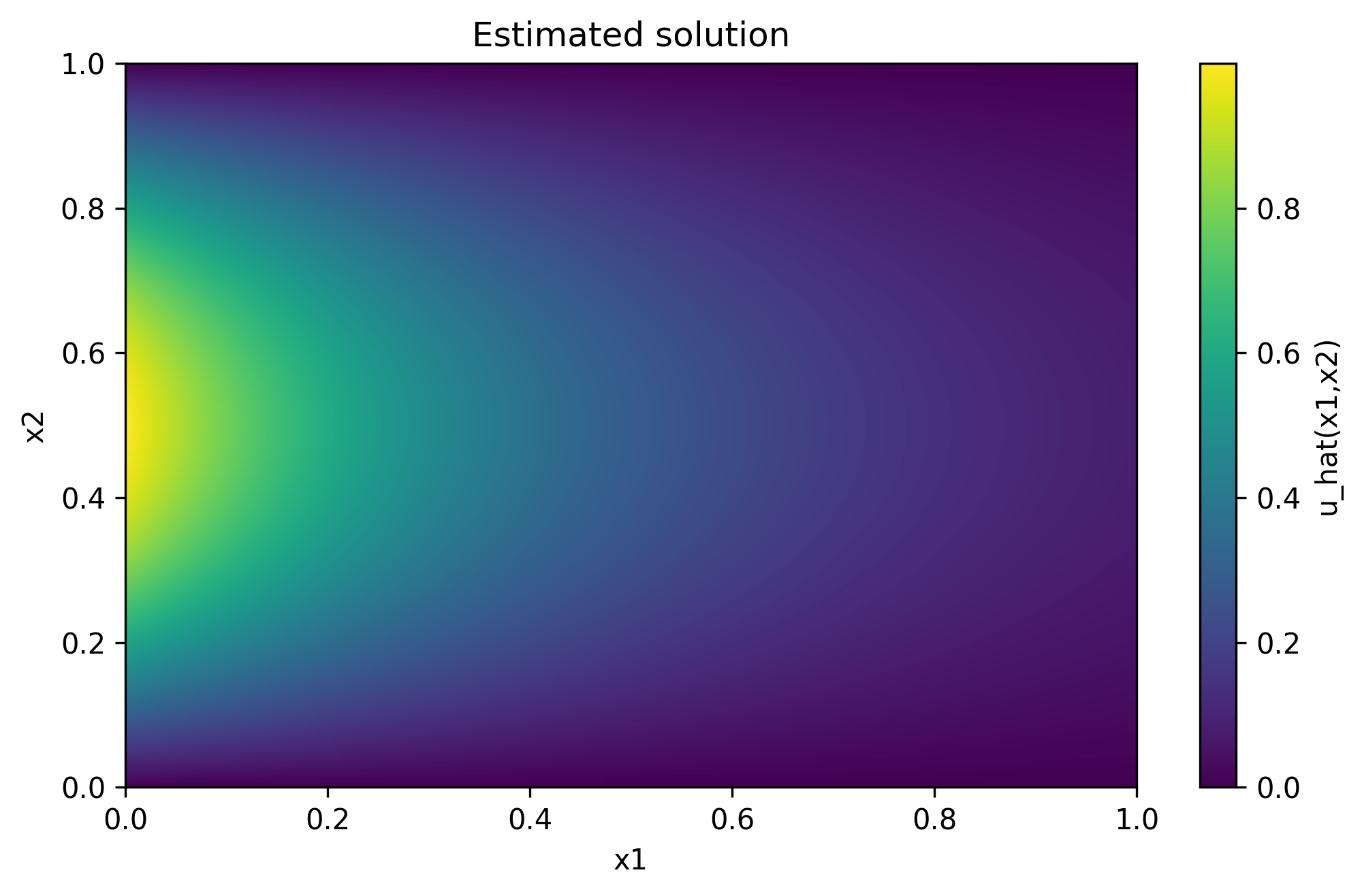}
    \caption{Estimated solution}
    \label{fig:HEAT_estimated}
  \end{subfigure}%
  \begin{subfigure}[t]{0.32\textwidth}
    \centering
    \includegraphics[width=\linewidth]{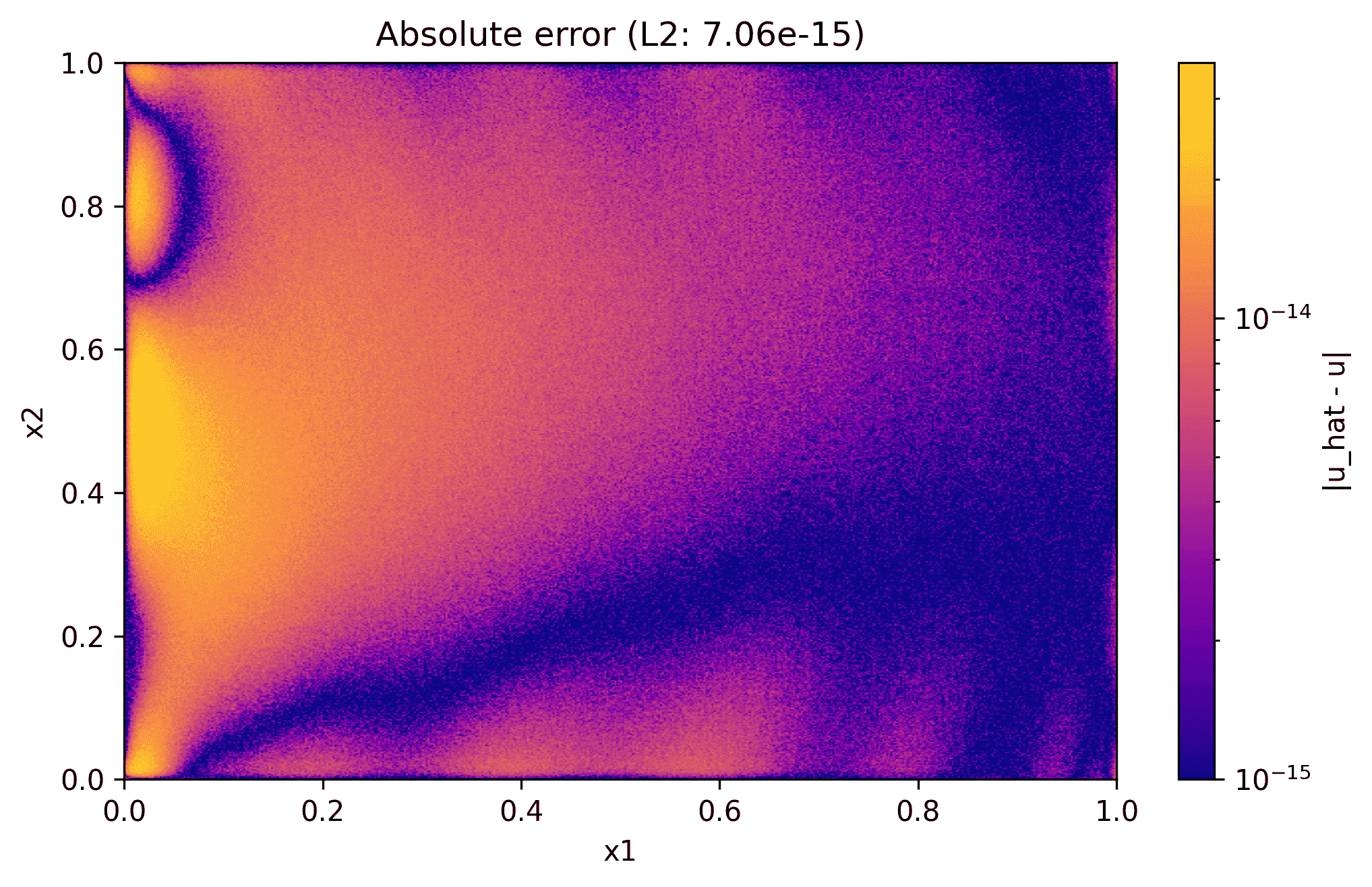}
    \caption{Error}
    \label{fig:HEAT_error}
  \end{subfigure}
  \caption{Results for the Heat equation (solution cutoff $10^{-14}$). The error remains uniformly low over the domain, illustrating the effectiveness of the adaptive multi-cutoff strategy.}
  \label{fig:heat_results}
\end{figure}

\todoAS[inline]{If time allows, increase the font size of the figure above to be more readable}

\subsection{Laplace Equations (L2D and L5D)}

For the Laplace equation in 2D, our method also demonstrates remarkable performance improvements over the baselines. The convergence is both faster and reaches a significantly lower error plateau.

\begin{figure}[H]
  \centering
  \includegraphics[width=\linewidth]{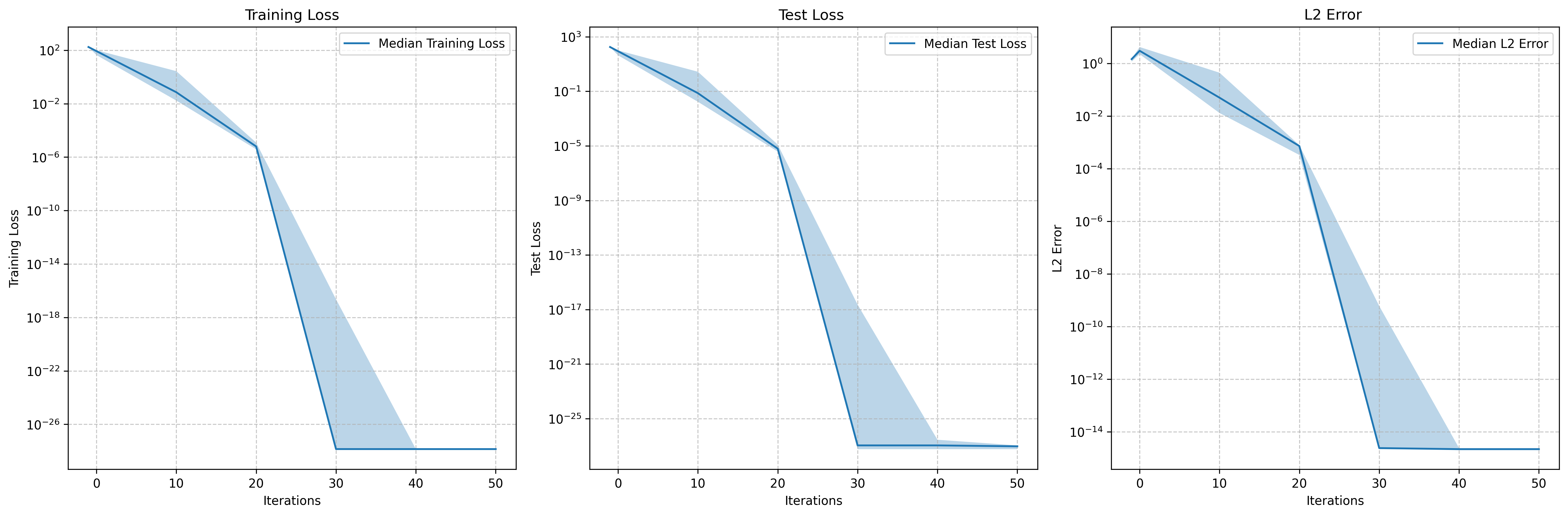}
  \caption{Convergence results for the Laplace 2D problem, showing the $L_2$ error over iterations. Our method (\method) achieves both faster convergence and lower final error compared to ANaGRAM and other baseline methods. The observed variance between runs can be explained by different speed of convergence depending on the initialization. 
  }
  \label{fig:laplace2d_convergence}
\end{figure}

\begin{figure}[H]
  \centering
  \begin{subfigure}[t]{0.32\textwidth}
    \centering
    \includegraphics[width=\linewidth]{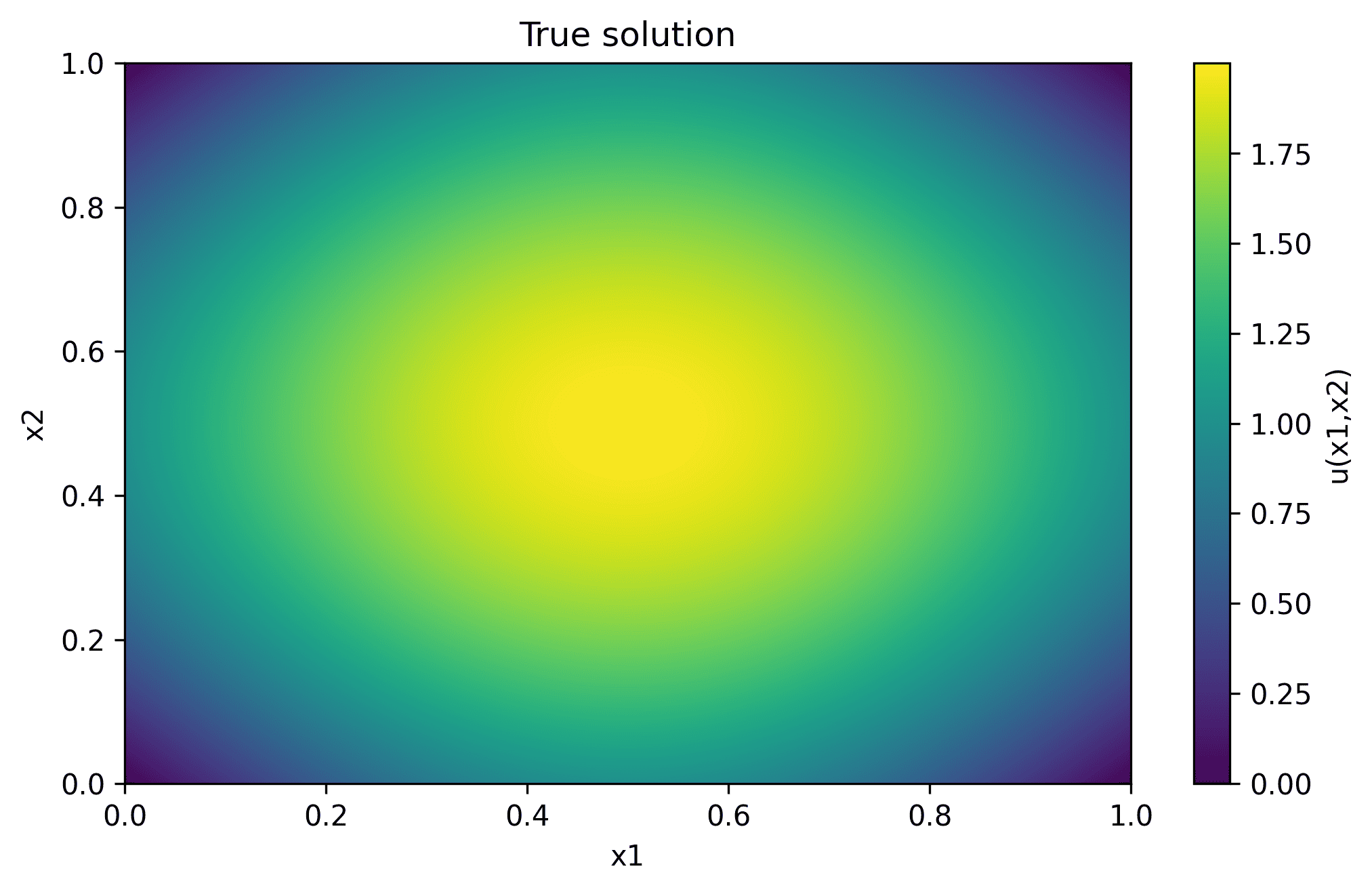}
    \caption{True solution}
    \label{fig:L2D_true}
  \end{subfigure}%
  \hfill
  \begin{subfigure}[t]{0.32\textwidth}
    \centering
    \includegraphics[width=\linewidth]{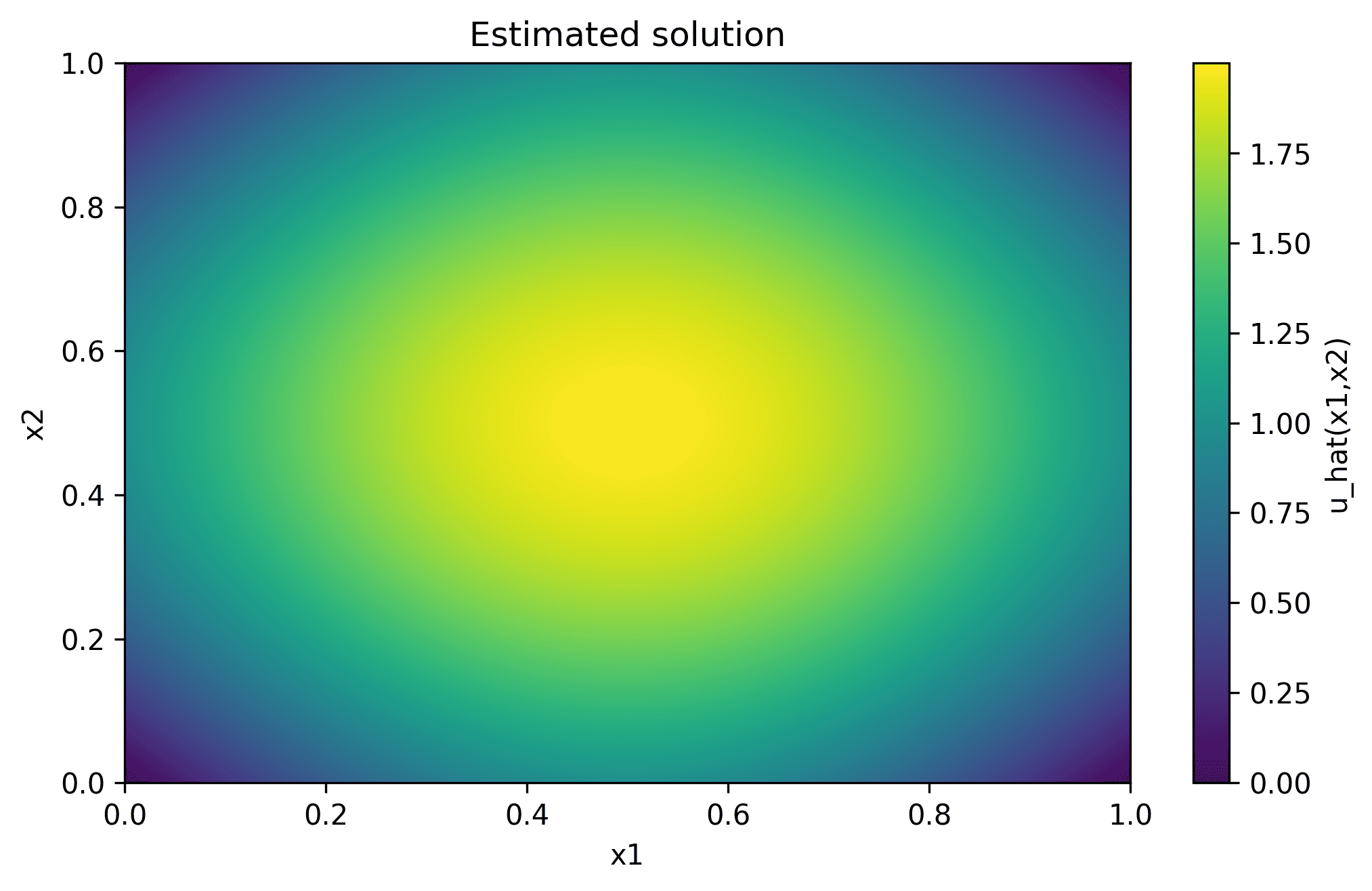}
    \caption{Estimated solution}
    \label{fig:L2D_estimated}
  \end{subfigure}%
  \hfill
  \begin{subfigure}[t]{0.32\textwidth}
    \centering
    \includegraphics[width=\linewidth]{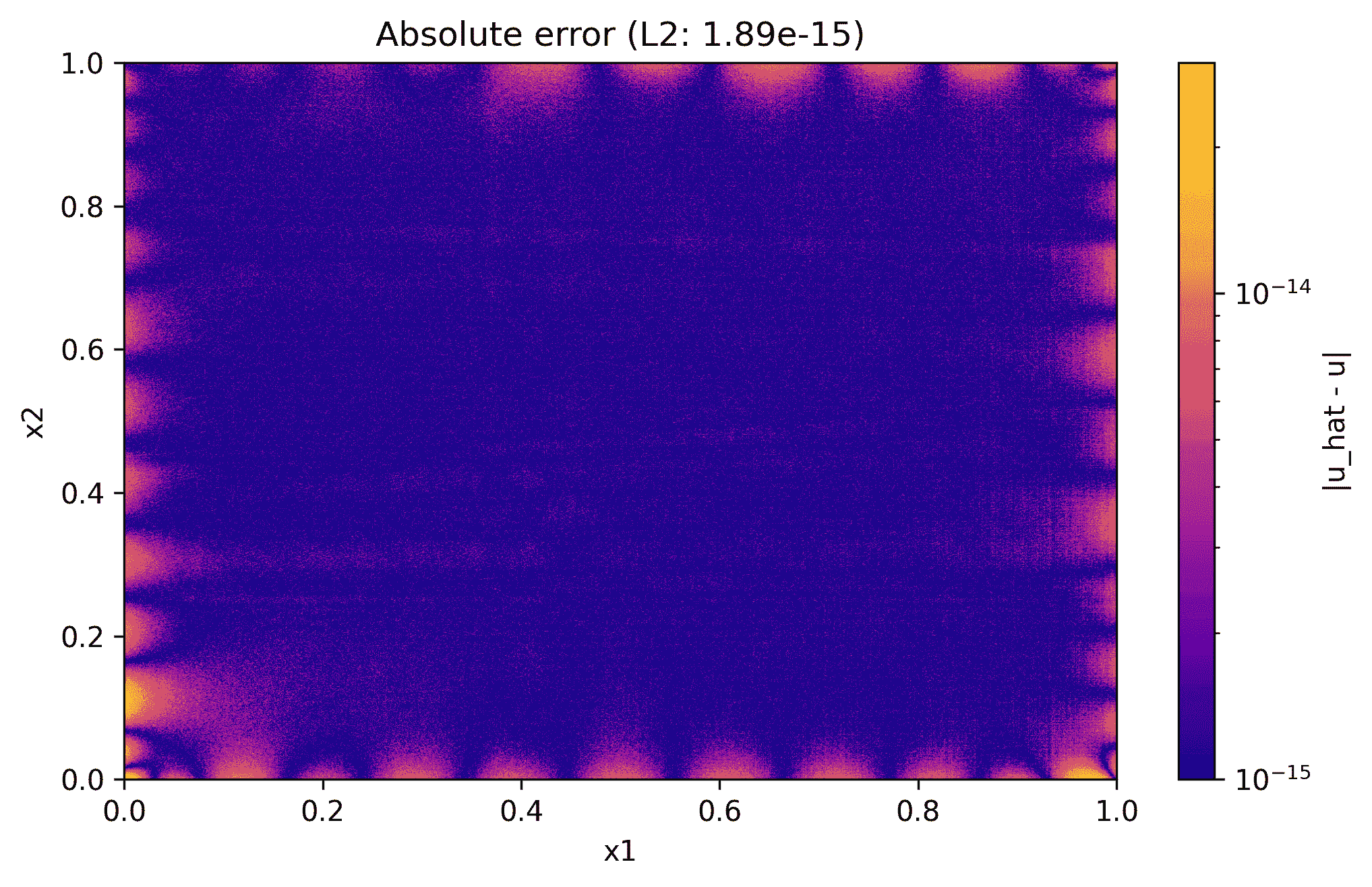}
    \caption{Error}
    \label{fig:L2D_error}
  \end{subfigure}
  \caption{Results for Laplace 2D Equation with cutoff $10^{-6}$.}
  \label{fig:l2d_results}
\end{figure}

\begin{figure}[H]
  \centering
  \includegraphics[width=\linewidth]{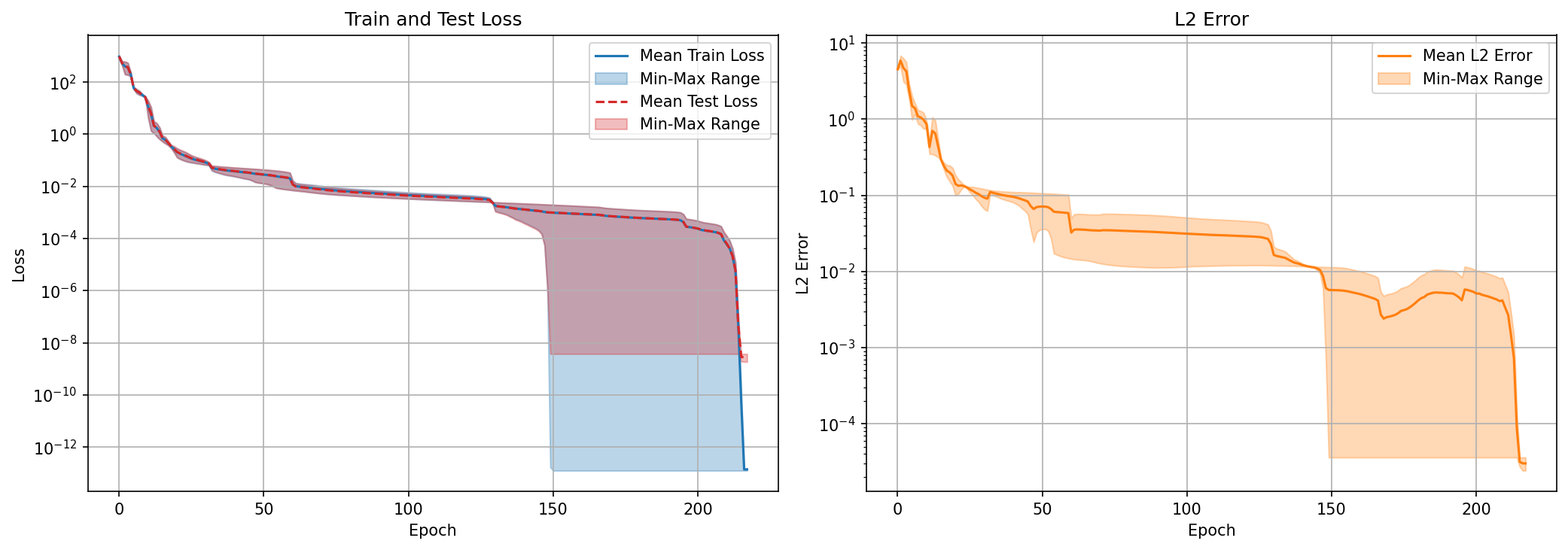}
  \caption{Convergence results for the Laplace 5D problem, showing the $L_2$ error over iterations. Our method (\method) achieves  faster convergence but not lower final error compared to ANaGRAM and other baseline methods. We see that seeds change the speed of convergence of the algorithm}
  \label{fig:laplace5d_convergence}
\end{figure}

\todoAS[inline]{The above figures have rather repetitive captions. I would remove the observations from the caption and just write one paragraph that summarizes the observations from all figures at once.}

\todoAS[inline]{If time allows, add equations for each PDE mentioned in this part of the Appendix.}

\subsection{Non Linear Poisson Equation}

\todoAS[inline]{see my comments from the subsections before}

To compare ourselves with \cite{urbanUnveilingOptimizationProcess2025}, we select (K=1).

\begin{figure}[H]
  \centering
  \includegraphics[width=\linewidth]{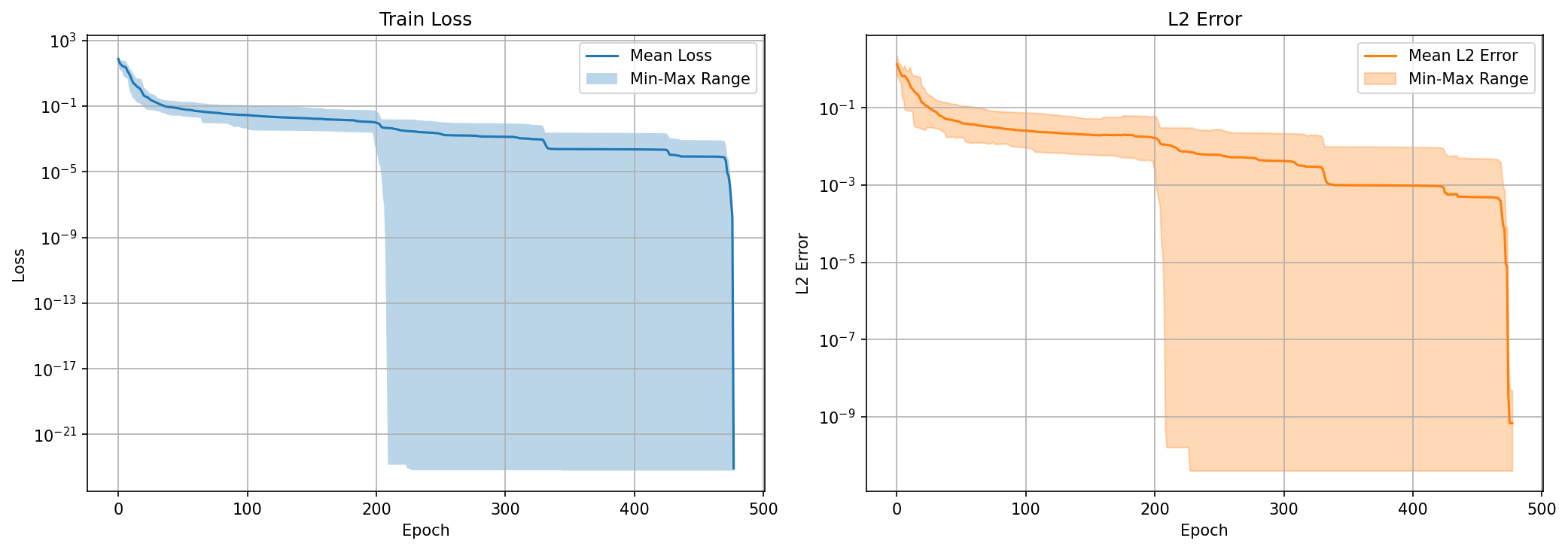}
  \caption{Convergence results for the Non Linear Poisson equation, showing the $L_2$ error over iterations. Our method (\method) achieves both faster convergence and lower final error compared to ANaGRAM and other baseline methods.}
  \label{fig:non_linear_poisson_convergence}
\end{figure}

\begin{figure}[H]
  \centering
  \begin{subfigure}[t]{0.32\textwidth}
    \centering
    \includegraphics[width=\linewidth]{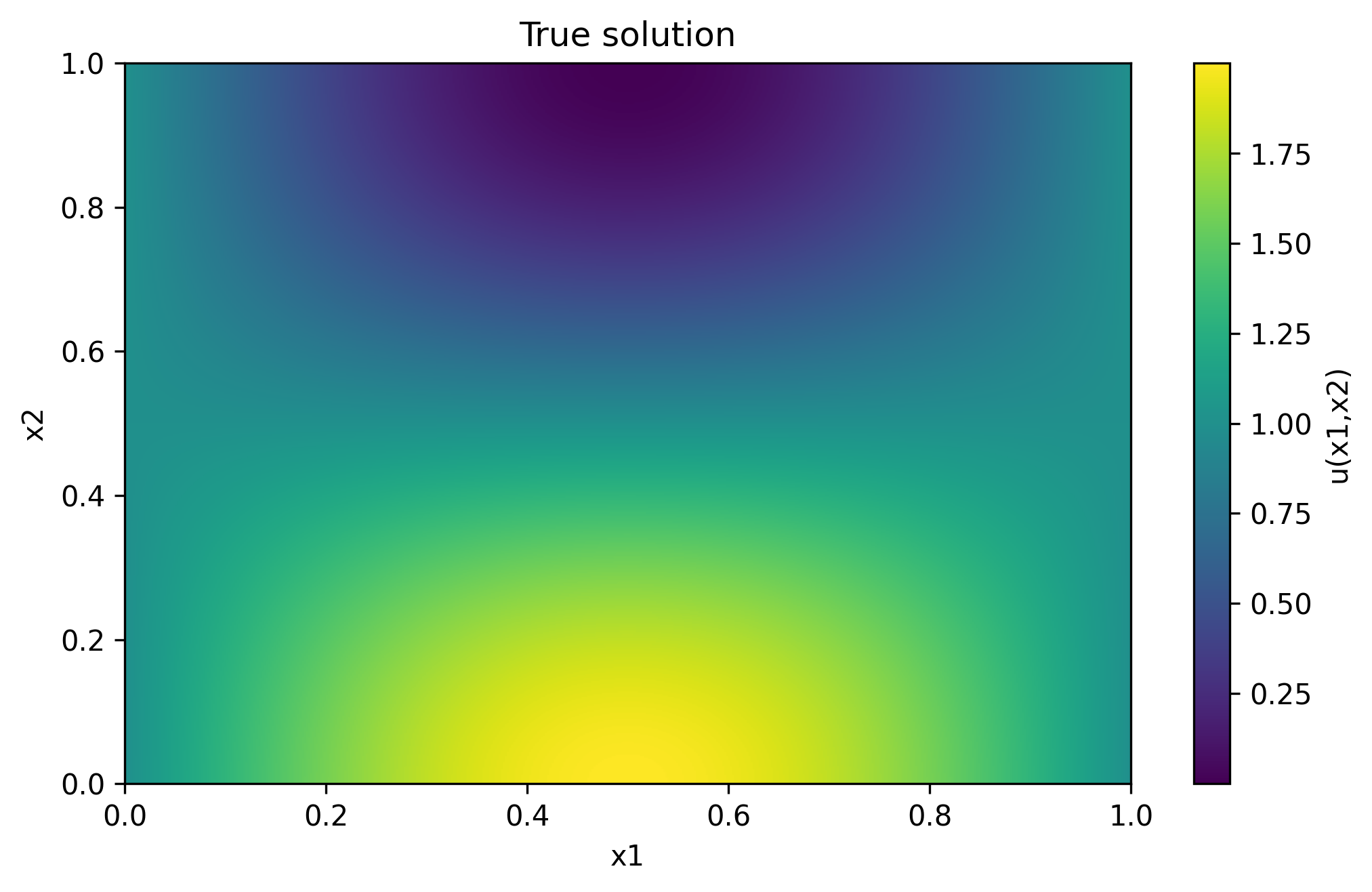}
    \caption{True solution}
    \label{fig:NLP_true}
  \end{subfigure}%
  \hfill
  \begin{subfigure}[t]{0.32\textwidth}
    \centering
    \includegraphics[width=\linewidth]{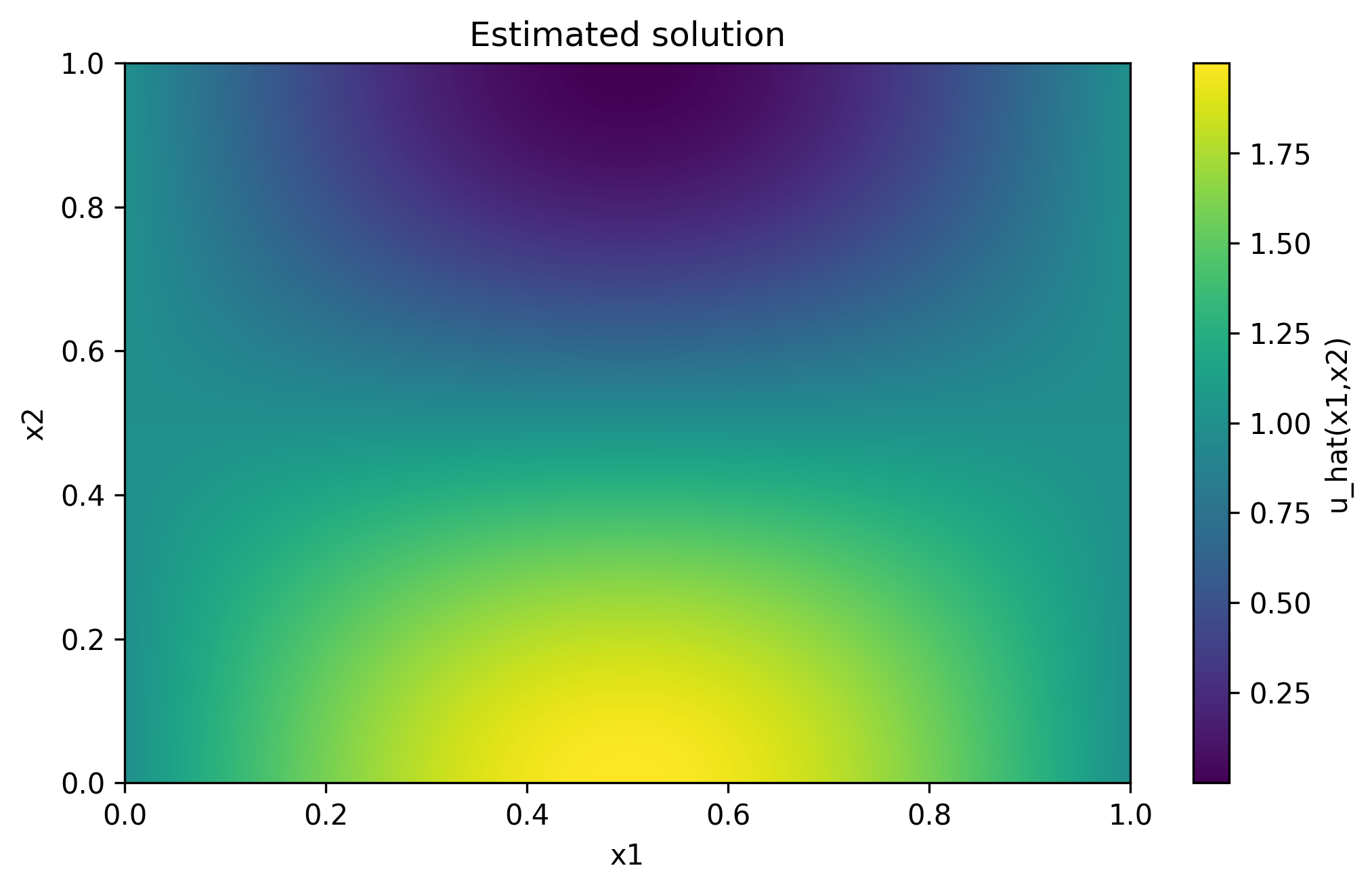}
    \caption{Estimated solution}
    \label{fig:NLP_estimated}
  \end{subfigure}%
  \hfill
  \begin{subfigure}[t]{0.32\textwidth}
    \centering
    \includegraphics[width=\linewidth]{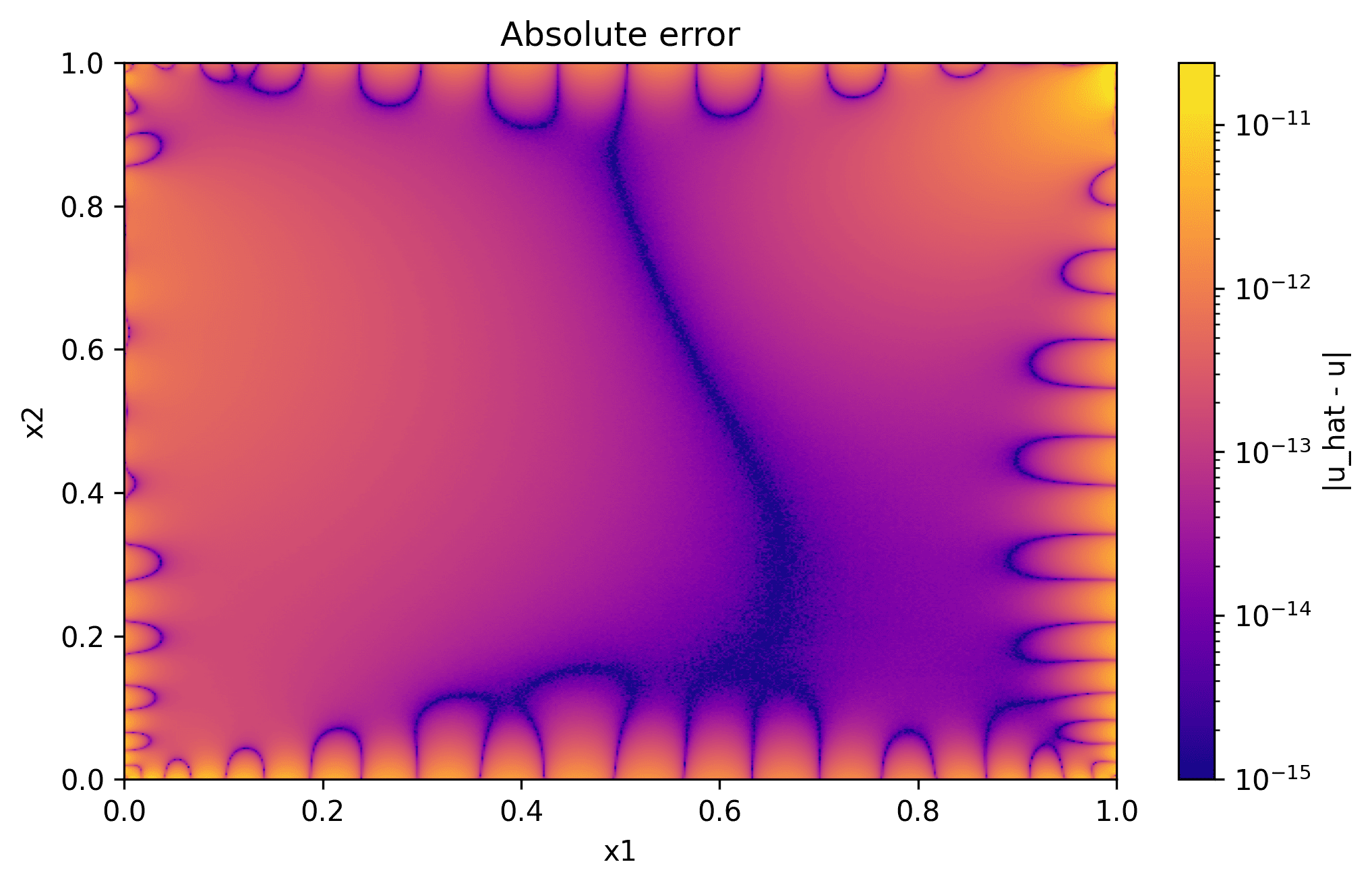}
    \caption{Error}
    \label{fig:NLP_error}
  \end{subfigure}
  \caption{Results for the Nonlinear Poisson equation (cutoff $10^{-4}$).}
  \label{fig:nlp_results}
\end{figure}

\subsection{Allen-Cahn Equation}

\begin{figure}[H]
  \centering
  \includegraphics[width=\linewidth]{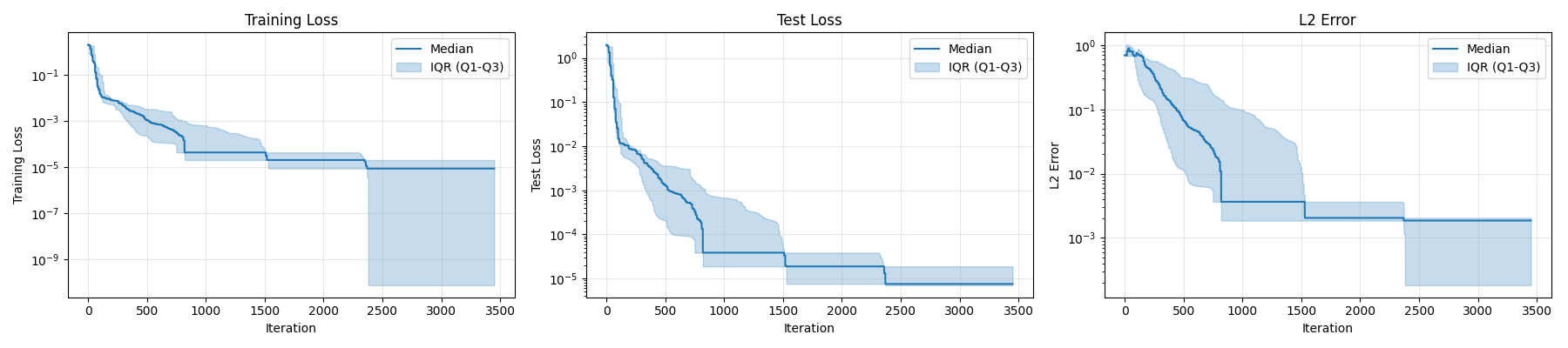}
  \caption{Training curves for the Allen-Cahn equation, showing the evolution of loss and error over iterations.}
  \label{fig:ac_training_curves}
\end{figure}

\begin{figure}[H]
  \centering
  \begin{subfigure}[t]{0.32\textwidth}
    \centering
    \includegraphics[width=\linewidth]{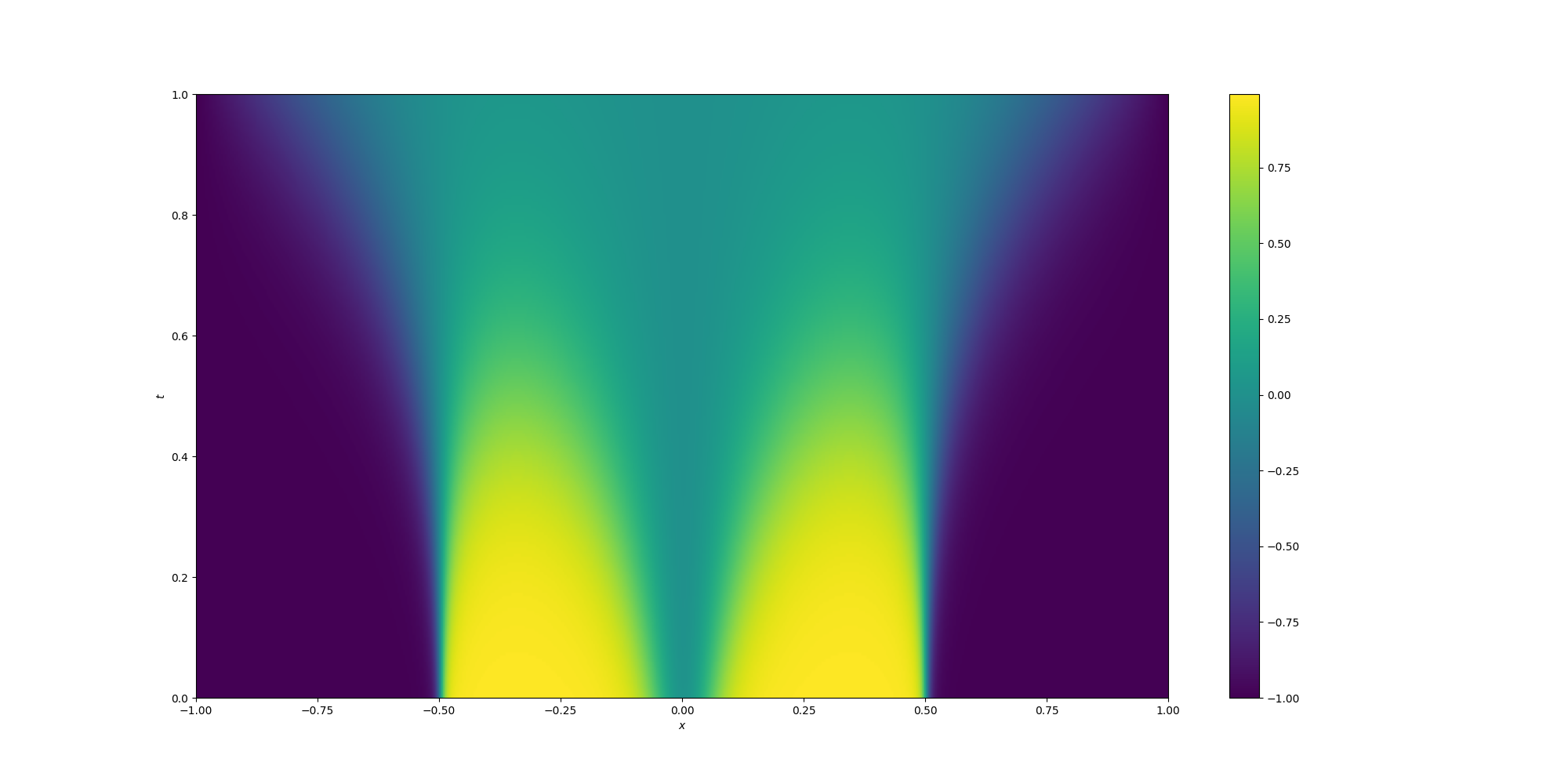}
    \caption{True solution}
    \label{fig:ac_true}
  \end{subfigure}%
  \hfill
  \begin{subfigure}[t]{0.32\textwidth}
    \centering
    \includegraphics[width=\linewidth]{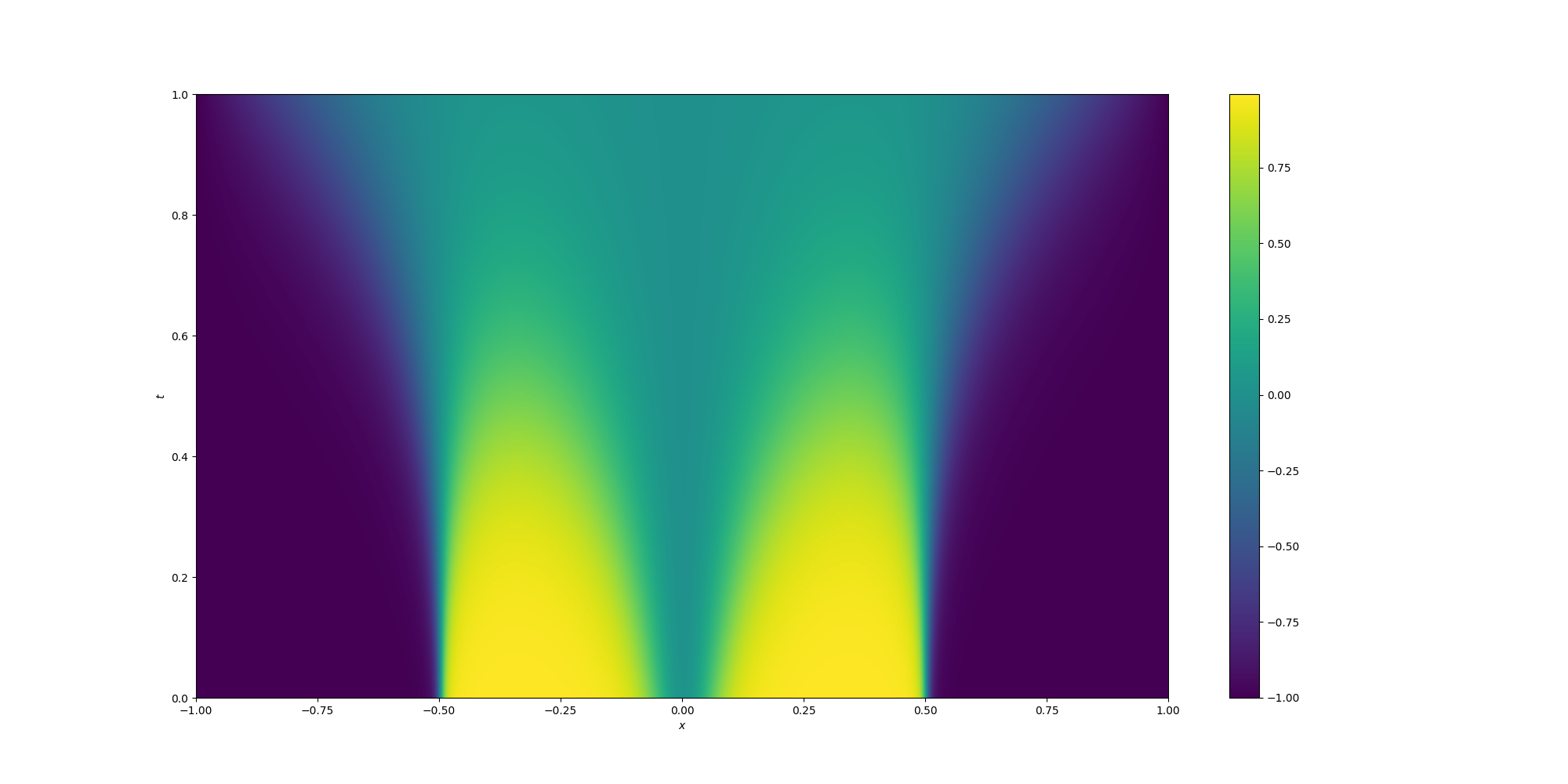}
    \caption{Estimated solution}
    \label{fig:ac_estimated}
  \end{subfigure}%
  \hfill
  \begin{subfigure}[t]{0.32\textwidth}
    \centering
    \includegraphics[width=\linewidth]{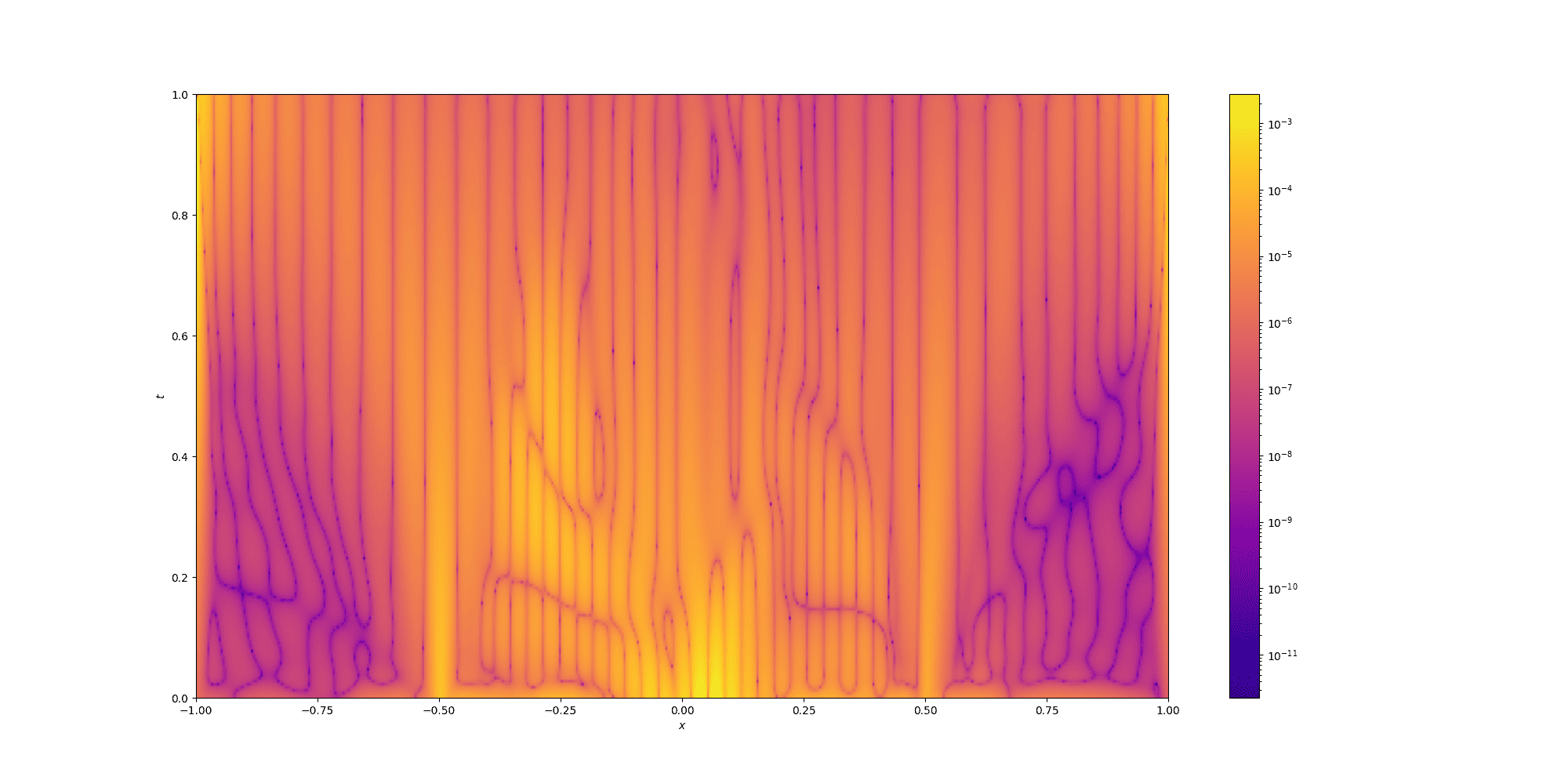}
    \caption{Error}
    \label{fig:ac_error}
  \end{subfigure}
  \caption{Results on the Allen-Cahn equation, showing the error distribution (left), model prediction (middle), and true solution (right). The error is mostly present in regions with the "sharpest" transitions, which exemplifies the challenge of accurately capturing sharp  interfaces still remains even for our advanced optimization approach.}
  \label{fig:ac_results}
\end{figure}

\section{Geometrical interpretation of regularizations}\label{app:geometric-interpretation-regularization}

\subsection{Why Regularization is Necessary}\label{app:why-regularization-is-necessary}

We recall that our goal is to solve the operator equation $D[u] = f$ by minimizing the squared residual
\begin{equation}
  \norm[{D[u] - f}]^2_{\dL^2(\Omega, \mu)}.
\end{equation}
For simplicity, assume $D$ is linear. Then the mapping
\begin{equation}
  u \in \cC^\infty(\Omega) \longmapsto \norm[{D[u]}]_{\dL^2(\Omega, \mu)}
\end{equation}
defines a semi-norm on $\cC^\infty(\Omega)$.
We can ``upgrade’’ this semi-norm into a true norm by introducing the following generalized Sobolev norm:
\begin{equation}\label{eqn:D-norm-definition}
  \norm_{\widetilde{\cH}_D}:\left\{
  \begin{array}{lll}
    \cC^\infty(\Omega \to \RR) & \to     & \RR^+                                                                                         \\
    u                          & \mapsto & \sqrt{\norm[u]^2_{\dL^2(\Omega \to \RR, \mu)} + \norm[D{[u]}]^2_{\dL^2(\Omega \to \RR, \mu)}}
  \end{array}
  \right.
\end{equation}
Clearly, for any $u$,
\begin{equation}\label{eqn:fondamental-inequality-1}
  \norm[{u}]_{\dL^2(\Omega, \mu)} \leq \norm[{u}]_{\widetilde{\cH}_D},
\end{equation}
which guarantees that $\norm_{\widetilde{\cH}_D}$ is \emph{definite}, i.e. $\norm[{u}]_{\widetilde{\cH}_D} = 0 \iff u=0$.

Completing $\cC^\infty(\Omega)$ with respect to $\norm_{\widetilde{\cH}_D}$ yields a generalized Sobolev space $\big(\cH_D,\norm_{\cH_D}\big)$.
This Hilbert space is the largest subspace of $\dL^2(\Omega, \mu)$ on which $D$ is continuous.
Indeed, for every $u \in \cH_D$,
\begin{equation}\label{eqn:fondamental-inequality-2}
  \norm[{D[u]}]_{\dL^2(\Omega, \mu)} \leq \norm[{u}]_{\cH_D}.
\end{equation}

\medskip

Since our goal is to solve $D[u] = f$, we need $D$ to be continuously invertible.
That is, we need the reverse inequality of \cref{eqn:fondamental-inequality-2} to hold (up to a constant $\alpha > 0$).
Formally, if $D$ were algebraically invertible (bijective as a mapping), this condition would read:
\begin{align}
  \label{eqn:chimeric-inequality}
  \begin{split}
           & \left( \exists \alpha > 0,\, \forall u \in \cH_D,\,
    \norm[u]_{\cH_D} \leq \alpha \norm[D{[u]}]_{\dL^2(\Omega \to \RR, \mu)} \right)                \\
    \iff\  & \left( \exists \alpha > 0,\, \forall u \in \cH_D,\,
    \norm[D^{-1}{[D{[u]}]}]_{\cH_D} \leq \alpha \norm[D{[u]}]_{\dL^2(\Omega \to \RR, \mu)} \right) \\
    \iff\  & \left( \exists \alpha > 0,\, \forall f \in \dL^2(\Omega \to \RR, \mu),\,
    \norm[D^{-1}{[f]}]_{\cH_D} \leq \alpha \norm[f]_{\dL^2(\Omega \to \RR, \mu)}
    \right)
  \end{split}.
\end{align}

\paragraph{Operator ill-conditioning.}
Even if $D$ is bijective, \cref{eqn:chimeric-inequality} may fail to hold, i.e. $D$ can be ill-conditioned.
Suppose there exists a subspace $\cH_K \subset \cH_D$ such that $D$ acts compactly on $\cH_K$ with infinite rank.
Then $D$ admits a singular value decomposition \citep[Theorem 15.16]{kressLinearIntegralEquations2014}: for $u \in \cH_K$,
\begin{equation}
  D[u] = \sum_{n \in \NN} e_n \lambda_n \braket{v_n}{u}_{\cH_D},
\end{equation}
with $(v_n)$ orthonormal in $\cH_D$, $(e_n)$ orthonormal in $\dL^2(\Omega,\mu)$, and $\lambda_n \to 0$ as $n \to \infty$.

For \cref{eqn:chimeric-inequality} to hold, we would need $\inf_n \lambda_n > 0$, contradicting $\lambda_n \to 0$.
This is exactly the classical inverse problem setting: $D$ is bijective but ill-conditioned, and regularization is unavoidable.
Among the many schemes developed, Tikhonov regularization is the canonical example \citep{kirschIntroductionMathematicalTheory2021}.

\paragraph{Non-bijectivity.}
If $D$ is not bijective, two additional issues may occur.

\subparagraph{Non-surjectivity.}
If $\Ima D$ is a closed subspace, we can still obtain a solution by replacing the target $f$ with its projection $\Pi_{\Ima D} f$.
Note that minimizing $\norm[{D[u]-f}]^2_{\dL^2(\Omega,\mu)}$ yields precisely this least-squares solution.

\subparagraph{Non-injectivity.}
The lack of injectivity is a much more subtle issue.
Since $D$ is linear and continuous, its null space $\Ker D$ is a closed subspace of $\cH_D$.
In principle, one could restrict the domain of $D$ to $\Ker D^\perp$ to make it injective.
The problem, however, is that identifying $\Ker D$ is typically just as hard as solving the original problem itself, since it amounts to characterizing all $u \in \cH_D$ such that $D[u] = 0$.
Therefore, unless one can rely on theoretical results that explicitly describe $\Ker D$, or construct a subspace $\cH_0 \subset \cH_D$ for which $\Ker D \cap \cH_0$ is explicitly known (so that $D$ can be restricted to $\cH_0$), it is generally impossible to ``get rid of’’ $\Ker D$ in practice.

\medskip

On the other hand, if we do not filter out $\Ker D$, this has the unwanted consequence of introducing ``spurious’’ low-energy signals.
To be concrete, suppose we approximate our solution in a space $\cH_K$ with orthonormal basis $(u_n)_{n \in \NN}$.
Assume there exists a subsequence $(u_n^S) \notin \Ker D$ converging towards $\Ker D$.
Since $\Ker D$ is closed (by continuity of $D$), this means
\begin{equation}
  \lim\limits_{n\to\infty} \norm[\Pi_{\Ker D} u_n^S - u_n^S]_{\cH_D}^2 = 0.
\end{equation}
Equivalently, after extraction, this can be rewritten for all $n \in \NN$ as
\begin{equation}\label{eqn:convenient-approaching-kernel-characterization}
  \frac{\norm[{ \Pi_{\Ker D} u_n^S }]_{\cH_D}^2}{\norm[{u_n^S}]_{\cH_D}^2} \;\geq\; 1 - 2^{-n}.
\end{equation}

Now consider normalized vectors $u_n^S / \norm[{u_n^S}]_{\cH_D}$.
We have
\begin{align}
  \begin{split}
    0 < \| D\!\left[ \tfrac{u_n^S}{\norm[{u_n^S}]_{\cH_D}} \right] \|_{\cH_D}^2
     & = \left\| D\!\left[ \tfrac{\Pi_{\Ker D^\perp} u_n^S + \Pi_{\Ker D} u_n^S}{\norm[{u_n^S}]_{\cH_D}} \right] \right\|_{\cH_D}^2 \\
     & = \Bigg\|
    D\!\left[ \tfrac{\Pi_{\Ker D^\perp} u_n^S}{\norm[{u_n^S}]_{\cH_D}} \right]
    + \underbracket{D\!\left[ \tfrac{\Pi_{\Ker D} u_n^S}{\norm[{u_n^S}]_{\cH_D}} \right]}_{=0}
    \Bigg\|_{\cH_D}^2                                                                                                               \\
     & = \left\| D\!\left[ \tfrac{\Pi_{\Ker D^\perp} u_n^S}{\norm[{u_n^S}]_{\cH_D}} \right] \right\|_{\cH_D}^2                      \\
     & \stackrel{(\ref{eqn:fondamental-inequality-2})}{\leq}
    \left\| \tfrac{\Pi_{\Ker D^\perp} u_n^S}{\norm[{u_n^S}]_{\cH_D}} \right\|_{\cH_D}^2                                             \\
     & = 1 - \frac{\|\Pi_{\Ker D} u_n^S\|_{\cH_D}^2}{\norm[{u_n^S}]_{\cH_D}^2}                                                      \\
     & \stackrel{(\ref{eqn:convenient-approaching-kernel-characterization})}{\leq} 2^{-n}.
  \end{split}
\end{align}

In particular, if (for simplicity) the normalized $(u_n/\norm[{u_n}]_{\cH_D})$ are right singular vectors of $D$, then the vectors $\big(u_n^S/\norm[{u_n^S}]_{\cH_D}\big)$ will correspond to singular values vanishing at least as fast as $(2^{-n})$.
Crucially, however, these vanishing singular values do not reflect an intrinsic ill-conditioning of $D$, but rather an \emph{artificial} ill-conditioning induced by the choice of approximation space $\cH_K$.
In other words, the spurious instability arises from how we approximate the operator, not from the operator itself.
For more details on this approximation-induced phenomenon, see \citet{adcockFramesNumericalApproximation2019,adcockFramesNumericalApproximation2020}.

\medskip

These remarks highlight the \emph{inevitable need for regularization} in practice.
In the next section, we will provide a geometric interpretation of the two regularization schemes introduced in \cref{sec:regularization-nat-grad}, emphasizing how fundamentally different they are in nature.

\begin{Rk}
  The above discussion becomes even more critical when we restrict ourselves to a finite-dimensional approximation space $\cH_\text{app} \subset \cH_D$.
  In this case, the restriction $D_\text{app}$ is automatically compact, since it is of finite rank.
  As a consequence, both types of ill-conditioning described above may occur simultaneously.
  This highlights once again why regularization is not merely convenient but \emph{unavoidable} in numerical practice.
\end{Rk}

\subsection{Ridge-regression}
\label{app:ridge-regression}

Returning to the definition given in \cref{sec:regularization-nat-grad}, recall that Ridge regression amounts to adding $\alpha^2 I_d$ (for some $\alpha > 0$) to the Gram matrix $G_\vtheta$ introduced in \cref{eqn:matrix-formula-natural-gradient-update}:
\begin{align*}
  \literalref{eqn:matrix-formula-natural-gradient-update}.
  \tag{\ref{eqn:matrix-formula-natural-gradient-update}}
\end{align*}
We can reformulate this observation in the following way: given our model
\begin{equation}
  u : \RR^P \to \dL^2(\Omega \to \RR,\mu),
\end{equation}
consider the \emph{regularized model}
\begin{equation}
  u^\alpha:\left\{
  \begin{array}{lll}
    \RR^P   & \to     & \dL^2(\Omega,\mu)\times \RR^P   \\
    \vtheta & \mapsto & (u_{\vtheta}, \alpha\,\vtheta).
  \end{array}\right.
\end{equation}
The Gram matrix of this regularized model is exactly $G_{\vtheta}+\alpha^2 I_d$.
Suppose further that regression is performed with respect to some function $f\in\dL^2(\Omega,\mu)$. Then we must adapt the objective to the regularized model, replacing $f$ with the pair
\begin{equation}
  (f,\alpha\,\vtheta)\in \dL^2(\Omega,\mu)\times\RR^P.
\end{equation}
A straightforward computation shows that, for all $1\leq p\leq \min(P,S)$,
\begin{align}
  \begin{split}
    \braket{\partial_p u_{\vtheta}^\alpha}{(f,\alpha\,\vtheta)-u^\alpha_{\vtheta}}_{\dL^2(\Omega,\mu)\times\RR^P}
     & =\braket{\partial_p u_{\vtheta}}{f-u_{\vtheta}}_{\dL^2(\Omega,\mu)}
    +
    \underbracket{\alpha \braket{\ve^{(p)}}{\vtheta-\vtheta}_{\RR^P}}_{=0}
    \\
     & =\braket{\partial_p u_{\vtheta}}{f-u_{\vtheta}}_{\dL^2(\Omega,\mu)}.
  \end{split}
\end{align}
Thus, regression of $(f,\alpha\,\vtheta)$ with the regularized model is exactly equivalent to Ridge regression.
Equivalently, Ridge regression corresponds to replacing the original model $u$ by the regularized model $u^\alpha$, and replacing the objective $f$ by $(f,\alpha\,\vtheta)$.
From this point of view, the choice of $\alpha\,\vtheta$ as the secondary target may be interpreted as a \emph{default assumption} in the absence of prior information on the parameters: one simply uses the current parameters as a reference target.

We can now extract several fundamental facts:

\begin{enumerate}
  \item As $\alpha \to 0$, the regularized model $u^\alpha$ tends in operator norm to the unregularized model $(u,0)$ (i.e. $u$ by abuse of notation). Indeed,
        \begin{equation}
          \sup_{\norm[\vtheta]_{\RR^P}=1}\norm[\dd u_{\vtheta}^\alpha-(\dd u_{\vtheta},0)]_{\dL^2(\Omega,\mu)\times\RR^P}
          =\alpha\sup_{\norm[\vtheta]_{\RR^P}=1}\norm[\vtheta]_{\RR^P}=\alpha.
        \end{equation}

  \item The model $u^\alpha$ is injective and continuous. Since $\dd u_{\vtheta}$ is continuous (as $\RR^P$ is finite-dimensional), the only possible source of non-injectivity is $\Ker \dd u^\alpha_{\vtheta}$. But
        \begin{equation}
          \Ker \dd u^\alpha_{\vtheta} = \Ker \dd u_{\vtheta}\cap \Ker(\alpha I_{\RR^P}) \subset \Ker(\alpha I_{\RR^P}) = \{0\},
        \end{equation}
        hence injectivity. Restricting $u^\alpha$ to its image makes it algebraically bijective, and the inverse is continuous since
        \begin{equation}
          \alpha \norm[\vtheta]_{\RR^P}\leq \norm[\dd u^\alpha_\vtheta]_{\dL^2(\Omega,\mu)\times\RR^P}.
        \end{equation}
        By the equivalence stated in \cref{eqn:chimeric-inequality}, this implies that $\left(\dd u^\alpha_\vtheta\right)^{-1}$ is continuous. Consequently, $\Ima \dd u^\alpha_\vtheta$ is closed in $\dL^2(\Omega,\mu)\times\RR^P$, since it is the inverse image of a closed set under $\left(\dd u^\alpha_\vtheta\right)^{-1}$. Therefore least-squares solution is well-defined.

  \item The least-squares solution of $u^\alpha=(f,0)$ is influenced by $\alpha$ as follows: 
        $(f,0)$ is projected onto
        \begin{equation}
          \Ima \dd u^\alpha_{\vtheta}=\Span\big((\partial_p u_{\vtheta}, \alpha e^{(p)}) : 1\leq p\leq P\big).
        \end{equation}
        In particular, even if $f\in\Ima \dd u_{\vtheta}$ and $f\neq 0$, we still have $(f,0)\not\in\Ima \dd u^\alpha_{\vtheta}$ (since $\dd u_{\vtheta}(0)=0$). Consequently,
        \begin{equation}
          \left(\Pi^\bot_{\Ima \dd u^\alpha_{\vtheta}} (f,0)\right)_1 \neq f,
        \end{equation}
        where the subscript $1$ denotes projection onto the first component in $\dL^2(\Omega,\mu)\times\RR^P$.
\end{enumerate}

We illustrate these phenomena in \cref{fig:ridge-regression-params}.

Building on the above analysis, we now show that Ridge regression can be extended to the functional setting.
To this end, let us reconsider the operator $D:\cH_D \to \dL^2(\Omega,\mu)$ introduced in \cref{app:why-regularization-is-necessary}.
Analogously to what we did for the parametric model $u$, we define the \emph{regularized operator} at level $\alpha>0$ as
\begin{equation}
  D^\alpha:\left\{
  \begin{array}{lll}
    \cH_D & \to     & \dL^2(\Omega,\mu)\times \cH_D \\
    u     & \mapsto & (D[u], \alpha u)
  \end{array}\right..
\end{equation}
The corresponding target becomes the \emph{regularized objective} $(f,\alpha u)$.

At this level of generality, the equivalence with Gram-matrix regularization no longer holds, since we are dealing with infinite-dimensional operators for which no direct Gram-matrix representation exists.
Nevertheless, the fundamental properties remain valid, namely:

\begin{enumerate}
  \item When $\alpha \to 0$, the regularized operator $D^\alpha$ converges to $(D,0)$ in the operator-norm sense, i.e.\ to $D$ by a mild abuse of notation. Indeed, we have
        \begin{equation}
          \sup_{\norm[u]_{\cH_D}=1}\norm[D^\alpha{[u]}-(D,0)]_{\dL^2(\Omega,\mu)\times\cH_D}
          =\alpha\sup_{\norm[u]_{\cH_D}=1}\norm[u]_{\cH_D}=\alpha.
        \end{equation}

  \item The operator $D^\alpha$ is injective and continuous.
        Indeed, $D$ is continuous by the very construction of $\cH_D$ (see \cref{app:why-regularization-is-necessary}), and injectivity follows since
        \begin{equation}
          \Ker D^\alpha = \Ker D \cap \Ker(\alpha I_{\cH_D}) \subseteq \Ker(\alpha I_{\cH_D}) = \{0\}.
        \end{equation}
        Restricting $D^\alpha$ to its image makes it algebraically bijective, and the inverse is continuous: we have $\alpha \norm[u]_{\cH_D}\leq \norm[D^\alpha{[u]}]_{\dL^2(\Omega,\mu)\times\cH_D}$, which by the equivalence in \cref{eqn:chimeric-inequality} implies that $\big(D^\alpha\big)^{-1}$ is continuous.
        Consequently, $\Ima D^\alpha$ is closed in $\dL^2(\Omega,\mu)\times\cH_D$, since it is the inverse image of a closed set under $\left(D^\alpha\right)^{-1}$. Therefore least-squares solution is well-defined.

  \item Least-squares solutions of the regularized problem $D^\alpha[u]=(f,0)$ are impacted by $\alpha$ in the following way: 
        we are projecting $(f,0)$ onto
        \begin{equation}
          \Ima D^\alpha = \Span\Big(\,(D[h], \alpha h)\,:\,h\in\cH_D\,\Big).
        \end{equation}
        In particular, even if $f\in \Ima D$ with $f\neq 0$, we have $(f,0)\notin\Ima D^\alpha$ (since $D[0]=0$), and hence
        \begin{equation}
          \big(\Pi^\bot_{\Ima D^\alpha}(f,0)\big)_1 \neq f,
        \end{equation}
        where the subscript $1$ denotes the first coordinate in $\dL^2(\Omega,\mu)\times \cH_D$.
\end{enumerate}

We illustrate these phenomena in \cref{fig:ridge-regression-functional}.

\medskip

In summary, Ridge regression can be interpreted as a modification of the operator $D$, rendering it injective and continuously invertible on its image.
However, this comes at a price: the regularized solutions are \emph{never} exact solutions of the original equation $D[u]=f$, even when $\alpha$ is arbitrarily small, since we are in fact solving a different operator equation.
This marks a fundamental distinction from cutoff regularization, which instead acts directly on the approximation space, as we shall see in the next section.

\begin{figure}[H]
  \centering
  \captionsetup[subfigure]{width=.95\linewidth}
  \begin{subfigure}{.49\textwidth}
    \centering
    \tikzsetnextfilename{ridge-regression-params}
    \resizebox{\textwidth}{!}{
\def\a{0.4}         
\def\Ds{.6}             
\def\u{2}           
\def\f{2.5}           
\pgfmathsetmacro{\ang}{atan(\a)}  
\def\R{1}         

\def\xmin{-1} \def\xmax{3}
\def\ymin{-.75} \def\ymax{2}
\def\zmin{-1} \def\zmax{3}
\def\sf{.1} 
\def\pstart{.5} 

\def\Dalpha{\mathrm{d}u^\alpha_\theta}
\def\Draw{\mathrm{d}u_\theta}
\def\Pspace{\mathbb{R}^P}
\def\Ospace{\mathrm{L}^2(\Omega,\mu)}

\tdplotsetmaincoords{70}{-45}

\begin{tikzpicture}[tdplot_main_coords, line cap=round, line join=round,scale=2.5]
\colorlet{colBlue}{blue!60!black}
\definecolor{colPurple}{rgb}{0.8,0.4,1}
\definecolor{colMagenta}{RGB}{255,50,50}

\draw[-{Latex[length=1mm]},green!70!black]   (\xmin,0,0) -- (\xmax,0,0) node[below=0.03\textwidth, left] {\scalebox{0.8}{$\Pspace$}};
\draw[-{Latex[length=1mm]},blue] (0,\ymin,0) -- (0,\ymax,0) node[below=0.03\textwidth, right] {\scalebox{0.8}{$\Pspace$}};
\draw[-{Latex[length=1mm]},blue]  (0,0,\zmin) -- (0,0,\zmax) node[above] {\scalebox{0.8}{$\Ospace$}};


\draw[black]
  ({\xmin+\pstart}, 0, {(\xmin+\pstart)*\Ds}) -- (\xmax, 0, {\xmax*\Ds})
  node[pos=1, below left, sloped] {\scalebox{0.8}{$\Gamma_{\Draw}$}};

\draw[black]
  ({\xmin+\pstart}, {(\xmin+\pstart)*\a}, 0) -- (\xmax, {\xmax*\a}, 0)
  node[pos=1, right, sloped] {\scalebox{0.8}{$\Gamma_{\alpha I_{\Pspace}}$}};

\draw[black]
  ({\xmin+\pstart}, {(\xmin+\pstart)*\a}, {(\xmin+\pstart)*\Ds}) -- (\xmax, {\xmax*\a}, {\xmax*\Ds})
  node[pos=1, right, sloped] {\scalebox{0.8}{$\Gamma_{\Dalpha}$}};

\newcommand{\setfprojcoor}[3]{%
  \pgfmathsetmacro{\ycoord}{#2*#3/(1+#2*#2)}%
  \pgfmathsetmacro{\zcoord}{#2*#2*#3/(1+#2*#2)}%
  \coordinate (#1) at (0,\ycoord,\zcoord);%
}

\def\coeffProj{\Ds/\a}

\draw[black, dashed]
  (0, {(\ymin+\pstart)}, {(\ymin+\pstart)*\coeffProj}) -- (0, {(\ymax-\sf)}, {(\ymax-\sf)*\coeffProj})
  node[pos=1, above right, sloped] {\scalebox{0.8}{$\mathrm{Im}\Dalpha$}};


\fill[colBlue] (\u,0,0) circle[radius=.002\textwidth]
  node[below right] {\scalebox{0.8}{$h$}};
\fill[colBlue] (\u,0,{\Ds*\u}) circle[radius=.002\textwidth]
  node[right] {\scalebox{0.8}{$(h,\Draw[h])$}};
\fill[colBlue] (\u,{\a*\u},0) circle[radius=.002\textwidth]
  node[above=0.005\textwidth, right=0.02\textwidth] {\scalebox{0.8}{$(h,\alpha I_{\Pspace}[h])$}};
\fill[colBlue] (\u,{\a*\u},{\Ds*\u}) circle[radius=.002\textwidth]
  node[left] {\scalebox{0.8}{$(h,\Dalpha[h])$}};

\fill[colBlue] (0,{\a*\u},0)
circle[radius=.002\textwidth]
  node[below left] {\scalebox{0.8}{$\alpha I_{\Pspace}[h]$}};
\fill[colBlue] (0,0,{\Ds*\u}) circle[radius=.002\textwidth]
  node[right] {\scalebox{0.8}{$\Draw[h]$}};
\fill[colBlue] (0,{\a*\u},{\Ds*\u}) circle[radius=.002\textwidth]
  node[left] {\scalebox{0.8}{$\Dalpha[h]$}};

\fill[colMagenta] (0,0,\f) circle[radius=.002\textwidth]
  node[right] {\scalebox{0.8}{$f$}};

\setfprojcoor{projPoint1}{\coeffProj}{\f}

\fill[colMagenta] (projPoint1) circle[radius=.002\textwidth]
  node[left] {\scalebox{0.8}{$\Pi_{\mathrm{Im}\Dalpha} f$}};

\draw[draw opacity=.5, dashed, colMagenta, -{Latex[length=1mm]}]
  (0, 0, \f) -- (projPoint1)
  node[pos=0.5, below, sloped, opacity=.5] {\scalebox{0.8}{$\Pi_{\mathrm{Im} \Dalpha}$}};

\setfprojcoor{projPoint15}{\Ds*3.75}{\f}
\setfprojcoor{projPoint2}{\Ds*5}{\f}
\setfprojcoor{projPoint3}{\Ds*10}{\f}
\setfprojcoor{projPoint4}{\Ds*20}{\f}
\setfprojcoor{projPoint5}{\Ds*40}{\f}

\draw[gray!70, dashed, -{Latex[length=1mm]},
      ]
  plot[smooth, tension=0.5] coordinates {
    (projPoint1)
    (projPoint15)
    (projPoint2)
    (projPoint3)
    (projPoint4)
    (projPoint5)
    (0,0,\f)};
  \node[gray!70, above left] at (projPoint3)
  {\scalebox{0.8}{$\alpha\to 0$}};



\path[fill=colPurple, fill opacity=0.75, draw=colPurple]
  (0,0,0) -- (\R,0,0)
  arc[start angle=0, end angle=\ang, radius=\R] -- cycle
  node[pos=0.5, colPurple, below=0.01\textwidth, right=0.06\textwidth] {\scalebox{0.8}{$\arctan(\alpha)$}};

\fill[colBlue] (0,0,0) circle[radius=.002\textwidth]
 node[below=.02\textwidth, right=.005\textwidth] {$0$};

\draw[densely dotted, -{Latex[length=1mm]}, gray!70]
  (\u,{\a*\u},{\Ds*\u}) -- (\u,0,{\Ds*\u})
  node[midway, above, sloped] {\scalebox{0.8}{$\alpha\to 0$}};

\draw[densely dotted, -{Latex[length=1mm]}, gray!70]
  (\u,{\a*\u},0) -- (\u,0,0)
  node[midway, below=-0.005\textwidth, sloped] {\scalebox{0.8}{$\alpha\to 0$}};

\draw[densely dotted, -{Latex[length=1mm]}, gray!70]
  (0,{\a*\u},{\Ds*\u}) -- (0,0,{\Ds*\u})
  node[midway, above, sloped] {\scalebox{0.8}{$\alpha\to 0$}};

\draw[densely dotted, -{Latex[length=1mm]}, gray!70]
  (0,{\a*\u},0) -- (0,0,0)
  node[midway, below=-0.005\textwidth, sloped] {\scalebox{0.8}{$\alpha\to 0$}};


\path[fill=blue!50, fill opacity=0.1]
  (0,{\ymin+\sf},{\zmin+\sf}) -- 
  (0,{\ymax-\sf},{\zmin+\sf}) -- 
  (0,{\ymax-\sf},{\zmax-\sf}) -- 
  (0,{\ymin+\sf},{\zmax-\sf}) -- 
  cycle;

\end{tikzpicture}}
    \caption{\textbf{Illustration of parametric Ridge regression.}\\
      The green region represents the solution space, while the blue regions denote the target spaces.
      As $\alpha \to 0$, the regularized graph $\Gamma_{\dd u_\vtheta^\alpha}$ of $\dd u_\vtheta^\alpha$ approaches the graph $\Gamma_{\dd u_\vtheta}$ of $\dd u_\vtheta$, with the angle between them vanishing at rate $\arctan(\alpha)$.
      The key consequence is that the projection of the objective $f$ onto $\Ima \dd u_\vtheta^\alpha$ follows a non-linear path as $\alpha \to 0$, coinciding with $\Pi_{\Ima \dd u_\vtheta} f$ only asymptotically.}
    \label{fig:ridge-regression-params}
  \end{subfigure}
  \begin{subfigure}{0.49\textwidth}
    \centering
    \tikzsetnextfilename{ridge-regression-functional}
    \resizebox{\textwidth}{!}{
\def\a{0.4}         
\def\Ds{.6}             
\def\u{2}           
\def\f{2.5}           
\pgfmathsetmacro{\ang}{atan(\a)}  
\def\R{1}         

\def\xmin{-1} \def\xmax{3}
\def\ymin{-.75} \def\ymax{2}
\def\zmin{-1} \def\zmax{3}
\def\sf{.1} 
\def\pstart{.5} 

\def\Dalpha{D^\alpha}
\def\Draw{D}
\def\Pspace{\mathcal{H}_D}
\def\Ospace{\mathrm{L}^2(\Omega,\mu)}

\tdplotsetmaincoords{70}{-45}

\begin{tikzpicture}[tdplot_main_coords, line cap=round, line join=round,scale=2.5]
\colorlet{colBlue}{blue!60!black}
\definecolor{colPurple}{rgb}{0.8,0.4,1}
\definecolor{colMagenta}{RGB}{255,50,50}

\draw[-{Latex[length=1mm]},green!70!black]   (\xmin,0,0) -- (\xmax,0,0) node[below=0.03\textwidth, left] {\scalebox{0.8}{$\Pspace$}};
\draw[-{Latex[length=1mm]},blue] (0,\ymin,0) -- (0,\ymax,0) node[below=0.03\textwidth, right] {\scalebox{0.8}{$\Pspace$}};
\draw[-{Latex[length=1mm]},blue]  (0,0,\zmin) -- (0,0,\zmax) node[above] {\scalebox{0.8}{$\Ospace$}};


\draw[black]
  ({\xmin+\pstart}, 0, {(\xmin+\pstart)*\Ds}) -- (\xmax, 0, {\xmax*\Ds})
  node[pos=1, below left, sloped] {\scalebox{0.8}{$\Gamma_{\Draw}$}};

\draw[black]
  ({\xmin+\pstart}, {(\xmin+\pstart)*\a}, 0) -- (\xmax, {\xmax*\a}, 0)
  node[pos=1, right, sloped] {\scalebox{0.8}{$\Gamma_{\alpha I_{\Pspace}}$}};

\draw[black]
  ({\xmin+\pstart}, {(\xmin+\pstart)*\a}, {(\xmin+\pstart)*\Ds}) -- (\xmax, {\xmax*\a}, {\xmax*\Ds})
  node[pos=1, right, sloped] {\scalebox{0.8}{$\Gamma_{\Dalpha}$}};

\newcommand{\setfprojcoor}[3]{%
  \pgfmathsetmacro{\ycoord}{#2*#3/(1+#2*#2)}%
  \pgfmathsetmacro{\zcoord}{#2*#2*#3/(1+#2*#2)}%
  \coordinate (#1) at (0,\ycoord,\zcoord);%
}

\def\coeffProj{\Ds/\a}

\draw[black, dashed]
  (0, {(\ymin+\pstart)}, {(\ymin+\pstart)*\coeffProj}) -- (0, {(\ymax-\sf)}, {(\ymax-\sf)*\coeffProj})
  node[pos=1, above right, sloped] {\scalebox{0.8}{$\mathrm{Im}\Dalpha$}};


\fill[colBlue] (\u,0,0) circle[radius=.002\textwidth]
  node[below right] {\scalebox{0.8}{$h$}};
\fill[colBlue] (\u,0,{\Ds*\u}) circle[radius=.002\textwidth]
  node[right] {\scalebox{0.8}{$(h,\Draw[h])$}};
\fill[colBlue] (\u,{\a*\u},0) circle[radius=.002\textwidth]
  node[above=0.005\textwidth, right=0.02\textwidth] {\scalebox{0.8}{$(h,\alpha I_{\Pspace}[h])$}};
\fill[colBlue] (\u,{\a*\u},{\Ds*\u}) circle[radius=.002\textwidth]
  node[left] {\scalebox{0.8}{$(h,\Dalpha[h])$}};

\fill[colBlue] (0,{\a*\u},0)
circle[radius=.002\textwidth]
  node[below left] {\scalebox{0.8}{$\alpha I_{\Pspace}[h]$}};
\fill[colBlue] (0,0,{\Ds*\u}) circle[radius=.002\textwidth]
  node[right] {\scalebox{0.8}{$\Draw[h]$}};
\fill[colBlue] (0,{\a*\u},{\Ds*\u}) circle[radius=.002\textwidth]
  node[left] {\scalebox{0.8}{$\Dalpha[h]$}};

\fill[colMagenta] (0,0,\f) circle[radius=.002\textwidth]
  node[right] {\scalebox{0.8}{$f$}};

\setfprojcoor{projPoint1}{\coeffProj}{\f}

\fill[colMagenta] (projPoint1) circle[radius=.002\textwidth]
  node[left] {\scalebox{0.8}{$\Pi_{\mathrm{Im}\Dalpha} f$}};

\draw[draw opacity=.5, dashed, colMagenta, -{Latex[length=1mm]}]
  (0, 0, \f) -- (projPoint1)
  node[pos=0.5, below, sloped, opacity=.5] {\scalebox{0.8}{$\Pi_{\mathrm{Im} \Dalpha}$}};

\setfprojcoor{projPoint15}{\Ds*3.75}{\f}
\setfprojcoor{projPoint2}{\Ds*5}{\f}
\setfprojcoor{projPoint3}{\Ds*10}{\f}
\setfprojcoor{projPoint4}{\Ds*20}{\f}
\setfprojcoor{projPoint5}{\Ds*40}{\f}

\draw[gray!70, dashed, -{Latex[length=1mm]},
      ]
  plot[smooth, tension=0.5] coordinates {
    (projPoint1)
    (projPoint15)
    (projPoint2)
    (projPoint3)
    (projPoint4)
    (projPoint5)
    (0,0,\f)};
  \node[gray!70, above left] at (projPoint3)
  {\scalebox{0.8}{$\alpha\to 0$}};



\path[fill=colPurple, fill opacity=0.75, draw=colPurple]
  (0,0,0) -- (\R,0,0)
  arc[start angle=0, end angle=\ang, radius=\R] -- cycle
  node[pos=0.5, colPurple, below=0.01\textwidth, right=0.06\textwidth] {\scalebox{0.8}{$\arctan(\alpha)$}};

\fill[colBlue] (0,0,0) circle[radius=.002\textwidth]
 node[below=.02\textwidth, right=.005\textwidth] {$0$};

\draw[densely dotted, -{Latex[length=1mm]}, gray!70]
  (\u,{\a*\u},{\Ds*\u}) -- (\u,0,{\Ds*\u})
  node[midway, above, sloped] {\scalebox{0.8}{$\alpha\to 0$}};

\draw[densely dotted, -{Latex[length=1mm]}, gray!70]
  (\u,{\a*\u},0) -- (\u,0,0)
  node[midway, below=-0.005\textwidth, sloped] {\scalebox{0.8}{$\alpha\to 0$}};

\draw[densely dotted, -{Latex[length=1mm]}, gray!70]
  (0,{\a*\u},{\Ds*\u}) -- (0,0,{\Ds*\u})
  node[midway, above, sloped] {\scalebox{0.8}{$\alpha\to 0$}};

\draw[densely dotted, -{Latex[length=1mm]}, gray!70]
  (0,{\a*\u},0) -- (0,0,0)
  node[midway, below=-0.005\textwidth, sloped] {\scalebox{0.8}{$\alpha\to 0$}};


\path[fill=blue!50, fill opacity=0.1]
  (0,{\ymin+\sf},{\zmin+\sf}) -- 
  (0,{\ymax-\sf},{\zmin+\sf}) -- 
  (0,{\ymax-\sf},{\zmax-\sf}) -- 
  (0,{\ymin+\sf},{\zmax-\sf}) -- 
  cycle;

\end{tikzpicture}}
    \caption{\textbf{Illustration of functional Ridge regression.}\\
      The green region represents the solution space, while the blue regions denote the target spaces.
      As $\alpha \to 0$, the regularized graph $\Gamma_{D^\alpha}$ of $D^\alpha$ approaches the graph $\Gamma_{\cH_D}$ of $D$, with the angle between them vanishing at rate $\arctan(\alpha)$.
      The key consequence is that the projection of the objective $f$ onto $\Ima D^\alpha$ follows a non-linear path as $\alpha \to 0$, coinciding with $\Pi_{\Ima D} f$ only asymptotically.}
    \label{fig:ridge-regression-functional}
  \end{subfigure}
  \caption{Illustrations of Ridge regression.}
  \label{fig:ridge-regression}
\end{figure}

\subsection{Cutoff regression}
\label{app:cutoff-regression}

As in \cref{app:ridge-regression}, let us return to the setting of \cref{sec:regularization-nat-grad}.
In \cref{eqn:cutoff-regularization}, we introduced cutoff regularization from the SVD perspective: given the differential $\dd u_{\vtheta}$ of the model $u$, at the point $\vtheta$, and its singular value decomposition $\dd u_{\vtheta} = \fVsing \fDsing \fUsing^T$, the cutoff-regularized pseudo-inverse $\dd u^{\dagger_\alpha}_{\vtheta}$ at level $\alpha > 0$ is defined as
\begin{align}
  \dd u^{\dagger_\alpha}_{\vtheta} & :=\fUsing \fDsing^{\dagger_\alpha} \fVsing^T\,;
                                   &                                                 
  \fDsi[p]^{\dagger_\alpha}        & :=
  \begin{cases}
    \fDsi[p]^{-1} & \text{if } \fDsi[p] \geq \alpha \\
    0             & \text{otherwise}
  \end{cases},
  \, 1 \leq p \leq P.
\end{align}

\medskip

Let us reinterpret this construction.
Denote by $N_\alpha \in \NN$ the number of singular values larger than $\alpha$.
Equivalently, assuming $(\fDsi[p])_{1\leq p\leq P}$ is non-increasing,
\begin{equation}
  N_\alpha := \argmax_{p\in\NN}\{\, \fDsi[p] \geq \alpha \,\}.
\end{equation}
Define
\begin{equation}
  \Theta_\alpha := \Span\{ \fUsi[p] : 1 \leq p \leq N_\alpha \},
  \qquad
  \TZN[N_\alpha] := \Span\{ \fVsi[p] : 1 \leq p \leq N_\alpha \},
\end{equation}
so that $\TZN[N_\alpha] = \dd u_\vtheta\left( \Theta_\alpha \right)$.
We then have
\begin{equation}\label{eqn:cutoff-reg-inverse}
  \left(\dd u_{\vtheta_{|\Theta_\alpha}}^{|\TZN[N_\alpha]}\right)^{-1} = \dd u^{\dagger_\alpha}_{\vtheta},
\end{equation}
meaning that the restriction $\dd u_{\vtheta}^\alpha := \dd u_{\vtheta_{|\Theta_\alpha}}$ of $\dd u_{\vtheta}$ to the domain $\Theta_\alpha$ becomes invertible once its codomain is restricted to its image $\TZN[N_\alpha]$, with inverse given precisely by the cutoff pseudo-inverse $\dd u^{\dagger_\alpha}_{\vtheta}$.
Moreover, for any $h \in \Theta_\alpha$,
\begin{equation}\label{eqn:chimeric-inequality-on-alpha-set}
  \norm[\dd u_{\vtheta}(h)]_{\dL^2(\Omega,\mu)}
  = \norm[\fVsing \fDsing \fUsing^T h]_{\dL^2(\Omega,\mu)}
  \stackrel{\fVsing\text{ unitary}}{=}
  \norm[\fDsing \fUsing^T h]_{\RR^P}
  \stackrel{h\in \Theta_\alpha}{\geq}
  \alpha \norm[\fUsing^Th]_{\RR^P}
  \stackrel{\fUsing\text{ unitary}}{=}
  \alpha \norm[h]_{\RR^P}.
\end{equation}
In other words, \cref{eqn:chimeric-inequality} is satisfied by $\dd u_{\vtheta}^\alpha$.

Thus, while ridge regularization modifies the model itself, cutoff regularization instead restricts the domain of the model so that, on this restricted domain, \cref{eqn:chimeric-inequality} holds and the model becomes invertible.
We summarize the fundamental properties:

\begin{enumerate}
  \item We have
        \begin{equation}
          \bigcap_{\alpha > 0} \left(\RR^P\backslash\Theta_\alpha\right) = \Ker \dd u_{\vtheta},
        \end{equation}
        that is, $\lim_{\alpha\to 0} \RR^P\backslash\Theta_\alpha = \Ker \dd u_{\vtheta}$, since for all $\alpha>\beta$ we have $\Theta_\alpha \subset \Theta_\beta$ and then $\RR^P\backslash\Theta_\beta \subset \RR^P\backslash\Theta_\alpha$.
        Similarly, $\lim_{\alpha\to 0} \TZN[N_\alpha] = \Ima \dd u_{\vtheta}$.
        Moreover, for each $\alpha>0$, the restriction $\dd u_{\vtheta}^\alpha$ coincides with $\dd u_{\vtheta}$ on $\Theta_\alpha$.

  \item By \cref{eqn:chimeric-inequality-on-alpha-set}, $\dd u_{\vtheta}^\alpha$ is injective and continuous.
        Restricting it to its image $\TZN[N_\alpha]$ makes it bijective and bicontinuous, with inverse exactly the cutoff pseudo-inverse $\dd u^{\dagger_\alpha}_{\vtheta}$.
        In particular $\dd u\left(\Theta_\alpha\right)$ is closed in $\dL^2(\Omega,\mu)$, since it is the inverse image of a closed set under $\dd u^{\dagger_\alpha}_{\vtheta}$. Therefore least-squares solution is well-defined.

  \item Solving the least-squares problem $\dd u_{\vtheta}^\alpha = f$ is now altered in the following way: the target $f$ is first projected onto $\TZN[N_\alpha] = \Ima \dd u_{\vtheta}^\alpha$.
        In particular, if for some $\alpha > 0$ we already have $f \in \Ima \dd u_{\vtheta}^\alpha$, then the regularized least-squares formulation recovers an \emph{exact solution} to the problem. This stands in sharp contrast with Ridge regression, where such exact recovery can only occur \emph{asymptotically} in the limit $\alpha \to 0$.
\end{enumerate}

As in \cref{app:ridge-regression}, we now need to reinterpret the cutoff regularization in order to extend it to the functional setting.
Let us return once more to the operator $D:\cH_D \to \dL^2(\Omega,\mu)$ introduced in \cref{app:why-regularization-is-necessary}.
In general, one cannot define an SVD for such an operator (except when it is compact).
We must therefore appeal to the spectral theorem for bounded self-adjoint operators, which relies on the notion of a \emph{projection-valued measure} (also called a resolution of the identity).
For our purposes, it will be sufficient to simply state the definition.

\begin{Def}[Projection-valued measure]\label{Def:projection-valued-measure}
  Let $(X,\cA)$ be a measurable space, where $\cA$ denotes its $\sigma$-algebra, and let $\cH$ be a Hilbert space.
  A \emph{projection-valued measure} (PVM) is a map
  \begin{equation*}
    \pi : \cA \to \cL_b(\cH\to\cH),
  \end{equation*}
  where $\cL_b(\cH\to\cH)$ denotes the set of bounded operators on $\cH$,
  such that for every $A \in \cA$, $\pi(A)$ is an orthogonal projection on $\cH$, and the following properties hold:
  \begin{enumerate}
    \item $\pi(\emptyset) = 0$ and $\pi(X) = I_{\cH}$, where $I_{\cH}$ is the identity operator on $\cH$;
    \item $\pi(A \cap B) = \pi(A)\pi(B)$ for all $A,B \in \cA$;
    \item For every countable family $(A_i)_{i=1}^\infty$ of disjoint sets in $\cA$,
          \begin{equation*}
            \pi\!\left(\bigcup_{i=1}^\infty A_i\right) = \sum_{i=1}^\infty \pi(A_i),
          \end{equation*}
          where the series converges in the strong operator topology.
  \end{enumerate}
\end{Def}

Since projection-valued measures are measures, one can define integrals with respect to them.
We refer to \cite[Chapter 13]{berezanskyFunctionalAnalysisVol1996} for details.
We may now state the spectral theorem.

\begin{restatable}{Th}{SelfAdjointDiag}
  \label{Th:self-adjoint-op-diag}
  Let $\cH$ be a Hilbert space and let $A:\cH\to\cH$ be a self-adjoint operator.
  Then there exists a projection-valued measure $\pi$ on the Borel $\sigma$-algebra of $\RR$ such that
  \begin{equation}
    A = \int_{\RR} \lambda \,\pi(\dd \lambda) = \int_{\Spectrum(A)} \lambda \,\pi(\dd \lambda),
  \end{equation}
  where $\Spectrum(A)$ denotes the spectrum of $A$.
\end{restatable}

A proof can be found in \cite[Theorem 4.1, Section 4.1, Chapter 13]{berezanskyFunctionalAnalysisVol1996}.
In particular, since $\pi$ is a projection-valued measure, we have by \cref{Def:projection-valued-measure}:
\begin{equation}
  I_{\cH} = \int_{\RR} \pi(\dd \lambda).
\end{equation}

Since $D$ is continuous, we can define its adjoint $D^*:\dL^2(\Omega,\mu)\to\cH_D$, and hence the self-adjoint operator $D^*D:\cH_D\to\cH_D$.
Applying \cref{Th:self-adjoint-op-diag}, we obtain a projection-valued measure $\pi_D$ on $\RR$ endowed with its Borel $\sigma$-algebra, such that
\begin{align}\label{eqn:spectral-decomposition-DstarD}
  D^*D      & = \int_{\RR_+} \lambda \,\pi_D(\dd\lambda), &
  I_{\cH_D} & = \int_{\RR_+} \pi_D(\dd\lambda),
\end{align}
where the integration is restricted to $\RR_+$ since $D^*D$ is a positive operator.
We can then define
\begin{equation}\label{eqn:Pi_HDalpha-def}
  \Pi_D^\alpha := \int_{\alpha^2}^{+\infty} \pi_D(\dd\lambda),
\end{equation}
which is an orthogonal projection in $\cH_D$ since $\pi_D$ is a projection-valued measure.
We then define the regularized space $\cH_D^\alpha$ at level $\alpha > 0$ by
\begin{equation}\label{eqn:HDalpha-def}
  \cH_D^\alpha := \Ima \Pi_D^\alpha \subset \cH_D.
\end{equation}

For any $u \in \cH_D^\alpha$, we compute
\begin{align}
  \begin{split}\label{eqn:coercivity-of-D-in-HDalpha}
    \norm[D{[u]}]^2_{\dL^2(\Omega,\mu)}
     & = \braket{D[u]}{D[u]}_{\dL^2(\Omega,\mu)}
    = \braket{u}{D^*D[u]}_{\cH_D}                                                                                \\
     & = \braket{u}{\int_{\RR_+}\lambda \pi_D(\dd\lambda) u}_{\cH_D}                                             \\
     & \stackrel{u\in\cH_D^\alpha}{=}
    \braket{u}{\int_{\RR_+}\lambda_1 \pi_D(\dd\lambda_1) \int_{\alpha^2}^{+\infty}\pi_D(\dd\lambda_2) u}_{\cH_D} \\
     & \stackrel{\pi_D \text{ PVM}}{=}
    \braket{u}{\int_{\alpha^2}^{+\infty}\lambda \pi_D(\dd\lambda) u}_{\cH_D}                                     \\
     & = \int_{\alpha^2}^{+\infty} \lambda \braket{u}{\pi_D(\dd\lambda) u}_{\cH_D}                               \\
     & \geq \alpha^2 \int_{\alpha^2}^{+\infty} \braket{u}{\pi_D(\dd\lambda) u}_{\cH_D}
    \stackrel{u\in\cH_D^\alpha}{=}
    \alpha^2 \braket{u}{u}_{\cH_D}
    = \alpha^2 \norm[u]_{\cH_D}^2.
  \end{split}
\end{align}
That is,
\begin{equation}
  \norm[D{[u]}]_{\dL^2(\Omega,\mu)} \;\geq\; \alpha \norm[u]_{\cH_D},
\end{equation}
so that \cref{eqn:chimeric-inequality} is verified on $\cH_D^\alpha$.
We denote
\begin{equation}
  D^\alpha := D_{|\cH_D^\alpha},
\end{equation}
the restriction of $D$ to the domain $\cH_D^\alpha$.
We can now list the fundamental properties:
\begin{enumerate}
  \item We have
        \begin{equation}
          \bigcap_{\alpha > 0}\left(\cH_D\backslash\cH_D^\alpha\right) = \Ker D
        \end{equation}
        that is, $\lim_{\alpha\to 0} \cH_D\backslash\cH_D^\alpha = \Ker D$, since for all $\alpha > \beta$, $\cH_D^\alpha \subset \cH_D^\beta$ by Property 3 of \cref{Def:projection-valued-measure}.
        Moreover, by continuity of $D$, we also have $\lim_{\alpha\to 0} D\left[\cH_D^\alpha\right] = \Ima D$.
        Finally, for each $\alpha > 0$, $D^\alpha$ coincides with $D$ on $\cH_D^\alpha$ by construction.
  \item As established by \cref{eqn:chimeric-inequality}, $D^\alpha$ is injective and continuous. When restricted to its image, it is therefore bijective and bicontinuous, hence invertible.
        In particular $D\left[\cH_D^\alpha\right]$ is closed in $\dL^2(\Omega,\mu)$, since it is the inverse image of a closed set under $\left(D^\alpha\right)^{-1}$. Therefore least-squares solution is well-defined.
  \item The least-squares solution of $D^\alpha = f$ is now modified as follows: one projects $f$ onto $\Ima D^\alpha$.
        In particular, if for some $\alpha > 0$ we already have $f \in \Ima D^\alpha$, then the regularized least-squares formulation recovers an \emph{exact solution} to the problem $D[u]=f$. This stands in sharp contrast with Ridge regression, where such exact recovery can only occur \emph{asymptotically} in the limit $\alpha \to 0$.
\end{enumerate}

\subsection{Connection to Green's Function}\label{sec:connection-with-green-functions}

To further highlight the difference between the two regularization schemes, we now reinterpret them through the lens of Green's functions of the operator $D$.
\citet[Theorem~2]{schwencke2025anagram} established in the finite-dimensional case a connection between the natural gradient for PINNs and Green's functions. Their proof relies on \citet[Proposition~3]{schwencke2025anagram}, which will be our starting point. We restate the relevant definitions and results for completeness.

\begin{Def}[{\citealp[Definition 9]{schwencke2025anagram}: generalized Green's function}]
  Let $\cH$ be an Hilbert space, $D:\cH\to\dL^2(\Omega,\mu)$ be a linear differential operator, $\cH_0\subset\cH$ a subspace isometrically embedded in $\cH$ and $f\in\dL^2(\Omega,\mu)$. A generalized Green's function of $D$ on $\cH_0$ is then any kernel function $g:\Omega\times\Omega\to\RR$ such that the operator:
  \begin{equation*}
    R_{\cH_0}:\left\{\begin{array}{lll}
      \dL^2(\Omega\to\RR,\mu) & \to     & \cH                                                         \\
      f                       & \mapsto & \Big(x\in\Omega\mapsto\int_\Omega g(x,s)f(s)\mu(\dd s)\Big)
    \end{array}\right.,
  \end{equation*}
  verifies the equation:
  \begin{equation}\label{eqn:generalized-green-operator-identity}
    D\circ R_{\cH_0} = \Pi^\bot_{D[\cH_0]}
  \end{equation}
\end{Def}
\begin{Prop}[{\citealp[Proposition 3]{schwencke2025anagram}}]
  Let $D:\cH\to\dL^2(\Omega,\mu)$ be a linear differential operator, and $\cH_0:=\Span(u_p : 1\leq p\leq P)\subset\cH$ a subspace isometrically embedded in $\cH$. Then the generalized Green's function of $D$ on $\cH_0$ is given by: $\forallt x,y\in\Omega$
  \begin{equation}
    g_{\cH_0}(x,y):=\sum_{1\leq p,q\leq P} u_p(x)\, G^\dagger_{p,q} D[u_q](y),
  \end{equation}
  with: $\forallt 1\leq p,q\leq P$,
  \begin{equation}
    G_{p,q} := \braket{D[u_p]}{D[u_q]}_{\dL^2(\Omega\to\RR,\mu)}.
  \end{equation}
\end{Prop}

\paragraph{Our goal.}
We aim to
\begin{enumerate}[label=(\roman*)]
  \item generalize \citet[Proposition~3]{schwencke2025anagram} to arbitrary Reproducing Kernel Hilbert Spaces;
  \item establish a direct connection to the regularization framework introduced earlier. This will provide a novel reinterpretation of the Green's function in the regularized operator setting.
\end{enumerate}

\paragraph{Operator framework.}
Consider the operator $D:\cH_D \to \dL^2(\Omega,\mu)$ from \cref{app:why-regularization-is-necessary}, and assume that there exists an RKHS $\cH_0$ isometrically embedded in $\cH_D$ (for instance, any finite-dimensional RKHS, see \citealp[Corollary~1]{schwencke2025anagram}).
For \citet[Definition~9]{schwencke2025anagram} to be well-posed, the range $D[\cH_0]$ must be a closed subspace of $\dL^2(\Omega,\mu)$. As argued earlier, this is guaranteed if $D$ is continuously invertible: indeed, in this case
\begin{equation}
  D[\cH_0] = (D^{-1})^{-1}[\cH_0],
\end{equation}
and the inverse image of a closed subspace under a continuous operator is closed.

\paragraph{Key observation.}
Thus, to generalize \citet[Proposition~3]{schwencke2025anagram}, we require $D$ to be continuously invertible. Conveniently, this is precisely the property enforced by the regularization schemes we introduced earlier.

In what follows, we first focus on the cutoff regularization, which offers the clearest interpretation in terms of Green's functions. We then briefly revisit the case of Ridge regression.
Before delving further into our main goal, let us first establish two general facts.

\begin{LM}\label{LM:equivalent-norm-yield-RKHS}
  Let $\big(\cH_0,\norm_{\cH_0}\big)$ be an RKHS on a set $X$ with reproducing kernel $k$. Suppose that $\norm_{bis}$ is a norm equivalent to $\norm_{\cH_0}$. Then $\big(\cH_0,\norm_{bis}\big)$ is also an RKHS.
\end{LM}
\begin{proof}
  The key point is to show that there exists a reproducing kernel for the inner product $\braket{\cdot}{\cdot}_{bis}$ associated with $\norm_{bis}$.
  Our argument follows the simple reasoning in \citet[Definitions~1--2]{paulsenIntroductionTheoryReproducing2016}.

  Since, for every $x \in X$, the point evaluation functional
  \begin{equation}
    \delta_x:u \in \cH_0 \mapsto u(x)
  \end{equation}
  is continuous with respect to $\norm_{\cH_0}$ by the definition of an RKHS, it is also continuous with respect to the equivalent norm $\norm_{bis}$
  Therefore, by the Riesz representation theorem, for each $x \in X$, there exists a unique element $k^{bis}_x \in \cH_0$ such that for all $u \in \cH_0$
  \begin{equation}
    \braket{k^{bis}_x}{u}_{bis} = u(x).
  \end{equation}
  In particular, this defines a reproducing kernel for the norm $\norm_{bis}$, given by
  \begin{equation}
    k_{bis}(x,y) = \braket{k^{bis}_x}{k^{bis}_y}_{bis} = k^{bis}_x(y), \qquad \forall x,y \in X.
  \end{equation}
  Hence $\big(\cH_0,\norm_{bis}\big)$ is indeed an RKHS.
\end{proof}

\begin{LM}\label{LM:adjoint-and-isometry}
  Let $\cH_A,\cH_B$ be two Hilbert spaces.
  If $U:\cH_\cA\to\cH_\cB$ is an isometry, then
  \begin{align}
    U^*U & = \dI_{\cH_\cA}, \quad &
    UU^* & = \Pi_{\Ima U}.
  \end{align}
  In particular $\Ima U$ is closed in $\cH_\cB$.
\end{LM}
\begin{proof}
  The first identity follows from the fact that for all $x,y\in\cH_\cA$,
  \begin{equation}
    \braket{x}{U^*U[y]}_{\cH_\cA}
    = \braket{U[x]}{U[y]}_{\cH_\cB}
    = \braket{x}{y}_{\cH_\cA}.
  \end{equation}
  Thus $\left(U^*U(y)-y\right)\in \cH_\cA^\bot$, \ie $U^*U=\dI_{\cH_\cA}$.
  For the second, the key point is to show that $\Ima U$ is closed, i.e.\ $\Ima U=\overline{\Ima U}$.

  Let $y\in\overline{\Ima U}$, and $(y_n)\in\Ima U^\NN$ with $y_n\to y$.
  Since $(y_n)$ is Cauchy, and $y_n=U(x_n)$ for some $(x_n)\in\cH_\cA^\NN$, we have
  \begin{equation}
    \norm[U(x_n)-U(x_m)]_{\cH_\cB}
    = \norm[x_n-x_m]_{\cH_\cA},
  \end{equation}
  so $(x_n)$ is also Cauchy and converges to $x\in\cH_\cA$, since $\cH_\cA$ is complete.
  Since $U$ is an isometry, we have for all $x\in\cH_\cA$
  \begin{equation}
    \norm[U(x)]_{\cH_\cB} = \norm[x]_{\cH_\cA}.
  \end{equation}
  In particular, $U$ is bounded with operator norm $\norm[U]=1$, and hence continuous.
  Thus $U(x)=y$, hence $y\in\Ima U$. We conclude that $\Ima U$ is closed in $\cH_\cB$.
  Finally:
  \begin{itemize}
    \item For $y\in\Ima U$, say $y=U(x)$, we have
          \begin{equation}
            UU^*(y) = U(U^*U)(x) = U(x) = y.
          \end{equation}
    \item For $y\in (\Ima U)^\perp$, we check that $UU^*(y)=0$. Indeed, for any $z\in\cH_\cB$,
          \begin{equation}
            \braket{z}{UU^*(y)}_{\cH_\cB}
            = \braket{UU^*(z)}{y}_{\cH_\cB}
            = 0,
          \end{equation}
          since $UU^*(z)\in\Ima U$. Thus $UU^*(y)\in \cH_\cB^\bot$, \ie $UU^*(y)=0$.
  \end{itemize}
  \vspace{-.8cm}
\end{proof}

We are interested in the restriction of $D$ to the domain $\cH_0$.
Since the restriction $D^*D:\cH_D\to\cH_D$ does not, \apriori, map $\cH_0$ into itself, we first need to adapt the setting in order to apply the spectral theorem of \cref{Th:self-adjoint-op-diag}.

Because $\cH_0 \subset \cH_D$ isometrically, we have for all $u,v\in\cH_0$:
\begin{align}
  \begin{split}
    \braket{D[u]}{D[v]}_{\dL^2(\Omega,\mu)}
     & = \braket{D\big[\Pi_{\cH_0}u\big]}{D\big[\Pi_{\cH_0}v\big]}_{\dL^2(\Omega,\mu)} \\
     & = \braket{\Pi_{\cH_0}u}{D^*D\big[\Pi_{\cH_0}v\big]}_{\cH_D}                     \\
     & = \braket{u}{\big(\Pi_{\cH_0}D^*D\Pi_{\cH_0}\big)[v]}_{\cH_D},
  \end{split}
\end{align}
where we used in the last step that $\Pi_{\cH_0}$ is self-adjoint.

We can therefore apply the spectral theorem \cref{Th:self-adjoint-op-diag} to the bounded self-adjoint operator $\Pi_{\cH_0}D^*D\Pi_{\cH_0}:\cH_0\to\cH_0$, obtaining the analogue of the decomposition in \cref{eqn:spectral-decomposition-DstarD}:
\begin{align}
  \Pi_{\cH_0}D^*D\Pi_{\cH_0} & = \int_{\RR_+} \lambda \,\pi^{\cH_0}_D(\dd\lambda),
                             &
  I_{\cH_0}                  & = \int_{\RR_+} \pi^{\cH_0}_D(\dd\lambda).
\end{align}

\paragraph{Regularized spaces.}
Fixing $\alpha>0$, and analogously to \cref{eqn:Pi_HDalpha-def,eqn:HDalpha-def}, we define the regularized projection and subspace:
\begin{align}
  \Pi_{D,\cH_0}^\alpha & := \int_{\alpha^2}^{+\infty} \pi^{\cH_0}_D(\dd\lambda),
                       &
  \cH_{D,\cH_0}^\alpha & := \Ima \Pi_{D,\cH_0}^\alpha \subset \cH_0 \subset \cH_D.
\end{align}
Let $k:\Omega\times\Omega\to\RR$ be the reproducing kernel of $\cH_0$. Then, by \citet[Theorem~2.5]{paulsenIntroductionTheoryReproducing2016}, $\cH_{D,\cH_0}^\alpha$ is an RKHS with reproducing kernel
\begin{equation}
  k_{\alpha}(x,y) := \Pi_{D,\cH_0}^\alpha[k(x,\cdot)](y), \qquad \forall x,y \in \Omega.
\end{equation}

\paragraph{Norm equivalence.}
Since $\cH_{D,\cH_0}^\alpha \subset \cH_0 \subset \cH_D$, inequality in \cref{eqn:fondamental-inequality-2} remains valid, i.e. for all $u \in \cH_{D,\cH_0}^\alpha$:
\begin{equation}
  \norm[D[u]]_{\dL^2(\Omega,\mu)} \leq \norm[u]_{\cH_D}.
\end{equation}
Furthermore, by an argument entirely analogous to \cref{eqn:coercivity-of-D-in-HDalpha}, we also have
\begin{equation}
  \norm[D[u]]_{\dL^2(\Omega,\mu)} \geq \alpha \norm[u]_{\cH_D}, \qquad \forall u \in \cH_{D,\cH_0}^\alpha.
\end{equation}

In particular, the functional
\begin{equation}
  \norm_D:\left\{
  \begin{array}{lll}
    \cH_{D,\cH_0}^\alpha & \to     & \RR                             \\
    u                    & \mapsto & \norm[D[u]]_{\dL^2(\Omega,\mu)}
  \end{array}\right.
\end{equation}
defines a norm equivalent to $\norm_{\cH_D}$ on $\cH_{D,\cH_0}^\alpha$.
By \cref{LM:equivalent-norm-yield-RKHS}, the pair $\big(\cH_{D,\cH_0}^\alpha,\norm_D\big)$ is itself an RKHS with a reproducing kernel $k_D$.

\paragraph{Isometry property.}
The crucial observation is that $D$ is an isometry with respect to this norm. Indeed, for all $u,v\in\cH_{D,\cH_0}^\alpha$,
\begin{equation}
  \braket{u}{v}_D = \braket{D[u]}{D[v]}_{\dL^2(\Omega,\mu)}.
\end{equation}
This allows us to characterize the associated Green’s function.

\GreenFunction*
\begin{proof}
  For all $f \in \dL^2(\Omega,\mu)$ and $x \in \Omega$,
  \begin{align}
    \begin{split}
      \int_\Omega g_D(x,s)f(s)\mu(\dd s)
       & = \braket{g_D(x,\cdot)}{f}_{\dL^2(\Omega,\mu)}    \\
       & = \braket{D[k_D(x,\cdot)]}{f}_{\dL^2(\Omega,\mu)} \\
       & = \braket{k_D(x,\cdot)}{D^*f}_{D}                 \\
       & = \big(D^*f\big)(x).
    \end{split}
  \end{align}
  Since $D$ is an isometry, \cref{LM:adjoint-and-isometry} gives $DD^* = \Pi_{D[\cH_{D,\cH_0}^\alpha]}$. Therefore,
  \begin{equation}
    D\Big[x \mapsto \int_\Omega g_D(x,s)f(s)\mu(\dd s)\Big]
    = D\big[D^*f\big]
    = \Pi_{D[\cH_{D,\cH_0}^\alpha]} f,
  \end{equation}
  which precisely shows that $g_D$ is a generalized Green’s function.
\end{proof}

The key insight of \cref{Th:general-generalized-green-function} is that, in the PINNs setting—and most notably in our algorithm—we implicitly construct the reproducing kernel $k_D$ associated with the norm $\norm_D$ on the regularized tangent space $T^\alpha_\theta\cM$ of the neural network manifold $\cM$, at cutoff level $\alpha$. This kernel is precisely the PINNs NNTK introduced by \citet{schwencke2025anagram}.

A crucial consequence is that the regularization of the Gram matrix is not merely a ``numerical trick'' to guarantee stability: it is the very mechanism that ensures the Green’s function is well defined.

\paragraph{Conceptual interpretation.}
This perspective also offers a profound interpretation of the procedure: rather than attempting to invert the operator $D$ directly, we build a kernel $k_D$ whose associated metric makes $D$ an isometry, and thus ensures that $D^*$ acts as the generalized left-inverse of $D$.
The magic of the kernel lies in the following facts:
\begin{enumerate}[label=(\roman*)]
  \item We never need to compute $D^*$ explicitly, since it is implicitly encoded in the relation
        \begin{equation}
          \braket{D[k_D(x,\cdot)]}{f}_{\dL^2(\Omega,\mu)}
          = \braket{k_D(x,\cdot)}{D^*f}_{D}.
        \end{equation}
  \item The same formula allows us to directly evaluate the generalized solution $D^*f$: indeed, for all $x\in\Omega$, the reproducing property gives
        \begin{equation}
          D^*f(x) = \braket{k_D(x,\cdot)}{D^*f}_{D}.
        \end{equation}
\end{enumerate}

\paragraph{Comparison with Ridge regression.}
An analogous analysis holds for Ridge regression. However, instead of inverting $D$ ``via isometry,'' we invert the augmented operator $\big(D,\alpha\dI_{\cH_D}\big)$.

\section{Proofs}
\subsection{Statement and proof of \cref{Prop:Natural-Gradient-formula}}\label{app:proof-gram-matrix-nat-grad-formula}
We start by recalling the following statements from \citet{schwencke2025anagram}.

\begin{Def*}[{\citealp[Definition 4]{schwencke2025anagram}}]
  A linear operator $A:\cH\to\cH$ is an integral operator given that there is $k:\Omega\times\Omega\to\KK$, $\KK\in\{\RR,\CC\}$, such that: $\forallt f\in\cH$, $\forallt x\in\Omega$
  \begin{equation}\label{eqn:integral-integrator-def}
    A\big(f\big)(x)=\braket{k(x,\cdot)}{f}_\cH.
  \end{equation}
\end{Def*}
\begin{LM*}[{\citealp[Lemma 1]{schwencke2025anagram}}]
  Let us be $\cH_0:=\Span(u_p : 1\leq p\leq P)\subset\cH$ and consider the Gram matrix $G_{pq} := \braket{u_p}{u_q}_{\cH}$ of $(u_p)$ and its eigen-decomposition $G=U\Delta^2U^t$. Then:
  \begin{literaleq}{eqn:SVD-left-basis}
    L_p:=\sum_{1\leq q\leq P}u_qU_{q,p}\Delta^\dagger_p,
  \end{literaleq}
  is an orthonormal basis of $\cH_0$.
  In particular, $\Pi_{\cH_0}$ is an integral operator whose kernel is:
  \begin{equation}
    k(x,y) = \sum_{1\leq p,q\leq P} u_p(x) G^\dagger_{p,q} u_q(y).
  \end{equation}
  Furthermore $L_p$ are the left-singular vector of the so-called \textbf{synthesis} operator\footnote{Name and notation are taken from \citet{adcockFramesNumericalApproximation2019}.}:
  \begin{equation}\label{eqn:synthesis-operator-def}
    \cT:\left\{\begin{array}{lll}
      \RR^P  & \to     & \cH_0                                   \\
      \alpha & \mapsto & \sum\limits_{1\leq p\leq P}\alpha_p u_p
    \end{array}\right..
  \end{equation}
\end{LM*}

\begin{Prop}\label{Prop:Natural-Gradient-formula}
  Given the scalar loss
  \begin{equation}\label{eqn:scalar-loss-def}
    \ell(\vtheta):=\cL\!\left(u_{\vtheta}\right)
    \stackrel{(\ref{eqn:functional-loss})}{=}
    \tfrac{1}{2}\,\norm[u_{\vtheta}-f]_{\dL^2(\Omega, \mu)}^2,
  \end{equation}
  the Natural Gradient update of \cref{eqn:functional-natural-gradient-update}
  \begin{align*}
    \literalref{eqn:functional-natural-gradient-update}\tag{\ref{eqn:functional-natural-gradient-update}}
  \end{align*}
  can be equivalently written as
  \begin{align*}
    \literalref{eqn:matrix-formula-natural-gradient-update}.
    \tag{\ref{eqn:matrix-formula-natural-gradient-update}}
  \end{align*}
\end{Prop}

\begin{proof}
  Since the tangent space $T_{\vtheta}\mathcal{M}$ of \cref{eqn:tangent-space}:
  \begin{equation*}
    \literalref{eqn:tangent-space}
    \tag{\ref{eqn:tangent-space}},
  \end{equation*}
  is finite-dimensional, we may invoke \citet[Lemma~1]{schwencke2025anagram}. This result shows that the \textbf{Natural Neural Tangent Kernel (NNTK)}, given by
  \begin{align}\label{eqn:NNTK-def-gram}
    {NNTK}_\vtheta(x,y) & :=\sum_{1\leq p,q\leq P} \big(\partial_p u_{\vtheta}(x)\big)\,{G_\vtheta^\dagger}_{pq}\,\big(\partial_q u_{\vtheta}(y)\big)^t,
                        &
    {G_{\vtheta}}_{p,q} & :=\braket{\partial_p u_{\vtheta}}{\partial_q u_{\vtheta}}_\cH,
  \end{align}
  is the kernel of the orthogonal projection $\Pi^\bot_{T_{\vtheta}\cM}$ onto $T_{\vtheta}\cM$. Therefore, by \cref{eqn:integral-integrator-def}, for all $x\in\Omega$,
  \begin{align}
    \begin{split}
      \Pi^\bot_{T_{\vtheta}\cM}\!\left(\nabla\cL_{|u_{\vtheta}}\right)(x)
       & = \braket{NNTK_\vtheta(x,\cdot)}{\nabla\cL_{|u_{\vtheta}}}_\cH \\
       & \stackrel{(\ref{eqn:NNTK-def-gram})}{=}
      \sum_{1\leq p,q\leq P} \partial_p u_{\vtheta}(x)\,{G_\vtheta^\dagger}_{pq}\,
      \braket{\partial_q u_{\vtheta}}{\nabla\cL_{|u_{\vtheta}}}_\cH.
    \end{split}
    \label{eqn:projection-formula}
  \end{align}

  Next, note that
  \begin{equation}\label{eqn:partial-ell-formula}
    \braket{\partial_q u_{\vtheta}}{\nabla\cL_{|u_{\vtheta}}}_\cH
    = \dd\cL_{|u_{\vtheta}}\!\big(\partial_q u_{\vtheta}\big)
    \stackrel{\text{chain rule}}{=}
    \partial_q \cL(u_{\vtheta})
    \stackrel{(\ref{eqn:scalar-loss-def})}{=}
    \partial_q \ell(\vtheta).
  \end{equation}

  Therefore, by linearity of $\dd u_{\vtheta}^\dagger$,
  \begin{equation}
    \dd u_{\vtheta}^\dagger\!\left(\Pi^\bot_{T_{\vtheta}\cM}\!\left(\nabla\cL_{|u_{\vtheta}}\right)\right)
    \stackrel{(\ref{eqn:projection-formula}),(\ref{eqn:partial-ell-formula})}{=}
    \sum_{1\leq p,q\leq P}
    \dd u_{\vtheta}^\dagger\!\big(\partial_p u_{\vtheta}\big)\,
    {G_\vtheta^\dagger}_{pq}\,\partial_q\ell(\vtheta).
  \end{equation}

  Finally, observe that $\partial_p u_{\vtheta}=\dd u_{\vtheta}(\ve^{(p)})$, where $\ve^{(p)}$ is the $p$-th canonical basis vector of $\RR^P$.
  If $\dd u_{\vtheta}$ were invertible, we would directly obtain
  \begin{equation}
    \dd u_{\vtheta}^\dagger\!\big(\partial_p u_{\vtheta}\big)=\ve^{(p)},
  \end{equation}
  which would complete the argument. However, this invertibility does not hold in general.

  To address this, recall that $\dd u_{\vtheta}$ can be identified with the synthesis operator $\cT$ introduced in \cref{eqn:synthesis-operator-def} of \citet[Lemma~1]{schwencke2025anagram}. From the final part of that lemma, we know that $\Ima\dd u_{\vtheta}^\dagger=\Ima G_\vtheta^\dagger$. Consequently,
  \begin{equation}\label{eqn:exact-implicit-inversion-trick}
    G_\vtheta^\dagger \ve^{(p)}
    = G_\vtheta^\dagger \dd u_{\vtheta}^\dagger\!\big(\partial_p u_{\vtheta}\big).
  \end{equation}

  Putting all pieces together yields the desired update rule, thereby completing the proof.
\end{proof}

\subsection{Ridge-regression implementation ANaGRAM}
\label{app:proof-SVD-ridge-reg}
In the following, we show that a Ridge-regression can be implemented in ANaGRAM's update rule given by \cref{eqn:eng-computation-trick}.
\begin{restatable}{Prop}{SVDSoftCutoffReg}\label{Prop:SVD-soft-cutoff-reg}
  A Ridge-regression can be implemented in the SVD-based update \cref{eqn:eng-computation-trick} by replacing the pseudo-inverse $\Dsing^\dagger$ 
  with
  \begin{equation}\literallabel{eqn:ridge-regression-SVD}
    {\left(\frac{\Dsi}{\Dsi^2+S\alpha}\right)_{1\leq i\leq \FRk}}.
  \end{equation}
\end{restatable}
\begin{proof}
  As shown in \cite[Section E]{schwencke2025anagram}, the ANaGRAM's update of \cref{eqn:eng-computation-trick}:
  \begin{align*}
    \literalref{eqn:eng-computation-trick},\tag{\ref{eqn:eng-computation-trick}}
  \end{align*}
  is equivalent to the update with the empirical matrix $\widehat{\cG}_{\vtheta}$:
  \begin{align}\label{eqn:marius-computation}
    \vtheta_{t+1}             & \gets \vtheta_t - \eta\,\widehat{\cG}_{\vtheta_t}^\dagger\nabla\ell(\vtheta_t)\,; &
    \widehat{\cG}_{\vtheta_t} & :=\frac{1}{S}\widehat{\phi}_{\vtheta_t}^t\widehat{\phi}_{\vtheta_t},
  \end{align}
  where $\ell$ is defined in \cref{eqn:empirical-quadratic-loss}:
  \begin{equation*}
    \literalref{eqn:empirical-quadratic-loss}.\tag{\ref{eqn:empirical-quadratic-loss}}
  \end{equation*}
  Thus, we get immediately
  \begin{equation}\label{eqn:nable-ell-expression}
    \nabla\ell(\vtheta_t) = \frac{1}{S}\widehat{\phi}^t\ELgrad_\vtheta
    =\frac{1}{S}\Using\Dsing\widehat{V}^t_\vtheta\ELgrad_\vtheta,
  \end{equation}
  where we used the SVD decomposition of $\feat$:
  \begin{align}\label{eqn:SVD-decomposition-phihat-alone}
    \literalref{eqn:SVD-decomposition-phihat}.
  \end{align}
  Using \cref{eqn:SVD-decomposition-phihat-alone} again, we have
  \begin{equation}
    \widehat{\cG}_{\vtheta} = \frac{1}{S}\Using\widehat{\Delta}^2_\vtheta\widehat{U}^t_\vtheta,
  \end{equation}
  thus for all $\alpha>0$
  \begin{equation}
    \widehat{\cG}_{\vtheta} + \alpha I_d
    = \frac{1}{S}\Using\widehat{\Delta}^2_\vtheta\widehat{U}^t_\vtheta + \alpha \Using\Using^t
    = \Using\left(\diag\left(\frac{\widehat{\Delta}^2_{\vtheta_i}}{S} + \alpha\right)_{1\leq i\leq \FRk}\right)\widehat{U}^t_\vtheta,
  \end{equation}
  which implies
  \begin{equation}
    \left(\widehat{\cG}_{\vtheta} + \alpha I_d\right)^{-1}
    = \Using\left(\diag\left(\frac{S}{\widehat{\Delta}^2_{\vtheta_i} + S\alpha}\right)_{1\leq i\leq \FRk}\right)\widehat{U}^t_\vtheta.
  \end{equation}
  This finally yields
  \begin{align}
    \left(\widehat{\cG}_{\vtheta} + \alpha I_d\right)^{-1}\nabla\ell(\vtheta_t)
     & \stackrel{(\ref{eqn:nable-ell-expression})}{=} \Using\left(\diag\left(\frac{S}{\widehat{\Delta}^2_{\vtheta_i} + S\alpha}\right)_{1\leq i\leq \FRk}\right)\widehat{U}^t_\vtheta \frac{1}{S}\Using\Dsing\widehat{V}^t_\vtheta\ELgrad_\vtheta \\
     & =\Using\left(\diag\left(\frac{\Dsi}{\widehat{\Delta}^2_{\vtheta_i} + S\alpha}\right)_{1\leq i\leq \FRk}\right)\widehat{V}^t_\vtheta\ELgrad_\vtheta,
  \end{align}
  which conludes the proof.
\end{proof}
\subsection{Proof of \cref{Prop:RCE-as-N-components-projection}}\label{app:proof-RCE-as-N-components-projection}

To prove \cref{Prop:RCE-as-N-components-projection}, we need the following lemma:
\begin{restatable}{LM}{RCENMFormula}\label{LM:RCE_N-RCE_M_formula}
  For $1\leq M\leq N\leq \FRk$:
  \begin{equation}\label{eqn:diff-of-squared-RCE}
    \left(\rce^S_M\right)^2-\left(\rce^S_N\right)^2 = \frac{1}{S}\norm[\Pi^M_N\Vsing^T\ELgrad]^2_{\RR^S}.
  \end{equation}
\end{restatable}
\begin{proof}
  Let us first recall the definition of the $\rce_N^S$ in \cref{eqn:reconstruction-loss-orthogonal}, namely
  \begin{equation}
    \literalref{eqn:reconstruction-loss-orthogonal}\tag{\ref{eqn:reconstruction-loss-orthogonal}}.
  \end{equation}
  Fixing $1\leq N\leq M\leq \FRk$ and
  applying  the same reasoning as in \cref{eqn:development-RCE} to $\rce_M^S$ and $\rce_N^S$ (see the proof of \cref{Prop:RCE-as-N-components-projection} in \cref{app:proof-RCE-as-N-components-projection}), we get
  \begin{align}
    S\left(\rce^S_M\right)^2
     & =\ELgrad_\vtheta^t\ELgrad_\vtheta
    -\sum_{p=1}^M\left(\widehat{V}_{\vtheta_p}^t\ELgrad_\vtheta\right)^2;
     &
    S\left(\rce^S_N\right)^2
     & =\ELgrad_\vtheta^t\ELgrad_\vtheta
    -\sum_{p=1}^N\left(\widehat{V}_{\vtheta_p}^t\ELgrad_\vtheta\right)^2,
  \end{align}
  and therefore
  \begin{align}
    \begin{split}
      S\left(\left(\rce^S_N\right)^2-\left(\rce^S_M\right)^2\right)
       & =
      \sum_{p=1}^N\left(\widehat{V}_{\vtheta_p}^t\ELgrad_\vtheta\right)^2
      -
      \sum_{p=1}^M\left(\widehat{V}_{\vtheta_p}^t\ELgrad_\vtheta\right)^2             \\
       & \stackrel{M\leq N}{=}
      \sum_{p=M+1}^N\left(\widehat{V}_{\vtheta_p}^t\ELgrad_\vtheta\right)^2
      =\sum_{p=M+1}^N\left(\ve^{(p)^t}\Vsing^t\ELgrad_\vtheta\right)^2                \\
       & =\sum_{p=M+1}^N\left(\ELgrad_\vtheta^t\Vsing\ve^{(p)}\right)
      \left(\ve^{(p)^t}\Vsing^t\ELgrad_\vtheta\right)                                 \\
       & =\ELgrad_\vtheta^t\Vsing
      \underbracket{\left(\sum_{p=M+1}^N\ve^{(p)}\ve^{(p)^t}\right)}_{=\Pi_N^M~ by~ \cref{eqn:PiMN-def}}
      \Vsing^t\ELgrad_\vtheta                                                         \\
       & =\braket{\Vsing^t\ELgrad_\vtheta}{\Pi_N^M\Vsing^t\ELgrad_\vtheta}_{\RR^S}    \\
       & \stackrel{\substack{
      \Pi_N^{M^2}=\Pi_N^M                                                             \\
          \Pi_N^{M^t}=\Pi_N^M}
      }{=}
      \braket{\Pi_N^M\Vsing^t\ELgrad_\vtheta}{\Pi_N^M\Vsing^t\ELgrad_\vtheta}_{\RR^S} \\
       & =\norm[\Pi_N^M\Vsing^t\ELgrad_\vtheta]_{\RR^S}^2,
    \end{split}
  \end{align}
  where we use in the penultimate equality, the fact that $\Pi^M_N$ is an orthogonal projection. 
\end{proof}
\begin{Rk}
  The above lemma provides an interesting property that gives a further understanding of what is happening during the flattening, \ie $\rce^S_M - \rce^S_N \approx 0$. In particular, as $ \left(\rce^S_M\right)^2-\left(\rce^S_N\right)^2 =  \left(\rce^S_M - \rce^S_N\right)\left(\rce^S_M + \rce^S_N\right) $, therefore flattening for the components in the range $[\Nflat, \rkk]$ means that $\frac{1}{S}\norm[\Pi^M_N\Vsing^T\ELgrad]^2_{\RR^S}  \approx 0$. In other words, the problem is "learned" for those components, as the projection of the functional gradient (which is propotional to the error) on their corresponding span is null. The proof of this lemma is provided in \cref{app:proof-RCE-as-N-components-projection}. 
\end{Rk}
\RCEproperties*
\begin{proof}
  The first statement is a direct consequence of \cref{LM:RCE_N-RCE_M_formula} proven above.

  Let us now show that the second statement takes place. Since $\FLgrad_{\utheta}\in\dL^2(\Omega,\mu)$ and $\Ima\dd \utheta\subset\dL^2(\Omega,\mu)$, the law of large numbers yields that for all $1\leq p,q\leq P$
  \begin{align}
    \label{eqn:squared-norm-nabla-cL}
    \lim\limits_{S\to\infty}\frac{1}{S}\sum_{i=1}^S \left[ \FLgrad_{\utheta}(x_i) \right]^2
     & =\lim\limits_{S\to\infty}\frac{1}{S}
    \ELgrad_\vtheta^t
    \ELgrad_\vtheta
    =\int_\Omega \left[\FLgrad_{\utheta}(x)\right]^2\mu(\dd x)\quad\textit{a.s},      \\
    \lim\limits_{S\to\infty}\frac{1}{S}\sum_{i=1}^S \partial_p \utheta(x_i)\FLgrad_{\utheta}(x_i)
     & =\lim\limits_{S\to\infty}\frac{1}{S}
    \widehat{\phi}_{\vtheta_p}^t
    \ELgrad_\vtheta
    =\int_\Omega\partial_p \utheta(x)\FLgrad_{\utheta}(x)\mu(\dd x)\quad\textit{a.s}, \\
    \lim\limits_{S\to\infty}\frac{1}{S}\sum_{i=1}^S \partial_p \utheta(x_i)\partial_q \utheta(x_i)
     & =\lim\limits_{S\to\infty}\frac{1}{S}
    \widehat{\phi}_{\vtheta_p}^t\widehat{\phi}_{\vtheta_q}
    =\int_\Omega\partial_p \utheta(x)\partial_q \utheta(x)\mu(\dd x)\quad\textit{a.s}.
  \end{align}

  \todoAS[inline]{It is unusual to see this constant switch between $\widehat{\phi}_{\vtheta_q}$ and $\FLgrad_{\utheta}(x)$. While essentially one is a discrete version of the other one. I wonder if we can either make therm more similar. I think at the very least we should use $\widehat{\phi}$ wherever we imply matrix-vector form and the other one for the functional representation.}

  In particular, this implies
  \begin{align}
    \lim\limits_{S\to\infty}\frac{1}{S}
    \feat^t\feat
    =G_\vtheta=U_\vtheta\Delta_\vtheta^2U_\vtheta^t\quad\textit{a.s}.
  \end{align}
  Since the eigenvectors \todoAS{I am not sure I am familiar with this result, maybe a citation here?} (and eigenvalues) are continuous functions of the matrix coefficients (by polynomial dependence through \todoAS{which characteristic polynomial?} the characteristic polynomial) and taking into account that $ \frac1S  \feat^t\feat = \frac1S \Using\Delta_\vtheta^2 \Using^t$, this yields
  \begin{align}
    \lim\limits_{S\to\infty}\Using & = U_\vtheta\quad\textit{a.s};
                                   &
    \lim\limits_{S\to\infty}\frac{1}{S}
    \Dsing^2
                                   & =\Delta_\vtheta^2\quad\textit{a.s}.
  \end{align}
  \todoAS[inline]{At first, it was not clear for me why $\lim\limits_{S\to\infty}\Using  = U_\vtheta$ holds, but then I remembered that the dimensionality of $\Using$ is $P\times \FRk$, so clearly if $S\to \infty$ then $\Using$ stops to depend on $S$, maybe through some polynomial which is continuous as you mentioned above. So this being said, I think it will be more clear if we clarify what polynomial you mean earlier and mention the argument about dimensionalities, what do you think?}
  By continuity of the square root and the inverse on $\RR^*_+$, we get that for all $1\leq p\leq P$ such that $\Delta_{\vtheta_p}>0$
  \begin{align}
    \lim\limits_{S\to\infty}\sqrt{S}\widehat{\Delta}_{\vtheta_p}^{-1}
    =\Delta_{\vtheta_p}^{-1}\quad\textit{a.s},
  \end{align}

  and thus for all $1\leq p\leq P$ such that $\Delta_{\vtheta_p}>0$, we have \textit{almost surely}
  \todoAS[inline]{So in the first equality we used $\feat = \Vsing \Dsing \widehat{U}^t_{\vtheta} \Rightarrow \Vsing = \feat \Using  \widehat{\Delta}^{-1}_{\vtheta} $. Also is $\sum_{q=1}^P\ve^{(q)}\ve^{(q)^t} = I_P$? I think we should mention all of it, otherwise it is harder to check the first equality.}

  \todoAS[inline]{By the way, why $\widehat{V}$ is with $\vtheta_p$, but $\ELgrad$ with just $\vtheta$? Also after the first equality, why $\widehat{\phi}$ is with $\vtheta$, while the other elements from $\widehat V$ are with  $\vtheta_p$?}
  \begin{align}
    \label{eqn:identification-of-limit-cn}
    \begin{split}
      \lim\limits_{S\to\infty}\frac{1}{\sqrt{S}}
      \widehat{V}_{\vtheta_p}^T\ELgrad_{\vtheta}
       & =\lim\limits_{S\to\infty}
      \sqrt{S}\widehat{\Delta}_{\vtheta_p}^{-1}\widehat{U}_{\vtheta_p}^t
      \left(\sum_{q=1}^P\ve^{(q)}\ve^{(q)^t}\right)
      \frac{1}{S}\feat^t\ELgrad_{\vtheta}                                                                           \\
       & =\sum_{q=1}^P
      \left(\lim\limits_{S\to\infty}
      \sqrt{S}\widehat{\Delta}_{\vtheta_p}^{-1}\right)
      \left(\lim\limits_{S\to\infty}\widehat{U}_{\vtheta_p}^t\ve^{(q)}\right)
      \left(\lim\limits_{S\to\infty}\frac{1}{S}\widehat{\phi}_{\vtheta_q}^t\ELgrad_{\vtheta}\right)                 \\
       & =\sum_{q=1}^P
      \Delta_{\vtheta_p}^{-1}U_{\vtheta_p}^t\ve^{(q)}\int_\Omega\partial_q \utheta(x)\FLgrad_{\utheta}(x)\mu(\dd x) \\
       & =\int_\Omega\dd \utheta\left(U_{\vtheta_p}\Delta_{\vtheta_p}^{-1}\right)(x)\FLgrad_{\utheta}(x)\mu(\dd x)  \\
       & =\int_\Omega V_{\vtheta_p}(x)\FLgrad_{\utheta}(x)\mu(\dd x),
    \end{split}
  \end{align}
  where we used in  the last equality,
  the identification of the singular vectors of $\dd \utheta$
  in \cite[Lemma 1 p. 24, section C.2]{schwencke2025anagram}.
  \todoAS[inline]{To be honest, I didn't completely understand the equlity before the last one. It looks like a definition of the $d u{\vtheta}$, but where is $\ve^{(q)}$? In general, don't hesitate to remind the definitions in the proofs, especially if you use them in your derivations, like it is done for $\rce$ below.}
  Returning to the definition of the $\rce_N^S$ in \cref{eqn:reconstruction-loss-orthogonal}, namely
  \begin{equation}
    \literalref{eqn:reconstruction-loss-orthogonal}\tag{\ref{eqn:reconstruction-loss-orthogonal}},
  \end{equation}
  we get
  \begin{align}
    \label{eqn:development-RCE}
    \begin{split}
      S\left(\rce^S_N\right)^2
       & = \braket{\Vsing\Pi^0_N\Vsing^t\ELgrad_\vtheta - \ELgrad_\vtheta}{\Vsing\Pi^0_N\Vsing^t\ELgrad_\vtheta - \ELgrad_\vtheta}_{\RR^S} \\
       & =
      \ELgrad_\vtheta^t\ELgrad_\vtheta
      + \ELgrad_\vtheta^t\Vsing\overbracket{\Pi^0_N\underbracket{\Vsing^t\Vsing}_{=I_d}\Pi^0_N}^{=\Pi^0_N}\Vsing^t\ELgrad_\vtheta
      -2 \ELgrad_\vtheta^t\Vsing\Pi^0_N\Vsing^t\ELgrad_\vtheta                                                                             \\
       & =\ELgrad_\vtheta^t\ELgrad_\vtheta
      -\ELgrad_\vtheta^t\Vsing\Pi^0_N\Vsing^t\ELgrad_\vtheta                                                                               \\
       & =\ELgrad_\vtheta^t\ELgrad_\vtheta
      -\ELgrad_\vtheta^t\Vsing\left(\sum\limits_{p=1}^N\ve^{(p)}\ve^{(p)^t}\right)\Vsing^t\ELgrad_\vtheta                                  \\
       & =\ELgrad_\vtheta^t\ELgrad_\vtheta
      -\sum_{p=1}^N\left(\widehat{V}_{\vtheta_p}^t\ELgrad_\vtheta\right)^2,
    \end{split}
  \end{align}
  where in the second equality, we use the fact that $\Vsing$ is orthogonal and $\Pi_N^0$ is a projection. Combining \cref{eqn:squared-norm-nabla-cL,eqn:identification-of-limit-cn}, this yields 
  \begin{align}
    \lim\limits_{S\to\infty}\left(\rce^S_N\right)^2
     & =\int_\Omega\FLgrad_{\utheta}(x)^2\mu(\dd x)
    -\sum_{p=1}^N\left(\int_\Omega V_{\vtheta_p}(x)\FLgrad_{\utheta}(x)\mu(\dd x)\right)^2\quad\textit{a.s}. \label{eqn:rce_interm}
  \end{align}
  By Fubini's theorem, we have \textit{almost surely}
  \begin{align}
    \sum_{p=1}^N\left(\int_\Omega V_{\vtheta_p}(x)\FLgrad_{\utheta}(x)\mu(\dd x)\right)^2
     & = \int_{\Omega^2} \FLgrad_{\utheta}(x)
    \left(\sum_{p=1}^NV_{\vtheta_p}(x)V_{\vtheta_p}(y)\right)
    \FLgrad_{\utheta}(y)\mu(\dd x) \mu(\dd y)\nonumber                                                            \\
     & = \int_{\Omega} \FLgrad_{\utheta}(x)
    \Pi^\bot_{\Span(V_{\vtheta_i}\,:\,1\leq i\leq N)}\FLgrad_{\utheta}(x)\mu(\dd x)       \label{eqn:fubini}      \\
     & =\norm[\Pi^\bot_{\Span(V_{\vtheta_i}\,:\,1\leq i\leq N)}\FLgrad_{\utheta}]^2_{\dL^2(\Omega,\mu)},\nonumber
  \end{align}
  where in the second equality, we used \cite[Theorem 4 p. 23, section C.2]{schwencke2025anagram} and the fact that $\left(\Pi^\bot_{\Span(V_{\vtheta_i}\,:\,1\leq i\leq N)}\right)^2=\Pi^\bot_{\Span(V_{\vtheta_i}\,:\,1\leq i\leq N)}$ in the third. Therefore,
  from \cref{eqn:rce_interm} and \cref{eqn:fubini}
  \begin{align}
    \lim\limits_{S\to\infty}\left(\rce^S_N\right)^2 & =\norm[\FLgrad_{\utheta}]^2_{\dL^2(\Omega,\mu)}
    -\norm[\Pi^\bot_{\Span(V_{\vtheta_i}\,:\,1\leq i\leq N)}\FLgrad_{\utheta}]^2_{\dL^2(\Omega,\mu)}\quad\textit{a.s},                                                                     \\
                                                    & =\norm[\FLgrad_{\utheta}-\Pi^\bot_{\Span(V_{\vtheta_i}\,:\,1\leq i\leq N)}\FLgrad_{\utheta}]^2_{\dL^2(\Omega,\mu)}\quad\textit{a.s},
  \end{align}
  where in the second equality, we use in the reverse order a reasoning similar to \cref{eqn:development-RCE}.
  Finally, we obtain
  \begin{align}
    \norm[\FLgrad_{\utheta}-\Pi^\bot_{\Span(V_{\vtheta_i}\,:\,1\leq i\leq N)}\FLgrad_{\utheta}]^2_{\dL^2(\Omega,\mu)}
    =
    \norm[\Pi^\bot_{\Span(V_{\vtheta_i}\,:\,1\leq i\leq N)^\bot}\FLgrad_{\utheta}]^2_{\dL^2(\Omega,\mu)},
  \end{align}
  which comes from the canonical decomposition in Hilbert spaces, \ie using that $\Span(V_{\vtheta_i}\,:\,1\leq i\leq N)$ is a closed subspace and
  \begin{equation}
    \FLgrad_{\utheta}
    =
    \Pi^\bot_{\Span(V_{\vtheta_i}\,:\,1\leq i\leq N)}\FLgrad_{\utheta}
    +
    \Pi^\bot_{\Span(V_{\vtheta_i}\,:\,1\leq i\leq N)^\bot}\FLgrad_{\utheta}.
  \end{equation}

  This completes the proof.
\end{proof}
\begin{Cor}\label{Cor:asymptotic-diff-of-squared-RCE}
  For $1\leq M\leq N\leq \FRk$:
  \begin{equation}
    \lim\limits_{S\to\infty}\left(\rce^S_M\right)^2-\left(\rce^S_N\right)^2
    = \norm[\Pi^\bot_{\TMN}\FLgrad_{\utheta}]_{\dL^2(\Omega)}^2
  \end{equation}
\end{Cor}
\begin{proof}
  Apply \cref{Prop:RCE-as-N-components-projection} to \cref{eqn:diff-of-squared-RCE} of \cref{LM:RCE_N-RCE_M_formula}.
\end{proof}

\end{document}